\documentclass[onefignum,onetabnum]{siamonline171218}
\def\jmpu{{\lbrack\!\lbrack \widetilde{\gamma}> \\
  <g,\widetilde{\Balpha}> \\
<g,\widetilde{\Bbeta}> \end{bmatrix}$, whereu\rbrack\!\rbrack}}

\def\sB{{\mathcal B}}
\def\sX{{\mathcal X}}
\def\sY{{\mathcal Y}}

\def\sL{{\mathcal L}}

\def\sG{{\mathcal G}}
\def\sM{{\mathcal M}}

\def\sN{{\mathcal N}}
\def\sQ{{\mathcal Q}}
\def\sF{{\mathcal F}}

\def\sR{{\mathcal R}}

\def\sK{{\mathcal K}}
\def\sE{{\mathcal E}}

\def\sG{{\mathcal G}}
\def\sP{{\mathcal P}}

\def\sA{{\mathcal A}}
\def\Balpha{\mbox{\boldmath$\alpha$}}
\def\Bbeta{\mbox{\boldmath$\beta$}}

\def\Blambda{\mbox{\boldmath$\lambda$}}

\newcommand{\LRc}[1]{\left\{ #1 \right\}}
\newcommand{\LRp}[1]{\left( #1 \right)}
\newcommand{\LRs}[1]{\left[ #1 \right]}
\newcommand{\norm}[1]{\left \Vert #1 \right \Vert}
\newcommand{\real}{{\mathbb{R}}}
\newcommand{\p}{{\mathbb{P}}}

\newcommand{\mb}[1]{\mathbf{#1}}

\def\bC{\mb{C}}

\def\bM{\mb{M}}

\newcommand{\lag}{\mathcal{L}}

\def\bR{\mb{R}}

\def\bW{\mb{W}}
\def\bX{\mb{X}}
\def\bY{\mb{Y}}

\def\bb{\mb{b}}
\def\bc{\mb{c}}

\def\bg{\mb{g}}

\def\bu{\mb{u}}

\def\bx{\mb{x}}
\def\by{\mb{y}}
\def\bz{\mb{z}}
\newcommand{\Ex}{{\mathbb{E}}}
\newcommand{\expect}{\Ex}
\newcommand{\pp}[2]{\frac{\partial #1}{\partial #2}} % adaptive partial derivatives/
\renewcommand{\epsilon}{\varepsilon}
\newcommand{\snor}[1]{\left| #1 \right|}

\newcommand{\mynote}[3]{
	\textcolor{#2}{\fbox{\bfseries\sffamily\scriptsize#1}}
		{\textsf{\emph{#3}}}
}

\newcommand{\krish}[1]{\mynote{Krish}{orange}{\textcolor{orange}{#1}}}
\usepackage{lipsum}
\usepackage{amsfonts}
\usepackage{graphicx}
\usepackage{epstopdf}
\ifpdf
  \DeclareGraphicsExtensions{.eps,.pdf,.png,.jpg}
\else
  \DeclareGraphicsExtensions{.eps}
\fi

\usepackage{enumitem}
\setlist[enumerate]{leftmargin=.5in}
\setlist[itemize]{leftmargin=.5in}

\newsiamremark{remark}{Remark}
\newsiamremark{hypothesis}{Hypothesis}
\crefname{hypothesis}{Hypothesis}{Hypotheses}
\newsiamthm{claim}{Claim}

\headers{Deep Neural Network Architecture Adaptation}{C G Krishnanunni, and Tan Bui-Thanh}

\title{%A Sequential Sparse Layerwise %Sparsifying Training and Sequential 
 An Adaptive and Stability-Promoting Layerwise Training Approach for Sparse Deep Neural Network Architecture
}

\author{C G Krishnanunni\thanks{Dept of Aerospace Engineering $\&$ Engineering Mechanics, UT  Austin
  (\email{krishnanunni@utexas.edu}).}
\and Tan Bui-Thanh \thanks{Dept of Aerospace Engineering $\&$ Engineering Mechanics, UT  Austin
  (\email{tanbui@utexas.edu}).}}
\usepackage{amsopn}

\makeatletter
\newcommand*{\addFileDependency}[1]{% argument=file name and extension
  \typeout{(#1)}% latexmk will find this if $recorder=0 (however, in that case, it will ignore #1 if it is a .aux or .pdf file etc and it exists! if it doesn't exist, it will appear in the list of dependents regardless)
  \@addtofilelist{#1}% if you want it to appear in \listfiles, not really necessary and latexmk doesn't use this
  \IfFileExists{#1}{}{\typeout{No file #1.}}% latexmk will find this message if #1 doesn't exist (yet)
}
\makeatother

\newcommand*{\myexternaldocument}[1]{%
    \externaldocument{#1}%
    \addFileDependency{#1.tex}%
    \addFileDependency{#1.aux}%
}
\usepackage{algorithm} %for writing pseudoalgorithms
\usepackage{algpseudocode} %for writing pseudoalgorithms
\usepackage{ntheorem}
\usepackage{tablefootnote}
\usepackage{mathtools}
\usepackage{amssymb}
\usepackage{comment}
\newenvironment{proof}{\paragraph{\textit{Proof:}}}{}

\myexternaldocument{ex_supplement}

\begin{document}
\maketitle
\begin{abstract}
%\it{``robustness"} is desirable for a deep neural network,  our adaptation strategy is based  on promoting  {\it{``robustness"}}
 This work presents a two-stage adaptive framework for progressively developing deep neural network (DNN) architectures that generalize well for a given training data set.
In the first stage, a layerwise training approach is adopted where a new layer is added each time and trained independently by freezing parameters in the previous layers.  
%In order to constrain the functions that should be learned by each layer,  
We impose desirable structures on the DNN by employing manifold regularization, sparsity regularization, and physics-informed terms. We introduce a $\epsilon-\delta-$ stability-promoting concept as a desirable property for a learning algorithm and show that employing manifold regularization yields a $\epsilon-\delta$ stability-promoting algorithm. Further, we also derive the necessary conditions for the trainability of a newly added layer and investigate the training saturation problem. In the second stage of the algorithm (post-processing),  a sequence of shallow networks is employed to extract information from the residual produced in the first stage, thereby improving the prediction accuracy. Numerical investigations on prototype regression and classification problems demonstrate that the proposed approach can outperform fully connected DNNs of the same size. Moreover, by equipping the physics-informed neural network (PINN) with the proposed adaptive architecture strategy to solve partial differential equations, we numerically show that adaptive PINNs not only are superior to standard PINNs but also produce interpretable hidden layers with provable stability. We also apply our architecture design strategy to solve inverse problems governed by elliptic partial differential equations.
 %suggests 
 %that the method could be employed for creating interpretable hidden layers in a deep network while outperforming equivalent baseline networks. 
 %\krish{The title has "Stability-Promoting Algorithm" since the whole work is about how to use manifold regularization to build a stability promoting algorithm. We define $\delta-$stability promoting algorithm in definition \ref{delta-def} which is the same thing as $\delta-$stability we discussed. However, it should be property of the algorithm rather than function. Once we use $\delta-$stability promoting algorithm to learn a function, we have $\delta-$stable function as defined in definition \ref{delta_deff}. This is the whole picture.}
\end{abstract}

% REQUIRED

% REQUIRED

\section{Introduction}
Deep neural networks (DNNs) have been
prominent in learning representative hierarchical features from image, video, speech, and audio inputs \cite{ellis1999size,hinton2007learning,lecun1995convolutional,montavon2010layer}.  Some of the problems associated with training such deep networks include %\cite{low2019stacking}:  
i) a possible large training set is necessary to overcome the over-fitting issue; ii) the architecture adaptability problem, e.g., any amendments to a pre-trained DNN require retraining even with transfer learning; and iii) GPU employment is almost  mandatory due to %the %huge size of the
massive network and data sizes. 
In particular, it is often unclear on the choice of depth and width of a network suitable for a specific problem. Existing neural architecture search algorithms rely on metaheuristic optimization, or combinatorial optimization, or reinforcement learning strategy to hopefully arrive at a reasonable architecture \cite{zoph2016neural, stanley2002evolving, suganuma2017genetic, elsken2018efficient, real2019regularized, balaprakash2018deephyper, miikkulainen2019evolving, liu2021survey}. However, these strategies involve training and evaluating many candidate architectures (possibly deep) in the process and are thus computationally expensive. Therefore, there is a need to develop a computationally tractable procedure for adaptively training/growing a DNN by leveraging information on the input data manifold and the underlying structure of the problems under consideration.

\subsection{Related work}

Layerwise training of neural networks is an approach that addresses the issue of the choice of depth of a neural network and the computational complexity involved with training \cite{xu1999training}. Many efforts have been proposed for growing neural architecture using this approach \cite{hettinger2017forward,kulkarni2017layer,chatterjee2017progressive,low2019stacking,xiao2019fast,nguyen2021analytic,rueda2015supervised}.  Hettinger et al. \cite{hettinger2017forward} showed that layers can be trained one at a time and the resulting DNN can generalize better. %and allows for the construction of DNNs with many sorts of learners.  
Belilovsky et al. \cite{belilovsky2019greedy} have further studied this approach for different convolutional neural networks. Kulkarni et al. \cite{kulkarni2017layer} employed a kernel analysis of the trained layerwise deep
networks. The weights for each layer are obtained by solving an optimization problem aiming at a better representation where a subsequent layer builds its representation on top of the features produced by a previous layer. Lengelle et al. \cite{lengelle1996training} conducted a layer-by-layer training of multilayer perceptron by optimizing an objective function for internal representations while avoiding any computation of the network's outputs. Bengio et al.  \cite{bengio2007greedy} proposed a greedy layerwise unsupervised learning algorithm by initializing weights for each layer in a
region near a good local minimum, giving rise to internal distributed representations
that are high-level abstractions of the input, thereby bringing better generalization. Nguyen et al. \cite{nguyen2021analytic} developed an analytic layerwise deep learning framework where an inverse layerwise training and a forward progressive learning
approach are adopted.

Alternately, researchers have also considered growing the width gradually by adding neurons for a fixed depth neural network \cite{wu2019splitting, wynne1993node}. Wynne-Jones \cite{wynne1993node}  considered splitting the neurons (adding neurons) based on a principal component analysis on the oscillating weight vector. Liu et al. \cite{wu2019splitting} developed a simple criterion for
deciding the best subset of neurons to split and a splitting gradient for optimally updating the off-springs. Chen et al. \cite{chen2015net2net} showed that replacing a model with an equivalent model that is wider (has more neurons in
each hidden layer) allows the equivalent model to inherit the knowledge from the existing one and can be trained to further improve the performance.

It is noteworthy that while the works cited above show promising results for image classification tasks, layerwise training procedure can introduce a large number of parameters  (especially in the context of a fully connected network) due to the addition of a new layer \cite{hettinger2017forward}. In such cases, it is important to recognize non-important parameters in an added layer and remove them. Further, layerwise training strategies in \cite{belilovsky2019greedy,hettinger2017forward,trinh2019greedy} do not ensure that the network output remains unchanged after a new layer is added thereby losing the performance achieved by previous layers.  Moreover, recent work by Trinh et al. \cite{trinh2019greedy} observed overfitting in the early layers while performing a layerwise training strategy.  Therefore, it is imperative to consider a layerwise training strategy that takes into account the following aspects: a) achieves light-weight hidden layers, i.e., automatically detects the non-important connections in an added layer and removes them; b) ensures that the network output remains unchanged after the addition of a new layer; c) mathematical principle to guide the design of each layer that prevents overfitting in early layers.

 %It is well known that each layer of the neural network learns a particular feature and that subsequent layers build their
%representation on top of the features produced by a previous layer. Moreover, recent work by Trinh et al. \cite{trinh2019greedy} showed that layerwise training quickly saturates due to the overfitting of early layers within the networks. Therefore, it is imperative to consider a  layerwise training strategy for efficient feature extraction at each layer while also preventing overfitting at early layers. 

\subsection{Our contributions}
 In this work, we use {\it{``robustness"}} as a desirable property for DNNs \cite{jin2020manifold, bubeck2021universal, yang2020closer} and use this as a criterion to devise a strategy for progressively adapting DNN architecture for a given data-set. While recent work shows that over-parametrization could be necessary for {\it{``robustness"}} \cite{bubeck2021universal,bubeck2021law}, it is generally unclear how to distribute the parameters in a network (number of active connections in each hidden layer, number of layers).  Our algorithm progressively builds a network by training one layer at a time while promoting {\it{stability}} for each layer and consequently promoting {\it{``robustness"}}  as we will show.
In particular, we set forth a sequential learning strategy where a learning task is divided into sub-tasks each of which is learned before the next. Our idea is inspired by the mechanism of how the human brain learns \cite{clegg1998sequence}. A layerwise training strategy (\cref{AlgoGreedyLayerwiseResNet}) where each layer is constrained to learn a specific task is adopted. To that end, when training each layer we incorporate a sparsity regularization for identifying the active/inactive parameters in each layer, a manifold regularization term \cite{belkin2006manifold} for promoting stability, and a physics-informed term \cite{raissi2019physics} aiming to respect the underlying physics (if any). The key to our approach is to stimulate
$\delta-$stable neural hidden layers. Once layerwise training saturates, which we shall prove,  we provide  \cref{Sequential} where a sequence of small networks is deployed to extract information from the residuals induced by the DNN generated from  \cref{AlgoGreedyLayerwiseResNet}.  Using  \cref{Sequential} together with \cref{AlgoGreedyLayerwiseResNet} results in robustness and accuracy as we shall show. Since the procedure emphasizes training a small network/layer each time, it is computationally tractable for large-scale problems and is free from vanishing gradient problems. Numerical investigation on prototype regression problems, physics-informed neural networks (PINNs) for solving elliptic partial differential equations, and MNIST classification problems suggest that the proposed approach can outperform ad-hoc baseline networks.

\section{Proposed methodology}

The proposed architecture adaptation procedure has two stages: a) A layerwise training strategy for growing the depth of DNNs; and b) a sequential residual learning strategy in which a sequence of shallow networks is added for robust and accurate predictions. A brief outline of the two stages is provided below in \cref{reguls} and \cref{seq_des}.

\subsection{Layerwise training strategy (\cref{AlgoGreedyLayerwiseResNet})}
\label{reguls}
    One of the key aspects in our layerwise training strategy is to employ the Residual Neural Network (ResNet)  \cite{he2016deep,he2016identity} which,  as we will show,
    %preserves the output value while 
    improves the training loss accuracy when
    adding new layers. In this work, all the matrices and vectors are represented in boldface. Notation $\LRp{.}^{(i)}$ denotes the quantities  (matrix, vector, scalar, set, or function) for the $i^{th}$ layer. %\tanbui{The notations for the $i^{th}$ matrix and the $i^{th}$ function is so similar that could lead to confusion. I read a bit further down and cannot find a reason why we need two different, but almost identical, to talk about things for the $i$-layer?}\krish{fixed the notation, consistently used $\LRp{.}^{(i)}$ } 
    Consider a regression/classification problem of $O$  outputs and $S$ input training features. Given inputs $\bx_i \in \mathbb{R}^S$ for $i \in \{1,2,...M\}$ organized
column-wise into a matrix $\bX \in \mathbb{R}^{S\times M}$, we denote the corresponding network outputs as $\by_i \in \mathbb{R}^O$ which can also be organized column-wise into a matrix $\bY \in \mathbb{R}^{O\times M}$. The corresponding true labels are denoted as $\bc_i\in \mathbb{R}^O$ and in stacked column-wise as $\bC \in \mathbb{R}^{O\times M}$.
%We explain a ResNet propagation  the layerwise training strategy. For the present approach, we consider an affine mapping that upsamples/ downsamples our inputs $\bX$.  
We denote $\bY^{(1)} = h^{(1)}\LRp{\bW^{(1)} \bX + \bb^{(1)}}$ as the output of the first (optional) upsampling/downsampling layer, 
where $\bW^{(1)}$ and $\bb^{(1)}$ represents the weight matrix and bias vector for this layer.
$\bY^{(1)}$ is then propagated forward through a $L$ layers of ResNet as:  
\begin{equation}
\begin{aligned}
&\bY^{(1)} = h^{(1)}\LRp{\bW^{(1)}\bX + \bb^{(1)}},\\
   & \bY^{(i+1)} = \sR^{(i)}\LRp{\bY^{(i)}} + h^{(i+1)}\LRp{\bW^{(i+1)}\sR^{(i)}\LRp{\bY^{(i)}} + {\bf b}^{(i+1)}}, \textrm{ for } i = 1,\dots,L-1,\\
    & \bY^{(L+1)} =h_{pred}\LRp{\bW_{pred}\sR^{(l)}\LRp{\bY^{(L)}}+\bb_{pred}},
    \end{aligned}
    \label{Res_two}
\end{equation}
where $h^{(i+1)}, \bW^{(i+1)}, {\bf b}^{(i+1)}$ are the activation function, weight matrix, and bias vector of the $(l+1)$th layer, while  $h_{pred}, W_{pred},\bb_{pred}$ are the activation function, weight matrix, and bias vector of the output layer, and $\sR^{(i)}$  represents an optional linear transformation for feature scaling on each hidden layer (we only need this for theoretical purpose  in \cref{stability_prop}). 
In order to impose desirable properties for the layerwise training strategy, we incorporate there different regularization terms, namely,  (a) Manifold regularization,  (b) Physics-informed regularization, and (c) Sparsity regularization.
 \begin{enumerate}
    \item \textit{Manifold Regularization (Data-dependent regularization):}
Manifold regularization exploits the geometry of marginal distribution where we assume that the input data is not drawn uniformly from input space $\real^S$ but lies on a submanifold $\sM\subset \real^S$ \cite{belkin2006manifold, fefferman2016testing}. Let $\mu$ be a probability measure with support on $\sM$. %\tanbuis{Let $\nabla_{\sM}f(x)$ denote the gradient of $f$ along the manifold $\sM$}{What is the definition of this manifold gradient?} 
Let $\nabla_{\sM}f(x)$ denote the gradient of $f$ along the manifold $\sM$ (see, e.g.,  Do Carmo \cite{do1992riemannian} for an introduction to differential geometry and definition of gradient along the manifold). To force the gradient of the learned function (each hidden layer in our case) to be small whenever the probability of drawing a sample is large, one can define the following  $H^1$-like regularization term \cite{belkin2006manifold,jin2020manifold,samad2021manifold,lee2015manifold}:
\begin{equation}
   \Phi_m=\int_{\sM}\norm{\nabla_{\sM}\ \by^{(l)}_{\boldsymbol{\theta}}(\bx)}^2_2\ d\mu (\bx) 
   %\xrightarrow[\text{}]{\text{approximated}}\frac{1}
   \approx \frac{1}{M^2}\sum_{i,j}\Bbeta_{ij}\norm{\by^{(l)}_{\boldsymbol{\theta}}(\bx_i)-\by^{(l)}_{\boldsymbol{\theta}}(\bx_j)}^2_2,
\label{Manifold}
\end{equation}
where $\by^{(l)}_{\boldsymbol{\theta}}(\bx)$ is the output of the $l^{th}$ hidden layer for an input $\bx$,
%\tanbuis{where $\by^{(l)}_{\boldsymbol{\theta}}(\bx)$ is the output of the $l^{th}$ hidden layer for an input $\bx$,}{what is the difference between $\by^{(l)}(\bx)$ and $\bY^{(l)}$? Very confusing for me.} \krish{fixed!} 
${\boldsymbol{\theta}}$ represents the trainable network parameters up to layer $l$, and the indices $i$  and $j$ on the %\tanbuis{right hand side}{Have we defined what right hand side is?} 
right hand side of \eqref{Manifold} varies from $1$ to $M$ to represent the pairwise comparison between all training samples. Note that the convergence of the right-hand side to the left-hand side of \eqref{Manifold} is valid under certain conditions and when choosing exponential weights for the similarity matrix  $\Bbeta_{ij}$ \cite{belkin2006manifold,belkin2008towards}. However, \eqref{Manifold} is expensive to compute when $M$ is large due to the 
computational complexity of  $\mathcal{O}(M^2)$. In order to reduce the computational complexity,
 the similarity matrix $\Bbeta_{ij}$ in \eqref{Manifold} is computed based on the pairwise must-link constraint set \cite{ma2018recent}: %\tanbui{Do we still have convergence in this case? Also, the convergence is in which sense?}:
\begin{equation}
    \Bbeta_{ij}=
\begin{cases}
\beta_p & \text{for }  \bc_i^{(p)}=\bc_j^{(p)},\\
0 & \text{otherwise},  
\end{cases}
\label{manifold_strength}
\end{equation}
where $\bc_i^{(p)}$ denotes the label assigned to training input $\bx_i$ belonging to cluster/class $C_p$ for a classification task, and 
 $\beta_p\geq 0$ and bounded such that $\beta_p\neq 0$ for at least one $p$. For regression problems, one may use 
clustering algorithms such as K-means clustering for computing $\Bbeta_{ij}$ in \eqref{manifold_strength}, where data points in the same clusters are assigned the same labels, and different clusters have different labels. %If one additionally assumes that each cluster $C_p$ contains the same number of data points, 
 Inspired by this choice of similarity matrix $\Bbeta_{ij}$ in practice, we consider a new form of manifold regularization $\Phi_m$   as follows:
 \begin{equation}
 \begin{aligned}
\Phi_m&=\sum_{p=1}^K\beta_p\int_{\sM_p}\int_{\sM_p}\norm{ \by^{(l)}_{\boldsymbol{\theta}}(\bx)-\by^{(l)}_{\boldsymbol{\theta}}(\by)}^2_2\ d\mu_p(\bx)\ d\mu_p(\by)\\
&\approx \sum_{p=1}^K\LRs{ \frac{\beta_p}{m_p^2}\sum_{\bx_i,\bx_j\in C_p}  \norm{\by^{(l)}_{\boldsymbol{\theta}}(\bx_i)-\by^{(l)}_{\boldsymbol{\theta}}(\bx_j)}^2_2},
 \end{aligned}
 \label{manif_modif}
 \end{equation}
where $\mu_p$ is the probability measure with support $\sM_p$ and $\cup_p \sM_p = \sM$, $K$ denotes the total number of clusters, and $m_p$ denotes the number of samples in  $C_p$.
%For our analysis in \cref{sect:analysis}, we shall use the manifold regularization in \cref{manif_modif}. 
   In \cref{stability_prop}, we will show that penalizing \eqref{manif_modif} helps in promoting stability for hidden layers during the learning process.
\begin{algorithm}[h!]
	\caption{Layerwise training Algorithm}
	\hspace*{\algorithmicindent} \textbf{Input}: Training data $\bX$, labels $\bC$, validation data $\bX_1$,  validation labels $\bC_1$, loss function $\Phi_d$, loss tolerance  $\epsilon_{\eta}$ for 
	%\hspace*{\algorithmicindent} \textbf{\hspace{1 cm}} 
	addition  
		of layers, node tolerance $\rho$ for thresholding of nodes, initial regularization parameters ($\alpha$, $\tau$, $\gamma$), similarity matrix $\Bbeta_{ij}$,
		%\hspace*{\algorithmicindent} \textbf{\hspace{1 cm}}
		%learning rate $\ell$, the number of epochs $E_e$,
	%learning rate decay $d_l$, 
 the number of neurons in each hidden
	%\hspace*{\algorithmicindent} \textbf{\hspace{1 cm}}
	layer $N_o$.\\
	\hspace*{\algorithmicindent} \textbf{Initialize}:  $\bf{U}, \bb^{(1)}, {\bf{W}}^{(2)}, {\bf b}^{(2)}, \bf{W}_{\mathrm{pred}}, {\bf b}_{\mathrm{pred}}$\\ 
	\begin{algorithmic}[1] 
		\State set $i = 2$, $\alpha^{(2)}=\alpha,\ \tau^{(2)}=\tau,\ \gamma^{(2)}=\gamma$.
		\State Minimize the loss function \eqref{total_loss_first}: %using Adam optimizer with parameters ($\ell$, $d_l$) and  $E_e$ epochs:
		\begin{equation}
		\begin{aligned}
		  \min_{\begin{array}{c}\scriptstyle \bW^{(1)}, \bb^{(1)}; \\[-4pt]
\scriptstyle \bW^{(2)},  {\bf b}^{(2)}; \\[-4pt]    
\scriptstyle  W_{\mathrm{pred}}, {\bf b}_{\mathrm{pred}}
\end{array}} &\Phi_d\LRp{\bC, h_{\mathrm{pred}}\LRp{\bW_{\mathrm{pred}}\sR^{(2)}\LRp{\bY^{(2)}} + {\bf b}_{\mathrm{pred}}}} + \alpha^{(2)} \ \Phi_s\LRp{\bW^{(1)}, \bb^{(1)}, \bW^{(2)}, {\bf b}^{(2)}} \\ 
		& +\   \tau^{(2)} \ \Phi_p\LRp{h_{\mathrm{pred}}\LRp{\bW_{\mathrm{pred}}\sR^{(2)}\LRp{\bY^{(2)}} + {\bf b}_{\mathrm{pred}}}}+\gamma^{(2)} \ \Phi_m\LRp{\bY^{(2)}},\\
		\mathrm{subject}\: \mathrm{to}\: &\bY^{(2)} = \sR^{(1)}\LRp{\bY^{(1)}} + h^{(2)}\LRp{\bW^{(2)} \sR^{(1)}\LRp{\bY^{(1)}}+ {\bf b}^{(2)}},\\
		&\bY^{(1)} = h^{(1)}(\bW^{(1)}\bX + \bb^{(1)}).
		\end{aligned}
		\label{total_loss_first}
		\end{equation}
		\State \textbf{Define}: $\eta^{(2)}= \text{data loss after minimizing (\ref{total_loss_first})}, \ \epsilon_v^{(2)}=\text{val. loss},\ \mathrm{set}\  \eta^{(1)}\approx 0,\ \epsilon_v^{(1)} > \epsilon_v^{(2)}$.
		\While{ $\Big | \frac{\eta^{(i-1)}-\eta^{(i)}}{\eta^{(i-1)}} \Big |>\epsilon_{\eta}$ \textbf{and} $\LRs{\epsilon_v^{(i)}< \epsilon_v^{(i-1)}}$} 
		\State $i = i+1$
        \State Update $(\alpha^{(i)},\ \tau^{(i)},\ \gamma^{(i)})$ based on the previous values $(\alpha^{(i-1)},\ \tau^{(i-1)},\ \gamma^{(i-1)})$.
		\State Extend the neural network by one layer with weights $\bW^{(i)}$ and ${\bf b}^{(i)}$ initialized as zero.
		\State Initialize $ \bW_{\mathrm{pred}}$ and ${\bf b}_{\mathrm{pred}}$ with values inherited from the previously trained network.
		\State Freeze the weights $\bW^{(1)}, \bb^{(1)}, \{\bW^{(l)}\}_{l=1}^{i-1}, \{{\bf b}^{(l)}\}_{l=1}^{i-1}$.
		\State Minimize the loss function \eqref{loss_total}: %objective using Adam optimizer with parameters ($\ell$, $d_l$, $E_e$):
  \begin{equation}
		\begin{aligned}
		  \min_{\begin{array}{c}\scriptstyle  \bf{W}^{(i)}, {\bf b}^{(i)}; \\[-4pt]
\scriptstyle   W_{\mathrm{pred}}, {\bf b}_{\mathrm{pred}}
\end{array}} &\Phi_d\LRp{\bC, h_{\mathrm{pred}}\LRp{\bW_{\mathrm{pred}}\sR^{(i)}\LRp{\bY^{(i)}} + {\bf b}_{\mathrm{pred}}}} + \alpha^{(i)} \times \Phi_s\LRp{ \bW^{(i)}, {\bf b}^{(i)}} \\ 
		+\ \tau^{(i)}& \times \Phi_p\LRp{h_{\mathrm{pred}}\LRp{\bW_{\mathrm{pred}}\sR^{(i)}\LRp{\bY^{(i)}} + {\bf b}_{\mathrm{pred}}}}+\gamma^{(i)} \times \Phi_m\LRp{\bY^{(i)}},\\
		\mathrm{subject}\: \mathrm{to}\: &\bY^{(l+1)} = \sR^{(l)}\LRp{\bY^{(l)}} + h^{(l+1)}\LRp{\bW^{(l+1)}\sR^{(l)}\LRp{\bY^{(l)}} + {\bf b}^{(l+1)}},\ \ l=1,\dots i-1\\
		&\bY^{(1)} = h^{(1)}(\bW^{(1)}\bX + \bb^{(1)}).
		\end{aligned}
  \label{loss_total}
		\end{equation}
		\State Threshold  weights, biases: if  $\Big |w_{jk}\Big | < \rho$ then set $w_{jk} = 0$; if  $ \ \Big |b_{j}\Big | < \rho$ then set $ \ b_{j} = 0$. 
  \State Store the training loss $\eta^{(i)}$ and corresponding best validation loss $\epsilon_v^{(i)}$.
  \EndWhile 
	\end{algorithmic}
 \label{AlgoGreedyLayerwiseResNet}
\end{algorithm}

\item \textit{Physics-informed regularization:}  In addition to the training data, if one is also provided with information on the underlying physics of the problem, then each hidden layer in our proposed approach can be interpretable (see section \ref{PIANN}). For example, given a forward operator $\sG$ of algebraic/ differential equation type:
\begin{equation}
    \sG(\hat{\bx},\ \hat{\by})=\bf{0}, \ \  \ \ \hat{\bx} \in \real^S, \ \ \hat{\by} \in \real^O,
\end{equation}
a physics-informed regularization term can be defined as
\begin{equation}
    \Phi_p=  f\LRs{\sG(\hat{\bx}_1,\ \sN(\hat{\bx}_1, \ {\boldsymbol{\theta}})),\ \sG(\hat{\bx}_2,\ \sN(\hat{\bx}_2, \ {\boldsymbol{\theta}})), \dots \ \sG(\hat{\bx}_r,\ \sN(\hat{\bx}_r, \ {\boldsymbol{\theta}}))}\in \real,
    \label{physics}
\end{equation}
where $\sN$ represents the output of the ResNet in  \eqref{Res_two} with parameters ${\boldsymbol{\theta}}$, $\{ \hat{\bx}_1\dots \hat{\bx}_r\} $ is the set of collocation points, and $f$ is a user-defined loss. 

 \item In addition,  we also employ the $L_1$ regularization (sparsity promoting regularization) denoted as $\Phi_s({\boldsymbol{\theta}})$ to promote learning only the important weights/biases in a layer.    
\end{enumerate}
Putting all the regularizations together, our layerwise training process is shown in \cref{AlgoGreedyLayerwiseResNet}. \cref{AlgoGreedyLayerwiseResNet} starts with training (minimizing the loss function \eqref{total_loss_first})  a two-hidden layer network where the first layer is the (optional) upsampling/downsampling layer. Once, this network is trained, a new layer is added with the weights and biases initialized as zero (line 6 in  \cref{AlgoGreedyLayerwiseResNet}). Keeping parameters in the previous layers fixed, the newly added layer is trained  (minimizing the cost function  \eqref{loss_total}). The procedure is repeated until a termination criteria is satisfied (line 4 in  \cref{AlgoGreedyLayerwiseResNet}).
%Regularization parameters ($\alpha^{(i)}$, $\tau^{(i)}$, $\gamma^{(i)}$) is employed while training the $i^{th}$ hidden layers. 
Other details are provided in \cref{AlgoGreedyLayerwiseResNet}.

\subsection{Sequential residual learning strategy (\cref{Sequential})}
\label{seq_des}
As we shall show (see \cref{TPOP_cor}), even though \cref{AlgoGreedyLayerwiseResNet} proposed in this work prevents overfitting in early layers, training saturates after a certain critical layer.  To overcome this issue, we add a post-processing stage for \cref{AlgoGreedyLayerwiseResNet} in order to further decrease the training loss without overfitting the data. 
\begin{algorithm}[h!]
	\caption{Sequential residual learning}
	\hspace*{\algorithmicindent} \textbf{Input}: Training data $\bX$, labels $\bC$, validation data $\bX_1$, validation labels $\bC_1$, trained network $\sN$ 
 from \cref{AlgoGreedyLayerwiseResNet} and the corresponding validation loss $\epsilon_v^{(1)}$, loss function $\Phi_d$, training epoch $E_e$.\\
	\hspace*{\algorithmicindent} \textbf{Initialize}:  Maximum number of networks: $N_n$, initialize each network: $\sQ_r$ with $\leq 2$ hidden 
layers.\\
	\begin{algorithmic}[1] 
		\State set $r =1$, set $\epsilon_v^{(0)}>\epsilon_v^{(1)}, \sQ_1=\sN$
		%\While{$i \le N_n$ \textbf{and} $\Big |\frac{\eta^{(i-1)}-\eta^{i}}{\eta^{(i-1)}}\Big |>\epsilon_{e}$ \textbf{and} $\LRs{(\epsilon_v)^{i}\leq (\epsilon_v)^{i-1}}$} 
            \While{$r \le N_n$ \textbf{and} $\LRs{\epsilon_v^{(r)}< \epsilon_v^{(r-1)}}$} 
		\State Load trained network $\sQ_r$
		\State $\bC=\bC-\sQ_r(\bX) $
        \State Initialize network $\sQ_{r+1}$ with output layer parameters as $0$.  
		\State Train network $\sQ_{r+1}$ for $E_e$ epochs with data $\bX$, labels $\bC$ and loss function $\Phi_d$. 
		%\State Restore the best network $\sQ_{i+1}$, training loss $\eta^{(i+1)}$ and validation loss $(\epsilon_v)^{(i+1)}$
            \State Restore the best network $\sQ_{r+1}$,  and validation loss $\epsilon_v^{(r+1)}$.
		\State $r = r+1$
		\EndWhile
		\State Compute the net output as: $\bY=\sN(\bX)+\sQ(\bX),\quad \mathrm{where}\ \sQ(\bX)=\sQ_2(\bX)+\dots \sQ_{r-1}(\bX)$.
	\end{algorithmic} \label{Sequential}
\end{algorithm}
This stage is named the ``sequential residual learning strategy" where one trains a sequence of neural networks to extract information from the residual resulting from \cref{AlgoGreedyLayerwiseResNet}.  The procedure is shown in \cref{Sequential}. 
\cref{Sequential} starts by generating new training labels (residuals) using a pre-trained network $\sN$ as demonstrated in  line 4 of \cref{Sequential}. A shallow network $\sQ_2$ is then trained for a limited number of epochs as shown in line 6 of \cref{Sequential}. Note that this is essential for preventing overfitting on the residuals and promoting $\delta-$stability as will be discussed later in  \cref{math_sequential}.   The procedure is then repeated until a termination criteria is satisfied (line 2 in \cref{Sequential}).
\section{An analysis of  \cref{AlgoGreedyLayerwiseResNet} and  \cref{Sequential}}
\label{sect:analysis}
In this section, we introduce key concepts\textemdash $\delta-$stability and approximate $\delta-$robustness\textemdash that drive the development of our approach. In particular, we prove how the design of \cref{AlgoGreedyLayerwiseResNet} promotes $\delta-$stability for each hidden layer (\cref{relevmanif}) while also analyzing training saturation problems that might arise from certain hyperparameter settings (\cref{train_sat_alg}). Lastly, \cref{robust_seq} analyzes the role of \cref{Sequential} in promoting robustness.

\subsection{Layerwise training  \cref{AlgoGreedyLayerwiseResNet}}
\begin{figure}[h!]
    \hspace{-1.0 cm}
     \includegraphics[scale=0.55]{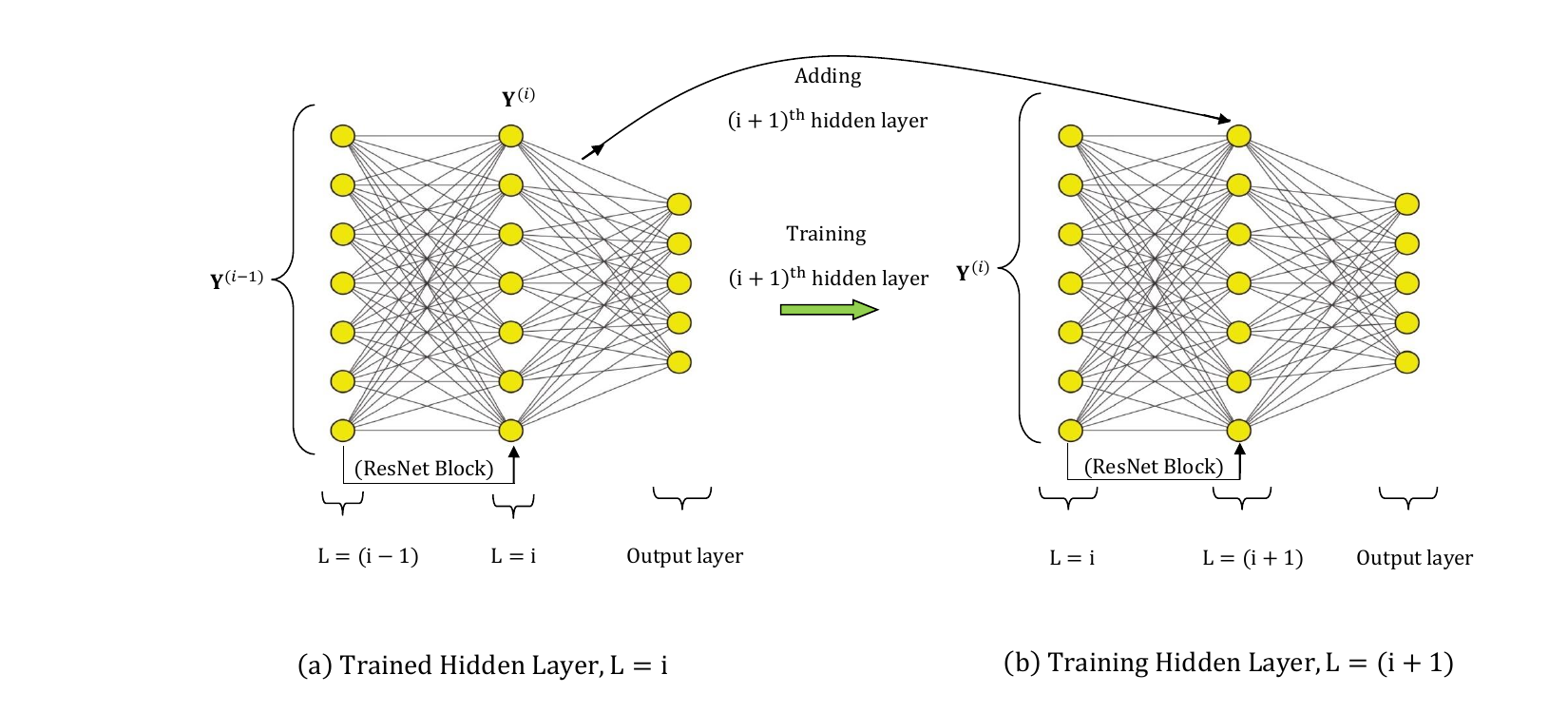}
    \caption{Schematic of  \cref{AlgoGreedyLayerwiseResNet}: Training the $(i+1)^{th}$ hidden layer. }
    \label{Algo_I_scheme}
\end{figure}

In  \cref{AlgoGreedyLayerwiseResNet}, the input features are transformed through layers with previously trained weights and biases to create a new learning problem for each  additional layer. Note that each layer of a deep neural network can be regarded as an ``information
filter" \cite{lengelle1996training} transferring information from a previous layer to next layer. In particular, it is interesting to look at how 
 regularizations defined in section \ref{reguls} helps the layerwise training strategy. Further, it is also important to know the conditions under which each additional added layer is trainable. To that end, let us define the following:

\begin{definition}[Input space and neural transfer map]
\label{ntmm}
Let the input data lie on $\sM^{(0)}\subset \real^S$  such that $\sM^{(0)}=\bigcup \limits_{j=1}^{K}\sM_j^{(0)}$, where each  $\sM_j^{(0)}$ contains similar data points based on some  predefined similarity measure. Further, let $C_j\subset \sM_j^{(0)}$ be the associated training data clusters.
We  define the neural transfer map $\sN^{(i)}$ with domain $\sM^{(i-1)}$ as:
\begin{equation}
    \begin{aligned}
     \sN^{(i)}\LRp{\bx}&= h^{(i)}\LRp{ \bW^{(i)*}\bx + \bb^{(i)*}},\  \mathrm{for} \ \ i=1\ \ \mathrm{and},\\
     \sN^{(i)}\LRp{\bx}&= \bx+ h^{(i)}\LRp{ \bW^{(i)*}\bx + \bb^{(i)*}},\  \mathrm{for} \ \ i=2,3,\dots L
\end{aligned}
\label{ntm}
\end{equation}
%\begin{equation}
  %  \sN^{(i)}\LRp{\bY^{(i-1)}}= h^{(i)}\LRp{\bY^{(i-1)} \bW^{(i)*} + \bb^{(i)*}},\  \mathrm{for} \ \ i=1,
%\end{equation}
%\begin{equation}
 %   \sN^{(i)}\LRp{\bY^{(i-1)}}= \bY^{(i-1)}+ h^{(i)}\LRp{\bY^{(i-1)} \bW^{(i)*} + \bb^{(i)*}},\  \mathrm{for} \ \ i=2,3,\dots
%\end{equation}
where  $\bW^{(i)*}$ and $\bb^{(i)*}$ denote the trained biases and weights of the $i^{th}$ layer, $N_o$ represents the number of neurons in each hidden layer, and the domain $\sM^{(i)}$ for each subsequent neural transfer map $\sN^{(i+1)}$  is obtained  as:
\begin{equation}
   \sM^{(i)}= \bigcup \limits_{j=1}^{K}\sM_j^{(i)}, \ \mathrm{where},\ \sM_j^{(i)}=\sR^{(i)} \LRp{\sN^{(i)}\LRp{\bM^{(i-1)}_j}}, \  \mathrm{for} \ \ i=1,2,\dots L-1
   \label{rescale}
\end{equation}
% \tanbuis{}{Have we defined what $\sM^{(0)}_j$, and hence $\sM^{(0)}$, are?} \krish{rectified! Introduced now in the beginning of the definition.}
\end{definition}
A key aspect of the algorithm is to learn transfer maps $\sN^{(i)}$  that allows for effective information  transfer through the network.  In this work, we focus on creating transfer maps $\sN^{(i)}$ that are $\delta-$ stable for a given input $\bx$ through the use of manifold regularization \cite{jin2020manifold, bubeck2021universal, yang2020closer}. A schematic of  \cref{AlgoGreedyLayerwiseResNet} is given in Figure \ref{Algo_I_scheme}.

\subsubsection{$\delta-$stability in layerwise training  \cref{AlgoGreedyLayerwiseResNet}}
\label{relevmanif}

In this section, we investigate the relevance of manifold regularization in our framework. The main motivation of manifold regularization is to promote $\delta-$stability. Here, stability  %is the property that 
means if two data points are “similar” to each other in some sense, then the network predictions on the two data points must be close to each other. Thus, stability in this sense is closely related to continuity. We now provide the details.  %We define the notion of $\delta$-stability for a learning algorithm as follows.
%as a property of the learning algorithm follows:
\begin{comment}
\begin{definition}{$\delta-$stability}{}
\label{delta-def}

\krish{We call a parameterized function $f_{\boldsymbol{\theta}}: \sX \subset \real^T \rightarrow \sY$ to have the property of $\delta-$stability on $\sX$ when it's stability is explicitly controlled through a hyper-parameter $\zeta>0$ of the learning algorithm that learns $\boldsymbol{\theta}$, i.e there exists a stability function 
$\delta(\zeta,\ \bx,\ \epsilon): [0,\ \zeta^u]\times \sX \times \real \rightarrow \real$, such that:
\begin{enumerate}
\item At any given $\bx \in \sX$,  $\forall \bx' \in B_\epsilon (\bx)  \cap \sX$ one has:
    \begin{equation}
 \norm{f_{\boldsymbol{\theta}}\LRp{\bx}-f_{\boldsymbol{\theta}}\LRp{\bx'}}_2 \leq {\delta}(\zeta,\ \bx,\ \epsilon), 
    \label{stability}
\end{equation}
where, $ B_\epsilon (\bx)$ denotes the open ball centered at $\bx$  with radius $\epsilon$ in the metric space $(\real^{T},\ \ell^2)$. (Note that the dependence of $\boldsymbol{\theta}$ on $\zeta$ is omitted for brevity).
    \item $\delta(\zeta^u,\ \bx,\ \epsilon)<\delta(0,\ \bx,\ \epsilon)$, 
     \label{qw_1}
     \item $\delta(\zeta,\ .,\ .)$ is continuous with respect to $\zeta$.
  %  \item Let $c \in \LRp{\delta(\zeta^u,\ \bx,\ \epsilon),\delta(0,\ \bx,\ \epsilon)}$, then $\exists \zeta\in \LRp{0,\zeta^u}$ such that $\delta(\zeta,\ \bx,\ \epsilon)=c$, i.e any desired stability can be achieved by varying the parameter $\zeta$.
        \label{qw}
\end{enumerate} 
}
\end{definition}
\end{comment}
\begin{definition}[$\epsilon-\delta$ stability promoting algorithm]
\label{delta-def}
Let $\sX=\bigcup \limits_{j=1}^{K}\sX_j$ denote the input space, where each  $\sX_j$ contains similar data points based on some assumed similarity measure. Consider a parameterized function $f_{\boldsymbol{\theta}}: \sX \subset \real^T \rightarrow \sY$ and a learning algorithm to find an ``optimal" parameter $\boldsymbol{\theta}^*$. Let $\zeta$ be a stability parameter characteristic of the learning algorithm and let $\epsilon$ be a given constant.  We call a learning algorithm to be $\epsilon-\delta$ stability promoting algorithm for $f_{\boldsymbol{\theta}}$ if there exists a stability function 
$\delta_j(\zeta): [\zeta^l,\ \zeta^u]\rightarrow \real$, such that:
\begin{enumerate}
\item $\delta_j(\zeta^u)=\epsilon, \ \delta_j(\zeta^u)< \delta_j(\zeta^l)$, 
     \label{qw_1}
     \item $\delta_j(\zeta)$ is continuous with respect to $\zeta$,
     \label{qw}
\item  For each cluster $ \sX_j$, there exists $\boldsymbol{\theta}^*=\boldsymbol{\theta}^*(\zeta)$ such that:
%\tanbuis{there exists $\zeta := \zeta\LRp{\boldsymbol{\theta}^*}$ such that}{Check: i) if my re-writing is correct and ii) Does $\zeta$ depend on $\bx$?}\krish{i) The correct way should be there exists $\boldsymbol{\theta}^*=\boldsymbol{\theta}^*(\zeta)$. For every pick of parameter $\zeta$, the weights/biases $\boldsymbol{\theta}^*$ changes so that below equation is true and that's how we control stability through $\zeta$!; ii) Also  $\zeta$ does not depend on $\bx$ as we only care about controlling stability in each cluster containing similar data-points, rather than controlling stability for each $\bx$ in the cluster itself, or in other words we only care about global stability for each cluster. I have modified to say $\forall \bx$ in below equation for clarity.}
%$\forall \bx' \in \sX_j$ one has:
    \begin{equation}
\norm{f_{\boldsymbol{\theta}^*}\LRp{\bx}-f_{\boldsymbol{\theta}^*}\LRp{\bx'}}_2 \leq {\delta}_j(\zeta), \quad \forall \bx,\ \bx' \in \sX_j.
    \label{stability}
\end{equation}
\label{qw_22}
%where, the dependence of $\boldsymbol{\theta}$ on $\zeta$ is omitted for brevity. 
  %   \item The learning algorithm uses a criteria to pick the best $\zeta^*\in \LRp{0,\ \zeta^u}$  that promotes stability.
  %\label{qw_5}
  %  \item Let $c \in \LRp{\delta(\zeta^u,\ \bx,\ \epsilon),\delta(0,\ \bx,\ \epsilon)}$, then $\exists \zeta\in \LRp{0,\zeta^u}$ such that $\delta(\zeta,\ \bx,\ \epsilon)=c$, i.e any desired stability can be achieved by varying the parameter $\zeta$.
\end{enumerate} 
\end{definition}
\begin{remark}
\label{rem_sts}
 Conditions \ref{qw_1}, and \ref{qw} implies that the algorithm can promote any desired stability $\delta_{desired} \in \LRp{\epsilon,\ \delta_j(\zeta^l)}$  by varying parameter $\zeta$, i.e $\exists \zeta^*\in \LRp{\zeta^l,\zeta^u}$ such that $\delta_j(\zeta^*)=\delta_{desired}$ (Intermediate value theorem). Note that $\epsilon$ is the best stability that can be achieved.
However, if the domain of $\delta_j(\zeta)$ is discrete, i.e,
 the hyperparameter $\zeta$ of the learning algorithm can only take discrete values in $[\zeta^l,\ \zeta^u]$, then we consider a  relaxation of condition \ref{qw} in \cref{delta-def}. To that end we define ``discrete $\epsilon-\delta$ stability promoting algorithm" in \cref{discrete_ep_del}.
 
 %\tanbui{But the third condition does not say so, as $\zeta := \zeta\LRp{\boldsymbol{\theta}}$ and thus you can never achieve $\zeta^*$? }\krish{$\boldsymbol{\theta}=\boldsymbol{\theta}(\zeta)$. So there exists a $\zeta^*$ and a corresponding $\boldsymbol{\theta}^*$ for which $\delta_j(\zeta^*)=\delta_{desired}$ can be achieved.}
%\begin{enumerate}
   % \item Condition \ref{qw_1} and \ref{qw} implies that desired stability $\delta_{desired} \in \LRp{\delta_j(\zeta^u),\delta_j(0)}$ can be achieved by varying the parameter $\zeta$, i.e $\exists \zeta^*\in \LRp{0,\zeta^u}$ such that $\delta_j(\zeta^*)=\delta_{desired}$ (Intermediate value theorem). $\zeta^u$ is typically chosen so as to have a large interval of stability $\LRp{\delta_j(\zeta^u),\delta_j(0)}$.
   % \item In practice,  the hyperparameter $\zeta^*$ in (\ref{stability}) is chosen based on the performance on a validation data-set (exploring the balance between over-stabilization and under-stabilization).
  %  \end{enumerate}
\end{remark}
\begin{definition}[Discrete $\epsilon-\delta$ stability promoting algorithm]
\label{discrete_ep_del}
  Consider \cref{delta-def} and
   assume that the parameter $\zeta$ takes only discrete values, i.e $\zeta \in \LRc{\zeta^l = \zeta_1,\ \zeta_2,\hdots, \zeta_{n-1},\ \zeta_n  =\zeta^u}$ and let the discrete domain of 
 $\delta_j(\zeta)$ be denoted as $D$. Then, we call a learning algorithm to be discrete $\epsilon-\delta$ stability promoting algorithm for $f_{\boldsymbol{\theta}}$ if $\delta_j(\zeta): D \rightarrow \real$ satisfies:
    \begin{enumerate}
    \item $\delta_j(\zeta^u)=\epsilon, \ \delta_j(\zeta^u)< \delta_j(\zeta^l)$, 
     \item $\delta_j(\zeta)$ is monotonically decreasing with respect to $\zeta$,
   \item   For each cluster $ \sX_j$, there exists $\boldsymbol{\theta}^*=\boldsymbol{\theta}^*(\zeta)$ such that:
         \begin{equation}
\norm{f_{\boldsymbol{\theta}^*}\LRp{\bx}-f_{\boldsymbol{\theta}^*}\LRp{\bx'}}_2 \leq {\delta}_j(\zeta), \quad \forall \bx,\ \bx' \in \sX_j.
\end{equation}
     \label{qw_discrete}
%\tanbuis{there exists $\zeta := \zeta\LRp{\boldsymbol{\theta}^*}$ such that}{Check: i) if my re-writing is correct and ii) Does $\zeta$ depend on $\bx$?}\krish{i) The correct way should be there exists $\boldsymbol{\theta}^*=\boldsymbol{\theta}^*(\zeta)$. For every pick of parameter $\zeta$, the weights/biases $\boldsymbol{\theta}^*$ changes so that below equation is true and that's how we control stability through $\zeta$!; ii) Also  $\zeta$ does not depend on $\bx$ as we only care about controlling stability in each cluster containing similar data-points, rather than controlling stability for each $\bx$ in the cluster itself, or in other words we only care about global stability for each cluster. I have modified to say $\forall \bx$ in below equation for clarity.}
%$\forall \bx' \in \sX_j$ one has:
%where, the dependence of $\boldsymbol{\theta}$ on $\zeta$ is omitted for brevity. 
  %   \item The learning algorithm uses a criteria to pick the best $\zeta^*\in \LRp{0,\ \zeta^u}$  that promotes stability.
  %\label{qw_5}
  %  \item Let $c \in \LRp{\delta(\zeta^u,\ \bx,\ \epsilon),\delta(0,\ \bx,\ \epsilon)}$, then $\exists \zeta\in \LRp{0,\zeta^u}$ such that $\delta(\zeta,\ \bx,\ \epsilon)=c$, i.e any desired stability can be achieved by varying the parameter $\zeta$.
\end{enumerate} 
\end{definition}
\begin{remark}
    Conditions \ref{qw_discrete} in \cref{discrete_ep_del} implies that the learning algorithm can continuously promote/improve the stability in a discrete manner by increasing  $\zeta$. However, the notion of discrete stability promoting algorithm in \cref{discrete_ep_del} is weaker than the one in \cref{delta-def} since it promotes only desired stability $\delta_{desired} \in \LRc{\delta_j(\zeta_1),\ \delta_j(\zeta_2),\hdots, \delta_j(\zeta_{n-1}),\ \delta_j(\zeta_n) }$.
    \end{remark}
\begin{definition}[$\delta-$stable function]
\label{delta_deff}
Consider the parameterized function $f_{\boldsymbol{\theta}}$ and an $\epsilon-\delta$ stability promoting algorithm as defined in \cref{delta-def} or \cref{discrete_ep_del}. Let $\zeta^*$ be the optimal stability parameter chosen based on a predefined criteria.  Then, we call the  function $f_{\boldsymbol{\theta}^*}$
to be $\delta_j(\zeta^*)-$stable on $\sX_j$ (or in short $\delta-$stable) if it satisfies \eqref{stability}.
%(\tanbuis{or in short $\delta-$stable}{stable locally at $\bx$ is not the same as stable globally on $\sX_j$.}
%\krish{I have modified the writing and cleared this confusion here. We are only looking at controlling global stability on each cluster $\sX_j$ (containing similar data points) throughout the paper!}).
\end{definition}
\begin{remark}
\label{tunin}
    In this work, the optimal $\zeta^*$ is defined as the one  that minimizes the validation loss. That is, one looks at the performance on a validation data-set (exploring the balance between over-stabilization and under-stabilization) to pick  $\zeta^*$. A numerical demonstration in the context of choosing the best manifold regularization parameter is provided in  \cref{bost_sec}, Figure \ref{Boston_training} (middle figure). Here, one  plots the $\LRp{\zeta,\mathrm{validation \ loss}}$ curve and chooses  $\zeta^*$ that gives the lowest validation loss. More details are provided in \cref{bost_sec}. 
    %\tanbuis{A numerical demonstration is provided in  Figure \ref{Boston_training}, \cref{bost_sec}.}{does this example also shows how to pick $\zeta^*$? if yes, spell it out as it is not clear to me to how pick it at all except the principle for choosing it as you stated.}\krish{Yes, it shows how to pick $\zeta^*$. I have rewritten the sentence above.}
\end{remark}
For learning a $\delta-$ stable function in the sense of \cref{delta_deff}, we first need to create an $\epsilon-\delta$ stability promoting algorithm. To that end, we first determine the conditions under which one has an $\epsilon-\delta$ stability promoting algorithm through the use of manifold regularization.
%(refer to \cref{delta-def}).
For analysis purpose, let us first consider a re-normalization of the loss \eqref{loss_total} as follows:
\begin{equation}
 \sL_r= \LRp{\gamma^u-\gamma^{(i)}} \times  \Phi_d + \LRp{\gamma^u-\gamma^{(i)}}\alpha^{(i)} \times \Phi_s \\ 
		+\LRp{\gamma^u-\gamma^{(i)}}\tau^{(i)} \times \Phi_p +\gamma^{(i)} \times \Phi_m,
  \label{renormalized_loss}
  \end{equation}
  where $\gamma^u$ is a predefined upper bound such  that when $\gamma^{(i)}=\gamma^u$, $\Phi_m$ (manifold regularization) is the sole component in the total loss $\sL_r$, and when $\gamma^{(i)}=0$, the manifold regularization is absent. Now, we are in the position to show the existence of a stability function when employing manifold regularization.
To that end we will first prove the following lemma. %\ref{upper_hemicontinuity}.
\begin{lemma}[Set of minimizers $\boldsymbol{\theta}^*\LRp{\gamma^{(i)}}$ is upper hemicontinuous with respect to $\gamma^{(i)}$]
\label{upper_hemicontinuity}
Consider the loss \eqref{renormalized_loss} associated with training the layer $\sN^{(i)}$. For a given $\{ \alpha^{(i)},\ \tau^{(i)}\}$ we define the loss function \eqref{renormalized_loss} as $\Phi\LRp{\boldsymbol{\theta},\ \gamma^{(i)}}$, where $\boldsymbol{\theta}$ represents the weights and biases to be optimized. Suppose the set of global minimizers, $S\LRp{\gamma^{(i)}} = \LRc{\boldsymbol{\theta}^*: 
\boldsymbol{\theta}^* = \arg\min_{\boldsymbol{\theta}}\Phi\LRp{\boldsymbol{\theta},\ \gamma^{(i)}}}$ corresponding to the loss function \eqref{loss_total} is non-empty and commonly bounded\footnote{Commonly bounded means that $\exists M<\infty$, independent of $\gamma^{(i)}$, such that $\sup_{\gamma^{(i)}\in [0,\ \gamma^u]} \snor{\LRp{\boldsymbol{\theta}^*\LRp{\gamma^{(i)}}}}\leq M$.} for all $\gamma^{(i)}\in [0,\ \gamma^u]$, where $\gamma^u$ is a predefined upper bound. %\footnote{Since $\gamma^{(i)}$ is a regularization parameter, we would like it to be finite, but smaller than an upper bound to avoid over-regularization.} %How to pick  such an upper bound or the best regularization parameter is a well-known problem and is not a subject of interest of this paper.} 
%Let \tanbuis{the set of global minimizers  be denoted as  $\boldsymbol{\theta}^*\LRp{\gamma^{(i)}}$}{too much abuse of notation here: $\boldsymbol{\theta}^*\LRp{\gamma^{(i)}}$ is more suitable for a the minimizer.}. 
Then, 
 %correspondence $\boldsymbol{\theta}^*\LRp{\gamma^{(i)}}$ 
 $S\LRp{\gamma^{(i)}}$ is upper hemicontinuous in % with respect to
 $\gamma^{(i)}$.
 %https://math.stackexchange.com/questions/481952/union-of-infinitely-many-bounded-sets-is-not-bounded
\end{lemma}
\begin{proof}
 %With the assumptions in lemma \ref{upper_hemicontinuity}, let us denote the loss function \eqref{loss_total} as $\Phi\LRp{\boldsymbol{\theta},\ \gamma^{(i)}}$, where $\boldsymbol{\theta}$ represents the weights and biases to be optimized. 
 We shall use the Berge maximum theorem (refer to Chapter 17 in \cite{charalambos2013infinite}) to show that the set of minimizers %$\boldsymbol{\theta}^*\LRp{\gamma^{(i)}}$ 
 $S\LRp{\gamma^{(i)}}$ is upper hemicontinuous with respect to $\gamma^{(i)}$. To that end the following observations can be made:
 \begin{enumerate}
     \item   $\Phi\LRp{\boldsymbol{\theta},\ \gamma^{(i)}}$ is continuous with respect to both $\boldsymbol{\theta}$ and $\gamma^{(i)}$, since each component of the total loss \eqref{renormalized_loss}  is continuous. %\tanbui{you need to provide the reason/justification for each of the observation.}
     \item Let $\real^c$ be the Euclidean space with $c$ being the total number of parameters and $\sP(\real^s)$ be the set of all subsets of $\real^s$. There exists a compact-valued correspondence  $C:[0,\ \gamma^u] \rightarrow \sP(\real^s)$ with $C\LRp{\gamma^{(i)}}\neq \emptyset,\ \forall \gamma^{(i)} \in [0,\ \gamma^u]$ such that:
     \[
        \boldsymbol{\theta}^*\LRp{\gamma^{(i)}}=\arg\min_{\boldsymbol{\theta} \in C\LRp{\gamma^{(i)}}} \Phi\LRp{\boldsymbol{\theta},\ \gamma^{(i)}}= \arg \min_{\boldsymbol{\theta} \in \real^c} \Phi\LRp{\boldsymbol{\theta},\ \gamma^{(i)}}.
      \]
      Indeed, since the set of global minimizers is non-empty $\forall \gamma^{(i)}\in [0,\ \gamma^u]$, the existence of a compact-valued correspondence is straightforward to verify by defining the constraint set to be 
 the same for each $\gamma^{(i)}\in [0,\ \gamma^u]$, i.e  $C(\gamma^{(i)})=\overline{\bigcup_{\gamma^{(j)}\in [0,\ \gamma^u]}S\LRp{\gamma^{(j)}}}$, where $\overline{(.)}$ denotes the closure of the set. Further, $C(\gamma^{(i)})$ is closed an bounded, and hence compact, due to  the  assumption that the set of global minimizers is commonly bounded $\forall \gamma^{(i)}\in [0,\ \gamma^u]$,
    \item Since the compact constraint set $C(\gamma^{(i)})$ defined above is constant independent of $\gamma^{(i)}$, it is continuous (refer to Chapter 17 in \cite{charalambos2013infinite}).
 \end{enumerate}
 Thus, by Berge maximum theorem %(refer to Chapter 17 in \cite{charalambos2013infinite}) 
 the 
 correspondence $\boldsymbol{\theta}^*\LRp{\gamma^{(i)}}$ is upper hemicontinuous with respect to $\gamma^{(i)}$.
\end{proof}
\begin{proposition}[$\epsilon-\delta$ stability promoting algorithm via manifold regularization]
\label{stability_prop}
Consider the input space and neural transfer map introduced in  \cref{ntmm}. Further assume that:
\begin{enumerate}
   % \item The  training input set $\sD_{\bx}$ is supported on some compact subset $\sM^{(0)}\subset \real^S$ such that there exists $K$ data clusters in $\sD_{\bx}$. 
 %  \item Each cluster  contains the same number of data points. %\krish{There is a difference between $ \sX_p$ and $C_p$. $C_p$ is the actual training data clusters containing finite samples formed from training data points and $C_p\subset \sX_p$. $\sX_p$ is the actual input space. $C_p$ is defined in \cref{ntmm} and also in the very beginning near equation 2.3. The notation $\sX_p$ in definition is adopted as a general notation for describing stability. In our case of layerwise training, $\sX_j$  will coincide with $\sM^{(0)}_j$ in definition 3.1.}
  %  \label{gl_1}
    \item   The samples in cluster $C_p$\footnote{Note that we do not assume that labels are available for all the inputs.} are independent draws from  $\mu_p^{(0)}$ with support $\sM_p^{(0)}\subset \sM^{(0)}$ where $\sM_p^{(0)}$ is compact $\forall p$. Let $\mu_p^{(i)}=(\sR^{(i)}\sN^{(i)})_{\#}\mu_p^{(i-1)}$ be the  push forward measure associated with the map (\ref{rescale}) for $i=1,\dots$.
    \label{gl_2}
    \item The linear transformation $\sR^{(i)}$ in (\ref{rescale}) rescales the features to $[-M,\ 0]$, where $M$ is a positive number and the activation $h^{(i)}\in \{\mathrm{ELU},\ \mathrm{ReLU},\ \mathrm{Leaky  ReLU},\ \mathrm{SELU}\}$ for $i=2,\dots L$.  
   % \item Let the input set for the subsequent neural transfer maps $\sN^{(i+1)}(\bx)$ be given as $\sM^{(i)}=\bigcup \limits_{j=1}^{K}\sM_j^{(i)}$, where $\sM_j^{(i)}=\sR\LRp{\sN^{(i)}\LRp{\sM_j^{(i-1)}}}$ with the associated push forward measure denoted as $\mu_j^{(i)}=\sN^{(i)}_{\#}\mu_j^{i-1}$ for $i=2,\dots\ etc$.
   \label{gl_3}
     \item The set $S\LRp{\gamma^{(i)}}$ in lemma \ref{upper_hemicontinuity} is singleton $\forall \gamma^{(i)} \in [0,\ \gamma^u]$. That is, the global minimizer is unique $\forall \gamma^{(i)} \in [0,\ \gamma^u]$ and for any choice of $\beta_p$ in \eqref{manifold_strength}. 
         \label{global_minimizer_assumption}
        \end{enumerate}
Then, employing manifold regularization $\Phi_m$ of the form \eqref{manif_modif} with $\beta_p=1,\ \forall p$  yields a probabilistic $\epsilon-\delta$ stability promoting algorithm for $\sN^{(i+1)}$.
%in the sense of \cref{delta-def}.
%, in the sense of \cref{delta-def}.
In particular,  there exists a stability function $\tilde{\delta}_{p}\LRp{\gamma^{(i)},\ \xi^{(i)}}: [0,\ \gamma^u] \times \LRp{0,\ 1} \rightarrow \real$, 
%\tanbuis{($\gamma^u\rightarrow \infty$)}{What does it mean here: is $\gamma^u\rightarrow \infty$ or $\gamma^u\rightarrow \infty$ as something approaches something? If the former, say that as $\gamma^u$ approaches $\infty$. In any case, this needs clarification here to avoid misunderstanding about the finite value of $\gamma^u$ in Lemma 3.6. Also, the statement is very ambiguous to me. Moreover, previously (even the third assumption above) we have the index $(i)$ for $\gamma$, but we do not have so for (3.4)?} 
that satisfies  conditions \ref{qw_1} and \ref{qw} for every given $\xi^{(i)}$,  such that
%\[\tilde{\delta}_{p}\LRp{\gamma^u}\rightarrow 0, \ \tilde{\delta}_{p}\LRp{\gamma^u}< \tilde{\delta}_{p}\LRp{0}, \quad \mathrm{as\ }\frac{1}{\gamma^u},\ \frac{\alpha^{(i)}}{\gamma^u},\ \frac{\tau^{(i)}}{\gamma^u}\rightarrow 0,\ \mathrm{in}\ \eqref{loss_total}\]
  %  \[\tilde{\delta}_j(\gamma^{(i+1)})= \int_{\sM_j^{(i)}}\int_{\sM_j^{(i)}}\norm{ \sN^{(i+1)}(\bx)-\sN^{(i+1)}(\by)}^2_2\ d\mu_j^{(i)}(\bx)\ d\mu_j^{(i)}(\by),\]
$\forall \bx,\ \bx'$ in $ \sM_p^{(i-1)}$:
\begin{equation}
 \norm{\sN^{(i)}\LRp{\bx}-\sN^{(i)}\LRp{\bx'}}_2^2 \leq \tilde{\delta}_{p}\LRp{\gamma^{(i)},\ \xi^{(i)}}, \quad i=2,\dots, L,
  \label{stability_transfer}\end{equation}
holds  with probability at-least $\LRp{1-\xi^{(i)}-\sE_p^{(i)}\LRp{m_p,\ \xi^{(i)}\epsilon}}$, 
%\tanbui{at nowhere you introduced $\varepsilon$ including in the proof in Appendix A. You need to define what $\epsilon$ is in this theorem and discuss how the right hand side of \eqref{stability_transfer} plays the role of the right hand side of 
%\eqref{stability}.} 
where $\sM_p^{(i-1)}$ is defined in (\ref{rescale}),  $\gamma^{(i)}$ is the 
%\tanbui{you meant the optimal weight after training (as in definition 3.4)? If yes, we need to resolve the notation issue here as you used $^*$ for optimized quantities before.} \krish{No. We do not talk about optimal weight yet. We are trying to prove $\delta-$stability promoting algorithm first. So $\gamma^{(i)}$ is still a variable here. There is a discussion to clarify this aspect in line 235.} 
weight associated with the manifold regularization term in  \eqref{loss_total}, $\sE_p^{(i)}\LRp{m_p,\epsilon}=2m_p \ \exp \LRp{-\omega_p^{(i)}m_p\epsilon^2}+2 \ \exp \LRp{-\vartheta_p^{(i)}m_p\epsilon^2}$, and $\omega_p^{(i)},\ \vartheta_p^{(i)}$ are positive constants that depends only on the cluster $C_p$ and the layer $i$.
\end{proposition}
\begin{proof}
The proof follows directly from analyzing the effect of manifold regularization applied to each neural transfer map. Consider the trained neural transfer map $\sN^{(2)}\LRp{\bx }$ which is the hidden layer of the ResNet architecture after the upsampling/downsampling layer. From \eqref{manif_modif} with $\beta_p=1,\ \forall p$, a sample average approximation of the manifold regularization term $\Phi_m$   can be written as: 
\begin{equation}
   \sum_{p=1}^K\LRs{ \frac{1}{m_p^2}\sum_{\bx_i,\bx_j\in C_p}  \norm{\sN^{(2)}(\bx_i)-\sN^{(2)}(\bx_j)}^2_2},
    \label{manif_pr}
\end{equation}
where, $m_p$ denotes the number of data points in cluster $C_p$, and $C_1,\ C_2\dots C_k$ denotes the data clusters. Now for each cluster $C_p$, one can show the following non-asymptotic bound (proof is provided in \cref{conv_prop}):
\begin{align}
  \label{sample_influence}
   & \p\LRs{\expect_{\mu_p^{(1)}}\LRs{\norm{ \sN^{(2)}(\bx)-\sN^{(2)}(\by)}^2_2}\leq \epsilon+\frac{1}{m_p^2}\sum_{\bx_i,\bx_j\in C_p}\norm{\sN^{(2)}(\bx_i)-\sN^{(2)}(\bx_j)}^2_2}\geq \sE_p^{(2)}\LRp{m_p,\epsilon},\\
    &\expect_{\mu_p^{(1)}}\LRs{\norm{ \sN^{(2)}(\bx)-\sN^{(2)}(\by)}^2_2}= \int_{\sM_p^{(1)}}\int_{\sM_p^{(1)}}\norm{ \sN^{(2)}(\bx)-\sN^{(2)}(\by)}^2_2\ d\mu_p^{(1)}(\bx)\ d\mu_p^{(1)}(\by).
     \label{sub_manifold_b}
\end{align}
%\begin{align}
 % \label{sample_influence}
%   & \p\LRs{\snor{\frac{1}{m^2}\sum_{\bx_i,\bx_j\in C_p}\norm{\sN^{2}(\bx_i)-\sN^{2}(\bx_j)}^2_2-\expect_{\mu_p^{(1)}}\LRs{\norm{ \sN^{(2)}(\bx)-\sN^{(2)}(\by)}^2_2}}\leq \epsilon}\geq 1-\frac{\sigma^2_p}{m^2\epsilon^2},\\
 %   &\expect_{\mu_p^{(1)}}\LRs{\norm{ \sN^{(2)}(\bx)-\sN^{(2)}(\by)}^2_2}= \int_{\sM_p^{(1)}}\int_{\sM_p^{(1)}}\norm{ \sN^{(2)}(\bx)-\sN^{(2)}(\by)}^2_2\ d\mu_p^{(1)}(\bx)\ d\mu_p^{(1)}(\by).
  %  \label{sub_manifold_b}
%\end{align}
where $\sE_p^{(2)}\LRp{m_p,\epsilon}=1-2m_p \ \exp \LRp{-\omega_p^{(2)}m_p\epsilon^2}-2 \ \exp \LRp{-\vartheta_p^{(2)}m_p\epsilon^2}$, $\omega_p^{(2)},\ \vartheta_p^{(2)}$ are positive constants that depends only on the cluster $C_p$ and the hidden layer, and $\expect_{\mu_p}$ is the expectation  defined in \eqref{sub_manifold_b}.
Now, note that the random variable $\norm{\sN^{(2)}(\bx)-\sN^{(2)}(\by)}_2^2$ in (\ref{sub_manifold_b}) is non-negative. Therefore, by Markov inequality \cite{vershynin2018high} we have:
\begin{equation}
    \norm{\sN^{(2)}\LRp{\bx}-\sN^{(2)}\LRp{\bx'}}_2^2 \leq \frac{\expect_{\mu_p^{(1)}}\LRs{\norm{ \sN^{(2)}(\bx)-\sN^{(2)}(\by)}^2_2}}{\xi^{(2)}},
     \label{hoeff}
\end{equation}
holds  $\forall \bx,\ \bx'$ on $ \sM_p^{(1)}$ with probability at-least $(1-\xi^{(2)})$.  Now using the non-asymptotic bound \eqref{sample_influence} in \eqref{hoeff} and applying the union bound, we have:
\begin{equation}
    \norm{\sN^{(2)}\LRp{\bx}-\sN^{(2)}\LRp{\bx'}}_2^2 \leq \frac{1}{\xi^{(2)} \times m_p^2}\sum_{\bx_i,\bx_j\in C_p}\norm{\sN^{(2)}(\bx_i)-\sN^{(2)}(\bx_j)}^2_2+\epsilon,
     \label{hoeffn}
\end{equation}
holds  $\forall \bx,\ \bx'$ on $ \sM_p^{(1)}$ with probability at-least $(1-\xi^{(2)}-\sE_p^{(2)}\LRp{m_p,\xi^{(2)}\epsilon})$. Therefore, from \eqref{hoeffn} the stability function  $\tilde{\delta}_{p}\LRp{\gamma^{(2)},\ \xi^{(2)}}$  is defined as follows:
\begin{equation}
 \tilde{\delta}_{p}\LRp{\gamma^{(2)},\ \xi^{(2)}}=\frac{1}{\xi^{(2)} \times m_p^2}\sum_{\bx_i,\bx_j\in C_p}\norm{\sN^{(2)}_{\boldsymbol{\theta}^*(\gamma^{(2)})}(\bx_i)-\sN^{(2)}_{\boldsymbol{\theta}^*(\gamma^{(2)})}(\bx_j)}^2_2+\epsilon,
 \label{expectation}
\end{equation}
where we have introduced  the subscript $\boldsymbol{\theta}^*(\gamma^{(2)})$ to  denote the optimal weights and biases of the trained layer $\sN^{(2)}$ and $\xi^{(2)}\in \LRp{0,\ 1}$.  We will first  prove that (\ref{expectation}) is a valid stability function that satisfies condition \ref{qw_1} and condition \ref{qw} in  \cref{delta-def} for every choice of $\xi^{(2)}$. Condition \ref{qw_1} is satisfied  by setting $\gamma^{(2)}=\gamma^u$ in \eqref{renormalized_loss}, since manifold regularization is the sole component in the total loss and consequently \eqref{manif_pr} is minimized. The minimizers in this case can be identified by noticing that: 
\begin{equation}
    \begin{aligned}
        &\norm{ \sN^{(2)}_{\boldsymbol{\theta}^*(\gamma^{(2)})}(\bx_i)-\sN^{(2)}_{\boldsymbol{\theta}^*(\gamma^{(2)})}(\bx_j)}^2_2=\\
&\norm{\bx_i-\bx_j+h^{(2)}\LRp{\bW^{(2)*} \bx_i+\bb^{(2)*}}-h^{(2)}\LRp{\bW^{(2)*} \bx_j+\bb^{(2)*}}}_2^2,        
        \end{aligned}
\end{equation}
which follows from (\ref{ntm}). Based on assumption \ref{gl_3}, we immediately notice that the minimizers are $\bb^{(2)*}=0$ and $\bW^{(2)*}=(-\lambda)\times \mathcal{I}$, where $\mathcal{I}$ is the identity matrix and $\lambda$ is a constant depending on the activation function employed. In this case, the minimizers $\bb^{(2)*},\ \bW^{(2)*}$ leads to the network learning a zero function and $\tilde{\delta}_p\LRp{\gamma^u,\ \xi^{(2)}}=\epsilon, \ \forall p$ in (\ref{expectation}).
Now since the minimizers $\bb^{(2)*},\ \bW^{(2)*}$ is unique for any choice of similarity matrix $\Bbeta$ in \eqref{manifold_strength}  (assumption \ref{global_minimizer_assumption}), we also note that $\tilde{\delta}_p\LRp{\gamma^u,\ \xi^{(2)}}=\epsilon$ for any $p$ if and only if the network learns a zero function. Therefore, by contraposition it follows that  $\forall p,\ \tilde{\delta}_p\LRp{0,\ \xi^{(2)}}>\epsilon$ since in the absence of manifold regularization term (i.e $\gamma^{(2)}=0$), it is necessary that the hidden layers learns a non-zero function to fit the training data. In conclusion, we have proved the following properties for $\tilde{\delta}_{p}\LRp{\gamma^{(2)},\ \xi^{(2)}}$:
\begin{equation}
    \tilde{\delta}_p\LRp{\gamma^u,\ \xi^{(2)}}=\epsilon,\quad \ \tilde{\delta}_p\LRp{0,\ \xi^{(2)}}>\epsilon,\ \ \forall p.
    \label{prop_stab}
\end{equation}
Now, in order to prove  condition \ref{qw} of \cref{delta-def}, it is sufficient to show that $\tilde{\delta}_{p}\LRp{\gamma^{(2)},\ \xi^{(2)}}$ is continuous with respect to $\gamma^{(2)}$. 
%To that end, let's denote the unique global minimizer (assumption \ref{global_minimizer_assumption}) for the loss \eqref{loss_total} as $\boldsymbol{\theta}^*(\gamma)$.
Note that, $\forall \gamma^{(2)} \in [0,\ \gamma^u]$ we have:
\begin{subequations}
  \begin{align}
   \label{a_con}
     \lim_{r\rightarrow 0} \tilde{\delta}_{p}\LRp{r+\gamma^{(2)},\ \xi^{(2)}}&=\lim_{r\rightarrow 0}  \LRs{\frac{1}{\xi^{(2)}\times m_p^2}\sum_{\bx_i,\bx_j\in C_p}\norm{ \sN^{(2)}_{\boldsymbol{\theta}^* (r+\gamma^{(2)})}(\bx_i)-\sN_{\boldsymbol{\theta}^* (r+\gamma^{(2)})}^{(2)}(\bx_j)}^2_2}+\epsilon\\
      \label{b_con}
    &= \frac{1}{\xi^{(2)}\times m_p^2}\LRs{\sum_{\bx_i,\bx_j\in C_p}\lim_{r\rightarrow 0}   {\norm{ \sN^{(2)}_{\boldsymbol{\theta}^* (r+\gamma^{(2)})}(\bx_i)-\sN_{\boldsymbol{\theta}^* (r+\gamma^{(2)})}^{(2)}(\bx_j)}^2_2}}+\epsilon\\
       \label{c_con}
   &= \frac{1}{\xi^{(2)}\times m_p^2}\sum_{\bx_i,\bx_j\in C_p}{\norm{ \sN^{(2)}_{\boldsymbol{\theta}^* (\gamma^{(2)})}(\bx_i)-\sN_{\boldsymbol{\theta}^* (\gamma^{(2)})}^{(2)}(\bx_j)}^2_2}_j+\epsilon\\
   &=\tilde{\delta}_{p}\LRp{\gamma^{(2)},\ \xi^{(2)}}
  \end{align}
\end{subequations}
where we have used the fact that $\boldsymbol{\theta}^*(\gamma^{(2)})$ is a continuous function in (\ref{c_con})  based on assumption \ref{global_minimizer_assumption} and using the results of lemma \ref{upper_hemicontinuity}.  Therefore,  condition \ref{qw} of  \cref{delta-def} is satisfied.
Now, applying the above arguments for $\delta-$stability recursively for subsequent neural transfer maps $\sN^{(i)}$, it follows that  property \eqref{stability_transfer}
holds for any subsequent neural transfer maps $\sN^{(i)}$, thereby concluding the proof.

\end{proof}
%\begin{remark}
%\label{remark_35}
%Note that in practice, having only low number of data points could affect the $\delta-$stability due to the violation of condition \ref{gl_2} in Proposition \ref{stability_prop}. 
%\begin{enumerate}
   % \item Note that in practice, having only low number of data points could affect the $\delta-$stability due to the violation of condition \ref{gl_2} in Proposition \ref{stability_prop}.
   % \item Further, $\delta-$stability could only be guaranteed in a probabilstic sense.
  %  \item Note that with some additional conditions (refer Appendix \ref{additional_condition}) one can show that  $\tilde{\delta}_{j}^{(i+1)}(\gamma)$ in (\ref{stability_transfer}) can be made as small as possible (even zero) for a sufficiently large choice of $\gamma$. 
%\end{enumerate}
%\end{remark}
\begin{remark}
\label{choose_manif}
%It is well known  that initial layers in a deep convolutional neural network \krish{cite} focus on creating meaningful representation of data while  the later layers focus on the actual classification/regression task in hand \cite{yosinski2014transferable}. \krish{see if there is a way for numerical demonstration}. Inspired by this observation and 
In order to reduce the number of hyperparameters, we consider a fixed growth/decay rate for the  manifold regularization weight, i.e. $\gamma^{(i+1)}=\kappa\gamma^{(i)}$ in \eqref{renormalized_loss}. When $\kappa \in \LRp{0,1}$, more emphasis is placed on minimizing the data loss  $\Phi_d$ in the later layers while the initial layers focus on stability. This strategy is suitable for regression/classification task with sufficient number of data points as demonstrated in 
\cref{bost_sec} and \cref{sub_image}. However, when the data set is sparse we consider choosing a small value for $\gamma^{(2)}$ and $\kappa>1$  such that the initial layers over-fits on the sparse data set and stability is imparted in the later layers. The feasibility of such a training procedure for sparse data sets is discussed in \cref{artific_clus}.
{\textit{In either case, this leads to tuning only parameter $\gamma^{(2)}$ in our procedure. \cref{stabili_cor} below shows that employing manifold regularization in this way yields an $\epsilon-\delta$ stability 
 promoting algorithm for the entire network $\sN$.} In \cref{stabili_cor},   we also look at the asymptotic behavior when $m_p\rightarrow \infty$, i.e one has access to unlimited training data points in each cluster.}
\end{remark}
\begin{corollary}
\label{stabili_cor}
Assume that conditions in Proposition \ref{stability_prop} are satisfied and each cluster contains infinitely many samples. Further, assume that the activation function employed is Lipschitz continuous and assume $\gamma^{(i+1)}=\kappa\gamma^{(i)}$ in (\ref{renormalized_loss}), where $\kappa$ is a fixed growth/decay rate  for the manifold regularization weight. Then, employing manifold regularization of the form \eqref{manif_modif} with $\beta_p=1,\ \forall p$  yields a probabilistic $\epsilon-\delta$ stability promoting algorithm for network $\sN$ with $\epsilon=0$ in the sense of \cref{delta-def}.  
 That is, there exists a stability function $\delta^{\sN}_p\LRp{\gamma,\ \xi^{(2)},\dots \xi^{(L)}}: [0,\ \gamma^u]\times \LRp{0,\ 1}^{L-1} \rightarrow \real$  that satisfies  conditions \ref{qw_1} and \ref{qw} for any fixed $\xi^{(2)},\dots \xi^{(L)}$, such that  in the limit $m_p\rightarrow \infty$\footnote{Inequality \eqref{total_stab} is an asymptotic property here. For more details, refer to the proof in \cref{corol_algo}. 
 }, $\forall \bx,\ \bx' \in \sM_p^{(0)} \subset \sM^{(0)}$:
 
%\begin{equation}
  %  \norm{\sN\LRp{\bx}-\sN\LRp{\bx'}}_2^2 \leq  c\epsilon_j \prod_{i=2}^L \LRp{\frac{\tilde{\delta}_{j}^{(i)}(\gamma)}{\xi^{(i)}}}^{\frac{1}{2^{(L-i+1)}}}=\delta_{\sN}\LRp{\gamma^{(2)},\dots \gamma^{(L)},\ j}, 
  %  \label{total_stab}
%\end{equation}
\begin{equation}
    \norm{\sN\LRp{\bx}-\sN\LRp{\bx'}}_2^2 \leq  \delta^{\sN}_p\LRp{\gamma,\ \xi^{(2)},\dots \xi^{(L)}}, 
    \label{total_stab}
\end{equation}
holds   with probability at-least $\LRp{1-\sum_{i=2}^L\xi^{(i)}}$, where $\gamma=\gamma^{(2)}$ is the initial weight chosen for the manifold regularization term in the loss \eqref{renormalized_loss}. 
\end{corollary}
\begin{proof}
Proof is provided in \cref{corol_algo}.    
\end{proof}
\begin{remark}
Note that for a given $\xi^{(2)},\dots \xi^{(L)}$, the stability function $\delta^{\sN}_p\LRp{\gamma,\ \xi^{(2)},\dots \xi^{(L)}}$ can be made as small as one wishes by varying the parameter $\gamma$ (also see remark \ref{rem_sts}). 
%For \eqref{total_stab} to hold with a high probability, one may choose a very small magnitude for $\xi^{(i)}$  for the initial layers (i.e for small $i$) since $\LRp{\xi^{(i)}}^{-\frac{1}{2^{(L-i+1)}}}\rightarrow 1$ for a  deep network (large $L$, and small $i$).
A numerical demonstration of controlling the  stability function by varying $\gamma$ is provided in 
Figure \ref{inverse_training} of \cref{Inverse_problem}. Further, by choosing the optimal stability parameter $\gamma^*$ in \eqref{total_stab}, we achieve a $\delta-$stable network $\sN$  in the sense of \cref{delta_deff}.
\end{remark}

\subsubsection{Layerwise training saturation problem}
\label{train_sat_alg}

Even though we have ensured that   \cref{AlgoGreedyLayerwiseResNet} promotes $\epsilon-\delta$ stability, it is imperative to analyse any training saturation problem that might arise from certain hyperparameter settings. In this section, we consider the case when $\kappa \in \LRp{0,1}$ in  \cref{stabili_cor} and derive the necessary conditions for trainability of a newly added layer (see \cref{TPOP}). To that end, let us define the following:
\begin{definition}{Layerwise training promoting property (LTP)}{}
\label{LTP_prop}

A non-linear activation function $h(\cdot)$  possesses the layerwise training promoting property (LTP) if $h'(0)\neq0$.

%\begin{itemize}
  %  \item $h(0)=0$,
%    \item $h'(0)\neq0$.
%\end{itemize}
\end{definition}

%It is easy to see that activation functions such as `ELU' and `Tanh' satisfy the LTP property, whereas `Sigmoid' and `ReLU' do not. However, any activation function can be forced to satisfy the LTP property by a simple transformation. In this work, the activation functions $h^{(i)}$ for $i=\{2,3,...L\}$ in \cref{AlgoGreedyLayerwiseResNet} is selected to be the exponential linear
%unit (ELU) (\cite{clevert2015fast}).

\begin{definition}{Trainability of an added layer}

A newly added $(i+1)^{th}$ layer with parameters $\bW^{(i+1)}$ and ${\bf b}^{(i+1)}$ initialized as $\bW_0$ and $\bb_0$, respectively, is trainable with respect to the $j^{th}$ training pair $\{\bY_j^{(i)}, \bC_j\}$ if

\[\LRs{\pp{\lag_j}{\bW^{(i+1)}}}_{\bW^{(i+1)}=\bW_0}\neq {\bf{0}}, \quad \LRs{\pp{\lag_j}{\bb^{(i+1)}}}_{\bb^{(i+1)}=\bb_0}\neq {\bf{0}},\]
where $\lag_j$ is the Lagrangian defined for the $j^{th}$ training sample in \eqref{lagrangian} in \cref{back_propagation}.

 %the loss function has strong non-zero gradients with respect to the  parameters of an added layer  when the previous layer is trained to a local minima.

\end{definition}

\begin{proposition}[Necessary condition for trainability of newly added layer]
%[\krish{Necessary conditions for training promoting-output preserving layer addition: I am confused with this necessary conditions. Recall that p implies q, then q is the necessary condition for p. This phrase "training promoting-output preserving layer addition" is challenging grammatically: do you mean output-preserving-promoting? even that it is very cumbersome. Yes Tan. That's right. I have modified teh title. This should be the correct title}]
\label{TPOP}
Consider training the $(i+1)^{th}$ hidden layer with the loss \eqref{loss_total} and parameters $\bW^{(i+1)}$ and ${\bf b}^{(i+1)}$  initialized as zero and  assume that $h^{(i+1)}(0) =0$ and $\alpha^{(i)}=\alpha^{(i+1)}=0$ (sparsity regularization). Further, assume that the $i^{th}$ hidden layer is trained  to a local minimum with respect to the $j^{th}$ training sample such that $\LRp{h^{(i)}}'\LRp{\bW^{(i)*} \sR^{(i-1)}\LRp{\bY^{(i-1)}_j}+ \bb^{(i)*}}\neq {\bf{0}}$, where $\bW^{(i)*}$ and $\bb^{(i)*}$ denote the optimal weights and biases of the $i^{th}$ layer and $\sR^{(i)}$ follows assumption  \eqref{gl_3}.
Then, the necessary conditions for trainability of  an added layer are:
\begin{enumerate}
    \item  $ \gamma^{(i+1)}\neq  \gamma^{(i)}$ or $ \tau^{(i+1)}\neq  \tau^{(i)}$; 
    \item Activation  $h^{(i+1)}$ should satisfy the LTP property.
\end{enumerate}

% Furthermore, let $E^{i*}$ denote the final value of the data loss function after training the $i^{th}$ layer. Then,  the loss at $n^{th}$ iteration in training  the $(i+1)^{th}$ layer, $E^{(i+1)}_{n}< E^{i*}$ only if $ \gamma^{(i+1)}<  \gamma^{(i)}$ and $ \tau^{(i+1)}<  \tau^{(i)}$ and the activation  $h^{(i+1)}$ satisfies the LTP property.

\end{proposition}

\begin{proof} 
The ResNet forward propagation (see \eqref{Res_two}) for the newly added layer with weights and biases initialized as $zero$ gives:
\begin{equation}
    \bY^{(i+1)} = \sR^{(i)}\LRp{\bY^{(i)}} + h^{(i+1)}\LRp{{\bf{0}}\times \sR^{(i)}\LRp{\bY^{(i)}} + {\bf{0}}}.
    \label{output_preserving}
\end{equation}
Since $h^{(l+1)}(0)=0$,  we have $\bY^{(i+1)} = \sR^{(i)}\LRp{\bY^{(i)}}$ on adding the new layer at the beginning of training and the network output $h_{\mathrm{pred}}\LRp{\bW_{\mathrm{pred}}\sR^{(i+1)}\LRp{\bY^{(i+1)}}+ {\bf b}_{\mathrm{pred}}}$ remains unchanged since  $\sR^{(i+1)}\sR^{(i)}\LRp{\bY^{(i)} }=\sR^{(i)}\LRp{\bY^{(i)} }$, which follows from assumption \eqref{gl_3}. Now, using  \eqref{output_preserving}, the gradient for the newly added layer evaluated at $\bf{0}$ for a particular $j^{th}$ training sample can be written as (see  \cref{back_propagation}): 

\begin{equation}
    \begin{aligned}
        &\pp{\lag_j}{\bb^{(i+1)}}=\\
        &-\LRp{\bR^{(i)}\LRp{\bW_{\mathrm{pred}}^{*}}^T\LRs{ h_{\mathrm{pred}}'\LRp{\bW_{\mathrm{pred}}^*\sR^{(i)}\LRp{\bY^{(i)}_j}+ {\bf b}_{\mathrm{pred}}^*}\circ{\Blambda_2^{(i+1)}}}-\pp{ \Phi_m^{(i+1)}}{\bY^{(i)}_j}}\circ {{\LRp{h^{(i+1)}}}'({\bf{0}})},        
        \end{aligned}
        \label{bias}
\end{equation}
and,
\begin{equation}
\pp{\lag_j}{\bW^{(i+1)}} 
=\pp{\lag_j}{\bb^{(i+1)}} \LRp{{\sR^{(i)}\LRp{\bY^{(i)}_j}}}^T,
\label{wei}
\end{equation}
where $\lag_j$ denotes the Lagrangian defined in  \cref{back_propagation},  \eqref{lagrangian}, $\bW_{\mathrm{pred}}^*, \ {\bf b}_{\mathrm{pred}}^*$ represents the optimal parameters for the output layer obtained after training the  $i^{th}$ layer, and $\Blambda_2^{(i+1)}$
is defined as:
\begin{equation}
  \Blambda_2^{(i+1)}=-\LRp{\pp{\Phi_d}{\bY^{(o)}_j}+\pp{\Phi_p^{(i+1)}}{\bY^{(o)}_j}}. 
  \label{lam_evolve}
\end{equation}
where, $\bY^{(o)}_j$ represents the output of the network.
Therefore, it is clear from \eqref{bias} that $\LRp{h^{(i+1)}}'({\bf{0}})\neq 0$ (LTP property) is a necessary condition for trainability of the newly added layer. Further, since the previous layer, i.e the $i^{th}$ layer is trained to  a local minima, then one also has the following condition (see \cref{back_propagation} equation \eqref{bias_gradient}):
\begin{equation}
    \LRp{\bR^{(i)}(\bW_{\mathrm{pred}}^{*})^T\LRs{ h_{\mathrm{pred}}'\LRp{\bW_{\mathrm{pred}}^*\sR^{(i)}\LRp{\bY^{(i)}_j}+ {\bf b}_{\mathrm{pred}}^*} \circ{\Blambda_2^{(i)}}}-\pp{ \Phi_m^{i}}{\bY^{(i)}_j}}\circ \bz={\bf{0}},
    \label{local_min_previous}
\end{equation}
where, $\bW^{(i)*},\ \bb^{(i)*}$ denotes the trained parameters of the $i^{th}$ layer and the vector $\bz$ given by:
\[\bz=\LRp{h^{(i)}}'\LRp{\bW^{(i)*}\sR^{(i-1)}\LRp{\bY^{(i-1)}_j} + \bb^{(i)*}}\neq {\bf{0}}, \]
by assumption in \cref{TPOP}. Therefore, condition \eqref{local_min_previous} translates to the following:
\begin{equation}
\begin{aligned}
   \pp{ \Phi_m^{i}}{\bY^{(i)}_j}= \bR^{(i)}(\bW_{\mathrm{pred}}^{*})^T\LRs{ h_{\mathrm{pred}}'\LRp{\bW_{\mathrm{pred}}^*\sR^{(i)}\LRp{\bY^{(i)}_j}+ {\bf b}_{\mathrm{pred}}^*} \circ \Blambda_2^{(i)}}.
    \label{optimality}
\end{aligned}
\end{equation}
Further from \eqref{lam_evolve}, we have:
\begin{equation}
  \Blambda_2^{(i+1)}=\Blambda_2^{(i)}+\pp{\Phi_p^{i}}{\bY^{(o)}_j}-\pp{\Phi_p^{(i+1)}}{\bY^{(o)}_j}.  
  \label{lamb_evolve}
\end{equation}
In addition from \eqref{bias}, for trainability of the newly added layer it is necessary that:
\[ \bR^{(i)}(\bW_{\mathrm{pred}}^{*})^T\LRs{ h_{\mathrm{pred}}'\LRp{\bW_{\mathrm{pred}}^*\sR^{(i)}\LRp{\bY^{(i)}_j}+ {\bf b}_{\mathrm{pred}}^*} \circ{\Blambda_2^{(i+1)}}}-\pp{ \Phi_m^{(i+1)}}{\bY^{(i)}_j}\neq \bf{0}. \]
The above condition can be simplified using  \eqref{optimality}, and \eqref{lamb_evolve} as:
\begin{equation}
\pp{ \Phi_m^{i}}{\bY^{(i)}_j}-\pp{ \Phi_m^{(i+1)}}{\bY^{(i)}_j}+\bR^{(i)}(\bW_{\mathrm{pred}}^{*})^T\LRs{ h_{\mathrm{pred}}'\LRp{\bW_{\mathrm{pred}}^*\sR^{(i)}\LRp{\bY^{(i)}_j}+ {\bf b}_{\mathrm{pred}}^*} \circ \LRp{\pp{ \Phi_p^{i}}{\bY^{(o)}_j}-\pp{ \Phi_p^{(i+1)}}{\bY^{(o)}_j}}}    \neq {\bf{0}}.    
\end{equation}
Using the \eqref{two-cost_a} in  \cref{back_propagation}, the above condition translates to:
\begin{equation}
  \LRp{\gamma^{(i)}-\gamma^{(i+1)}}  \pp{ \Phi_m}{\bY^{(i)}_j}+\LRp{\tau^{(i)}-\tau^{(i+1)}}(\bW_{\mathrm{pred}}^{*})^T\LRs{ h_{\mathrm{pred}}'\LRp{\bY^{(i)}_j\bW_{\mathrm{pred}}^*+ {\bf b}_{\mathrm{pred}}^*} \circ    \pp{ \Phi_p}{\bY^{(o)}_j}}    \neq {\bf{0}}.
  \label{final_optim_condition}
\end{equation}
Thus, it is necessary that $\gamma^{(i+1)}\neq \gamma^{(i)}$ or $\tau^{(i+1)}\neq \tau^{(i)}$ for trainability.
\end{proof}

\begin{corollary}[Training saturation]
\label{TPOP_cor}
%Let $\Phi_d^{(i)*}$ denote the  data loss  after training the $i^{th}$ layer to a local minimum with respect to each training sample.
%(assume stochastic gradient descent).
Assume that the necessary conditions of  \cref{TPOP} are satisfied, by 
choosing $\gamma^{(i+1)}=\kappa_1 \gamma^{(i)}$ and $\tau^{(i+1)}=\kappa_2 \tau^{(i)}$, where 
$\kappa_1,\kappa_2 \in \LRp{0,1}$. Then
\[
\lim_{i \to \infty}\pp{\lag_j}{\bb^{(i+1)}} = {\bf{0}}, \quad \text{ and }
\lim_{i \to \infty} \pp{\lag_j}{\bW^{(i+1)}} = {\bf{0}},
\]
for all  $j = 1,\hdots M$.
%\tanbui{is each term in each limit a number of vectors? If the latter (the proof says so), the limits need to be a zero vector.}
\end{corollary}

\begin{proof}
Since the $i^{th}$ layer is supposed to be trained to a local minimum with respect to each training sample,  we have:
\[\LRs{\pp{\lag_j}{\bb^{(i)}}}_{\bb^{(i)}=\bb^{(i)*}}={\bf{0}},\quad \LRs{\pp{\lag_j}{\bW^{(i)}}}_{\bW^{(i)}=\bW^{(i)*}}={\bf{0}}, \quad \forall j = 1, \hdots\]
where, $\bW^{(i)*}$ and $\bb^{(i)*}$ denote the optimal parameters of the $i^{th}$ layer.
Since $\lim_{i \to \infty}|\gamma^{(i+1)}-\gamma^{(i)}| = 0$, $\lim_{i \to \infty} |\tau^{(i+1)}-\tau^{(i)}| = 0$, it follows from  \eqref{final_optim_condition} and \eqref{bias} that:
\[\LRs{\pp{\lag_j}{\bW^{(i+1)}}}_{\bW^{(i+1)}={\bf{0}}}\rightarrow  {\bf{0}}, \quad \LRs{\pp{\lag_j}{\bb^{(i+1)}}}_{\bb^{(i+1)}={\bf{0}}}\rightarrow {\bf{0}},\quad \forall j = 1,\hdots M.\]
as $i$ approaches $\infty$.
\end{proof}

\begin{remark}{}{}
The results of \cref{TPOP_cor} implies that  for sufficiently large $i$, both $\pp{\lag_j}{\bb^{(i+1)}}$ and $\pp{\lag_j}{\bW^{(i+1)}}$ are small. In other words, we face the problem of vanishing gradient when adding new layer and
the training  saturates, i.e., either little or no  decrease in the total loss given by  \eqref{loss_total} for sufficiently large $i$.
\end{remark}
\begin{remark}
   Note that we have assumed $h^{(i)}(0)=0$ in  \cref{TPOP}. Further, the LTP property is a necessary condition for trainability of an added layer.  It is easy to see that activation function such as `ELU' and `Tanh' satisfy these conditions. In addition, activation functions such as $\{\mathrm{ReLU} \cite{clevert2015fast},\ \mathrm{Leaky  ReLU},\ \mathrm{SELU}\}$ can also be used in practice if one considers a non-zero subgradient at $0$. In this work, the activation functions $h^{(i)}$  in  \cref{AlgoGreedyLayerwiseResNet} are selected to be ELU. 
  % whereas `Sigmoid' does not. 
\end{remark}

\subsection{Sequential residual learning  \cref{Sequential}}
\label{math_sequential}
Recall that promoting $\epsilon-\delta$ stability (\cref{stabili_cor})  does not guarantee that the network classifies/regress correctly since training saturates for certain hyperparameter settings in \cref{AlgoGreedyLayerwiseResNet} (see \cref{TPOP_cor}). To overcome this issue, we incorporate a post-processing stage termed as the {\it{``sequential residual learning"}} step for further improving the predictions. The idea of sequential residual learning is to train a sequence of shallow neural networks to learn the residual (i.e lower the data loss $\Phi_d$ while generalizing well) from   \cref{AlgoGreedyLayerwiseResNet} where each network $\sQ_i$ is trained for a limited epochs to prevent overfitting on the residuals.
 \cref{Sequential} creates new training problem for each network $\sQ_{i+1}$ by generating data sets of the form:
\begin{equation}
\sK_{II,i+1}=\Big \{\bX, \ \bC_{i}\Big \}, \hspace{1 cm} \bC_{i}=\bC_{i-1}-\sQ_i(\bX),\hspace{1 cm} \ i= 1,\ 2, \dots \\
\label{resid}
\end{equation}
where, $\bC_{0}=\bC$ and $\sQ_1=\sN$ is the neural network produced by \cref{AlgoGreedyLayerwiseResNet}. For a given test data point $\bx$, the output $f(\bx)$ of our algorithm is computed as:
\begin{equation}
    f(\bx)=\sN(\bx)+\sQ(\bx),\quad \mathrm{where}\ \sQ(\bx)=\sQ_2(\bx)+\dots \sQ_{r-1}(\bx).
    \label{out_total}
\end{equation}

\subsubsection{$\epsilon-\delta$ stability and approximate $\delta-$robustness}
\label{robust_seq}
Note that while manifold regularization allows us to promote $\epsilon-\delta$ stability for the network $\sN$ (final network from \cref{AlgoGreedyLayerwiseResNet})  (see \cref{stabili_cor}), it is imperative to have a mechanism that promotes $\epsilon-\delta$ stability for $\sQ$ given in \eqref{out_total} during the sequential adaptation process. Further, the main motivation of  \cref{Sequential} is to promote robustness.   Robustness is the property that if a testing sample is “similar” to a training sample, then the testing error is close to the training error  along with the condition that the algorithm fits the training data “well-enough” \cite{xu2012robustness,jin2020manifold}. If the 
 prediction on the training sample is accurate, then the prediction on a similar testing sample will also be accurate. It is therefore immediately clear that learning a $\delta-$stable function (\cref{delta_deff}) is necessary for robustness. Additionally, by using a sequence of networks to learn  the residuals in \cref{Sequential},  one  ensures that the algorithm fits the training data “well-enough”. 
 Since in practice, one fits the training data points to at most $\iota$ tolerance (to prevent overfitting),  we define approximate $\delta-$robustness as follows: 
\begin{definition}[Approximate $\delta-$robustness]
\label{robust_def}
The function $f(\bx)=\sN(\bx)+\sQ(\bx)$ in \eqref{out_total} is called approximately $\delta-$robust at an input $\bx\in \sM_p^{(0)}$ if each of the networks $\sN$ and $\sQ$ are $\delta-$stable on $\sM_p^{(0)}$ as per  \cref{delta_deff}
and $\norm{f(\bx)-c}_2\leq \iota$, where $c$ is the correct label for $\bx$ and, $\iota$ is a small positive constant chosen to prevent overfitting.
\end{definition}
Note that we refrain from the use of any regularizers in \cref{Sequential} due to the small magnitude of residuals $\bC_{i}$ in \eqref{resid} which inhibits learning in the presence of explicit regularizers. Therefore,  one is naturally left with the question on how to design a $\epsilon-\delta$ stability promoting algorithm for  $\sQ$ in  \cref{Sequential}. Proposition \ref{Lipschitz}  provides  a practical implicit way of achieving this avoiding the use of any regularizers.  {\it{ In particular, proposition \ref{Lipschitz} explores how early stopping (a form of implicit regularization in machine learning where a network is trained for a limited number of iterations) can be used to design a $\epsilon-\delta$ stability promoting algorithm  as defined in \cref{discrete_ep_del}.}} In the below \cref{Lipschitz}, the term ``iterations" refer to the number of gradient updates made during the training process.
\begin{proposition}[$\epsilon-\delta$ stability promoting algorithm via early stopping]
\label{Lipschitz}
Consider the function $\sQ(\bx)$  in \cref{Sequential} with  the input space given in  \cref{ntmm} and assume that the activation function employed is 1-Lipschitz continuous \footnote{We call a function, $h:\real^n\rightarrow \real^m$, 1-Lipschitz continuous if $\forall \bx,\ \by\in \real^n,\quad \norm{h(\bx)-h(\by)}_2\leq \norm{\bx-\by}_2$.}   for each network $\sQ_i$. Further, assume that the input set $\sM_p^{(0)}$ is compact $\forall p$. Then, employing early stopping criteria yields a  $\epsilon-\delta$ stability promoting algorithm for network $\sQ$ in \eqref{out_total} with $\epsilon=0$. That is, there exists a stability function $\delta_p^{\sQ}(\zeta): \Big \{\frac{1}{m^u+1},\ \frac{1}{m^u},\dots 1\Big \}\mapsto \real$ satisfying  conditions in \cref{discrete_ep_del} %\footnote{Let $X$ and $Y$ be non-empty sets and let $\tau$ and $\nu$ be, respectively the discrete topologies on $X$ and $Y$. Every function $f:\LRp{X,\ \tau}\rightarrow \LRp{Y,\ \nu}$ is continuous.}
%with $\zeta=\frac{1}{m+1}$, and $m\in \{0,\ 1, \dots m^u\}$ is the number of training iterations and 
where $m^u \in \mathbb{Z}^+$ is some chosen upper bound on the training iterations. In particular, $\forall \bx,\ \bx' \in \sM_p^{(0)} \subset \sM^{(0)}$:
\begin{equation}
   \norm{\sQ\LRp{\bx}-\sQ\LRp{\bx'}}_2 \leq  \sum_{i=2}^{r-1}\ell^3\times \LRs{\prod_{k=1}^3 \LRp{\sum_{l=0}^{(1/\zeta)-1}\norm{\LRp{\bg^{(k)}_l}_i}_2}}\times \epsilon_p=\delta_p^{\sQ}\LRp{\zeta},  
   \label{lipschitz_stable}
\end{equation}
where $\epsilon_p=\underset{\bx,\ \bx'\in \sM_p^{(0)}}{\sup} \norm{\bx-\bx'}_2$,   $\LRp{\bg^{(k)}_l}_i$ represents the gradients for  $k^{th}$ layer weight matrix at the $l^{th}$ iteration of gradient descent  for network $\sQ_i$\footnote{$\LRp{\bg^{(k)}_0}_i$ represents the initial weights.}, $\ell$ denotes a constant learning rate.
%, and $\zeta^*$ is the chosen stability parameter.

%Then, network $\sQ_i$ is $\delta^{(i)}_j(\zeta^*)-$stable (in a discrete sense) on $\sM_j^{(1)}$ with hyperparameter $\zeta^*\in \Big \{\frac{1}{m^u+1},\ \frac{1}{m^u},\dots 1\Big \}$, where $m^u\in \mathbb{Z}^+$ is the upper bound on the training epochs. In particular,  the stability function $\delta_j^{(i)}(\zeta)$ in definition \ref{delta-def} is given by:

%\begin{equation}
  %  \delta_j^{(i)}(\zeta)=  \ell^3\times \LRs{\prod_{k=1}^3 \LRp{\sum_{l=0}^{(1/\zeta)-1}\norm{\LRp{\bg^{(k)}_l}_i}_2}}\times \epsilon_j,  
 %   \label{lipschitz_stable}
%\end{equation}
\end{proposition}
\begin{proof}
Proof is provided in \cref{proof_lips}.
\end{proof}
\begin{remark}[Approximate $\delta-$robustness by  \cref{Sequential}]
\label{stability_2}
Recall \cref{delta_deff} for $\delta-$stable function. By tuning parameter $\gamma$ in \eqref{total_stab}, and parameter $\zeta$  in \eqref{lipschitz_stable},  each network $\sN$ and $\sQ$ is $\delta-$stable. More information on tuning the parameters for $\delta-$stability are provided in \cref{tuning}. Lastly,
for an input $\bx$ in the training data set, the residual  $\norm{f(\bx)-c}_2\leq \iota$ in \cref{robust_def} is achieved after adding sufficient number of networks leading to approximate $\delta-$robustness.
\end{remark}

\begin{remark}[Manifold regularization vs. early stopping]
    Note that even though both manifold regularization and early stopping helps one design a $\epsilon-\delta-$ stability promoting algorithm, it is important to note the subtle difference between the stability functions $\delta^{\sN}_p\LRp{\gamma,\ \xi^{(2)},\dots \xi^{(L)}}$ in \eqref{total_stab} and $\delta_p^{\sQ}(\zeta)$ in \eqref{lipschitz_stable}. While $\delta^{\sN}_p\LRp{\gamma,\ \xi^{(2)},\dots \xi^{(L)}}$  explicitly depends  on the geometry (similarity) of data points,  the stability function $\delta_p^{\sQ}(\zeta)$ is a discrete function that does not directly depend on the geometry of the dataset. In our numerical examples (see results in \cref{bost_sec}),  we demonstrate that the $\delta-$stability enforced by manifold regularization in \cref{AlgoGreedyLayerwiseResNet} is indispensable for achieving better generalization performance in comparison to a baseline trained with a pure early stopping criteria.
\end{remark}

\section{Numerical Experiments}
In this section, we numerically demonstrate the proposed approach on both simulated data sets and real world data sets as follows:
\begin{enumerate}
    \item Regression task: Boston house price prediction problem.
    \label{prob_1}
    \item Physics informed adaptive neural network (PIANN).
     \label{prob_2}
     \item Adaptive learning of inverse maps from sparse data.
      \label{prob_3}
    \item Image classification problem: MNIST data set.
     \label{prob_4}
\end{enumerate}
General experimental settings for all the problems and descriptions of methods adopted for comparison are described in \cref{general_setting}.  We will show in our numerical experiments that the use of  \cref{Sequential} is inevitable for problems \ref{prob_1} and \ref{prob_4} to deal with the training saturation problem as described in  \cref{TPOP_cor}. However, the nature of problems \ref{prob_2} and \ref{prob_3} allows us to devise a different hyperparameter settings that evades the use of   \cref{Sequential} (by mitigating the training saturation problem).

\subsection{Regression task: Boston house price prediction problem}
\label{bost_sec}
This regression problem  aims to predict the housing prices in Boston and is a common prototype problem for testing ML regression algorithms \cite{chatterjee2017progressive, harrison1978hedonic, shahhosseini2019optimizing}.  The  input setting for \cref{AlgoGreedyLayerwiseResNet} and  \cref{Sequential} as well as other details are provided in \cref{boston_setting} and \cref{inputs}.

\begin{figure}[h!]      
  % \begin{subfigure}[b]{\textwidth}
\hspace{-0.7 cm}
  \begin{tabular}{c}

      \begin{tabular}{c}

          \centering
          \includegraphics[scale=0.445]{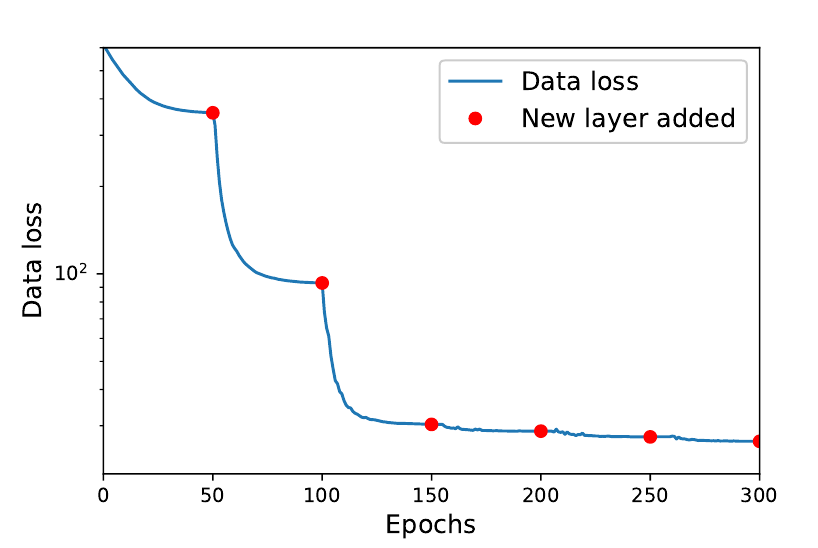}
       %   \includegraphics[scale=0.143]{Figures/Boston/training_boston_log.pdf}
         %   \includegraphics[scale=0.41]{Figures/Boston/training_boston_loglog.eps}
          % \caption{Effect of noise on DI and Tikhonov solutions}
          % \figlab{some_other_good_name}

      \end{tabular}

\hspace{-1 cm}

      \begin{tabular}{c}

          \centering
          \includegraphics[scale=0.345]{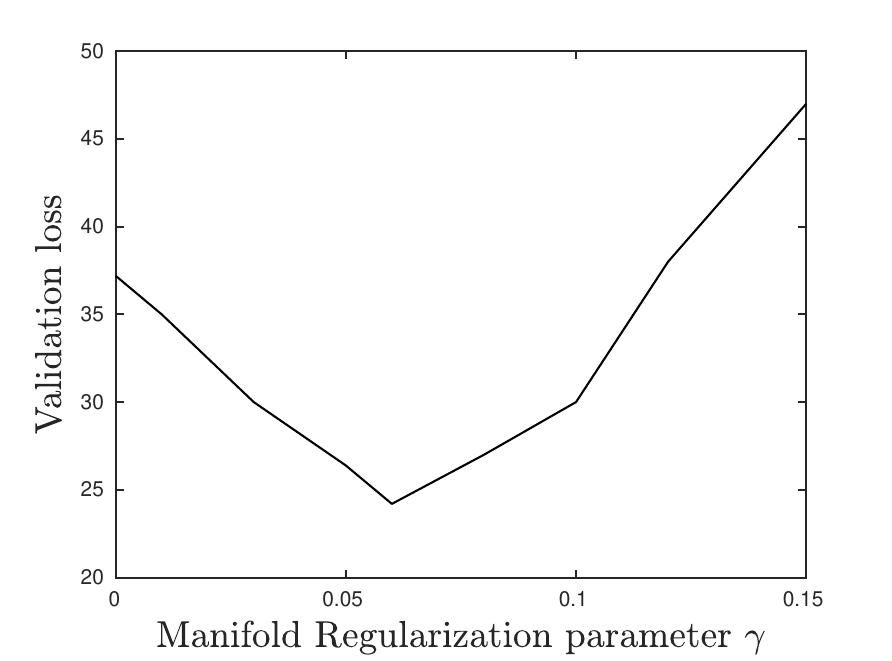}
          % \caption{Effect of noise on DI and Tikhonov solutions}
          % \figlab{some_good_name}

      \end{tabular}

\hspace{-0.7 cm}

      \begin{tabular}{c}

          \centering
          \includegraphics[scale=0.345]{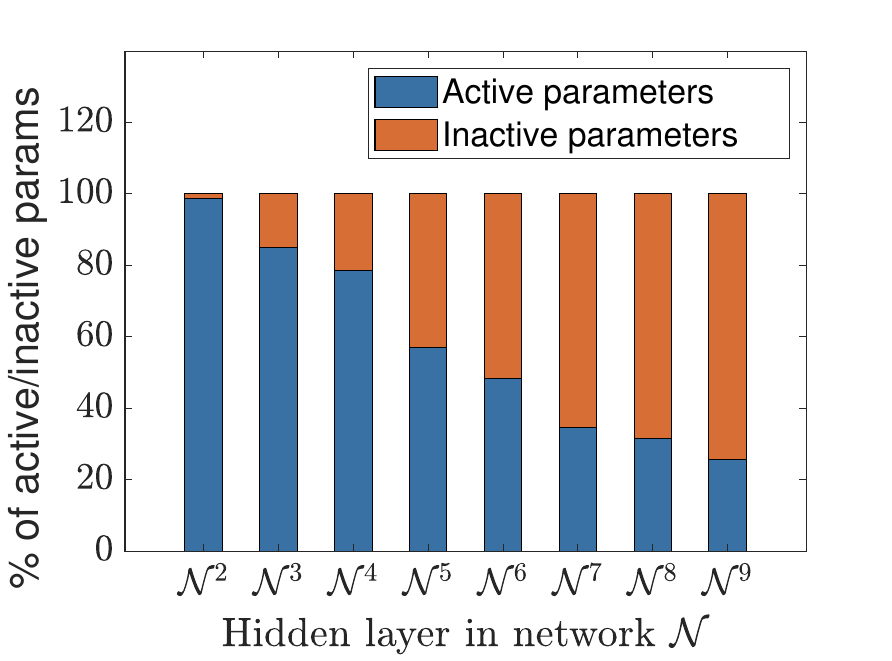}
          % \caption{Effect of noise on DI and Tikhonov solutions}
          % \figlab{some_good_name}

      \end{tabular}

  \end{tabular}
 \caption{Left to right: layerwise training curve on Boston house price prediction problem by   \cref{AlgoGreedyLayerwiseResNet}; Importance of  manifold regularization ($\gamma$) in   \cref{AlgoGreedyLayerwiseResNet}; Active and inactive parameters in each hidden layer.}
  \label{Boston_training}
  % \end{subfigure} \\
\end{figure}

%\newpage

%\begin{figure}[h!]      
  % \begin{subfigure}[b]{\textwidth}
 %         \includegraphics[scale=1]{Figures/Boston/training_boston_log.eps}
  %        \caption{a}
%\end{figure} 

%\newpage

The layerwise training curve is shown in Figure \ref{Boston_training} which shows that, subsequent layers are able to learn on top of the representations of previous layers. The percentage of active and inactive parameters (consequence of sparsity regularization) in each layer is shown in Figure \ref{Boston_training}. From Figure \ref{Boston_training} it is clear that initial layers matter more in a deep neural network  requiring more number of parameters \cite{raghu2017expressive}. We also note from  Figure \ref{Boston_training} that   \cref{AlgoGreedyLayerwiseResNet} is only capable of reducing the training loss to $24.2$ and training saturates (negligible decrease in the loss) as predicted in \cref{TPOP_cor}.

Further, Figure \ref{Boston_training} (middle figure) shows how to choose the optimal stability parameter (manifold regularization parameter $\gamma$) for $\delta-$stability as defined in  \cref{delta_deff}.   The optimal parameter ($\gamma^*=0.06$) is the one that gives the lowest validation loss.
From Figure \ref{Boston_training} (middle figure), it is evident that in the absence of manifold regularization (i.e $\gamma=0$), stability is not promoted and consequently the final validation loss achieved by \cref{AlgoGreedyLayerwiseResNet} is considerably higher in this case. 
Note that even though we do not have infinitely many training samples as dictated by Proposition \ref{stability_prop}, the effect of manifold regularization is quite pronounced in this case. In order to further decrease the loss (promote approximate $\delta-$robustness), we resort to  \cref{Sequential}. The inputs for  \cref{Sequential} are provided in \cref{inputs}.

\begin{figure}[h!t!b!]      
  % \begin{subfigure}[b]{\textwidth}
\hspace{-1.2 cm}
  \begin{tabular}{c}
      \begin{tabular}{c}
          \centering
          \includegraphics[scale=0.315]{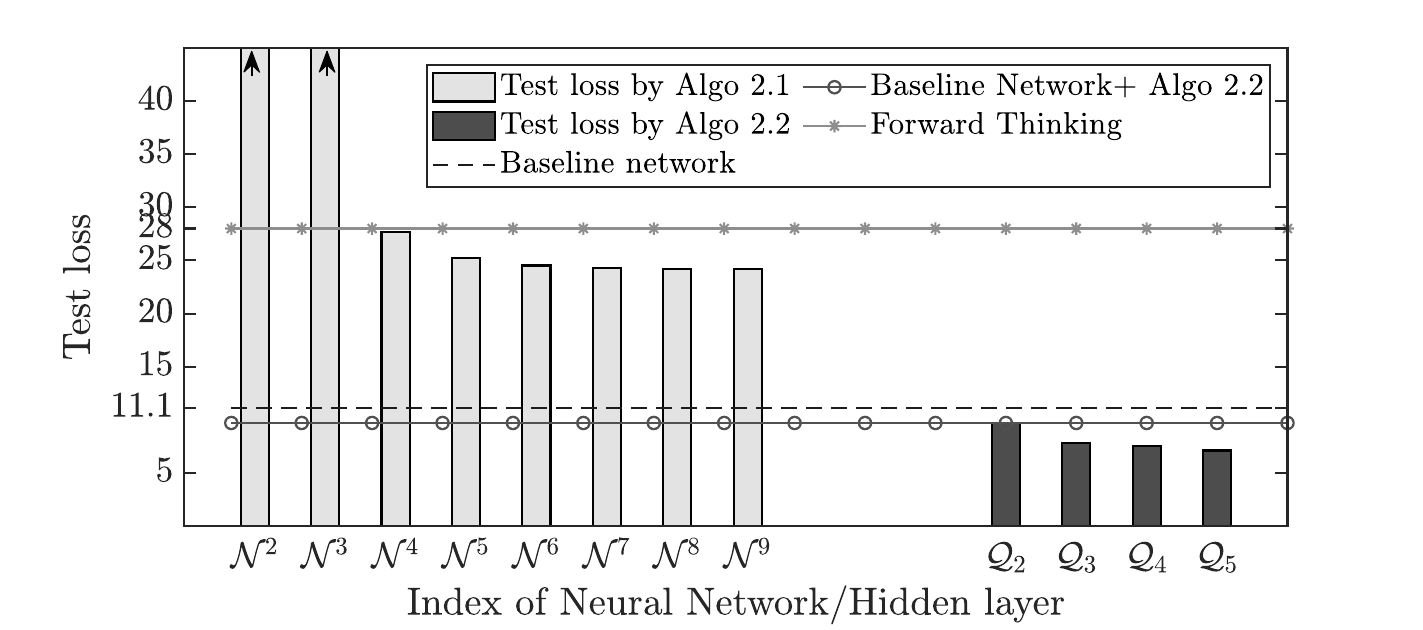}
          % \caption{Effect of noise on DI and Tikhonov solutions}
          % \figlab{some_other_good_name}
      \end{tabular}
      
\hspace{-0.6 cm}

           	\begin{tabular}{|c | c | c | c|}
        \hline
    Method &  Test   &  Params.  trained & Total \\ 
        &   loss &  simultaneously & params.  \\ \hline
            Proposed  & 7.11 & $11,601$&48, 474\\ \hline
    Baseline   & 11.2 & $82, 301$& 82, 301 \\ \hline
 %    Algo. \ref{Sequential}  & 9.8 & $11, 601$& 48, 021 \\ 
   %  (1 network) &  &  & \\ \hline
     Baseline +   & 10.5 & $82, 301$& 82,905 \\ 
     Algo. \ref{Sequential} &  &  & \\ \hline
Forward- &  &  & \\ 
Thinking \cite{hettinger2017forward} & 28.0 &  $11,601$& 82, 301\\ \hline
Approach \cite{hettinger2017forward} &  &  & \\ 
+  Algo. \ref{Sequential}& 11.5 &  $11,601$& 82, 905\\ \hline
	\end{tabular} 
  \end{tabular}
  \caption{Boston house price prediction problem.
%   Left to right: Summary of architecture adaptation results; Performance of baseline network and other layerwise training methods.
Comparison between the proposed approach and other methods. As can be seen, the proposed two-stage approach is the most accurate (by a large margin) with the least number of network parameters.}
\label{summary_boston}
  % \end{subfigure} \\
\end{figure} 
The adaptation results and performance comparison with other methods (the baseline and the approach from \cite{hettinger2017forward})
  are summarized in Figure \ref{summary_boston}.  From Figure \ref{summary_boston}, it 
  is clear that our proposed approach outperformed all the other methods by a good margin. The number of trained parameters in each step of  \cref{Sequential} and from other methods  is provided as a Table in  Figure \ref{summary_boston}. As can be seen, the proposed two-stage approach is the most accurate (by a large margin) with the least number of network parameters. Further, a random architecture search based on the approach in \cite{li2020random} produced a best test loss of $9.28$ which was also considerably higher than our approach.  It is noteworthy that using \cref{Sequential} on top of the baseline network trained with  early stopping criteria (shown in Figure \ref{summary_boston}) did not provide  much improvements on the result since  the baseline network  lacks the $\delta-$ stability  imposed by the manifold regularization.  Similar conclusion is observed when applying   \cref{Sequential} on top of forward-thinking approach \cite{hettinger2017forward} (see Table \ref{summary_boston}). 

\subsection{Physics informed adaptive neural network (PIANN)}
\label{PIANN}
Consider partial differential equation (PDE) of the form:
 \begin{align}
\sG(\bx, \by, a)&=0, \ \ \text{in} \ \ \Omega,\label{general}\\
\by&=g \ \ \ \text{in} \ \ \partial \Omega,\label{general_boundary}
\end{align} 
where, $a\in \sA$ is the PDE coefficient, $\sA$ is the parameter space, $\by\in \sB$ is the unknown solution, $\sB$ is the solution space and the operator $\sG: \sB \times \sA\rightarrow \sF$ is in general a non-linear partial differential operator.  Physics informed neural networks (PINNs) \cite{yu2018deep, lu2021deepxde} aims to solve the PDE \eqref{general} with the prescribed boundary condition \eqref{general_boundary}. A lot of skepticism exists regarding the solution time and accuracy of PINNs in comparison to traditional numerical methods \cite{grossmann2024can}.
Nevertheless several attempts have been made to accelerate the training of PINNs \cite{markidis2021old}. In this section we show how a layerwise training process can lead to faster and more accurate solution in comparison to traditional-PINNs. The key is to adapt the neural network architecture with PINN-type loss terms. If one denotes the discretized equations  for  \eqref{general} arising out of a weak form as $\hat{\sG}^i(y(\bx_1),\ y(\bx_2),\dots y(\bx_n)),\ \ i=1,\dots n$, the physics loss function is defined as:

\begin{equation}
    \Phi_p({\boldsymbol{\theta}})=   \sum_{i=1}^n \LRs { \hat{\sG}^i(y({\boldsymbol{\theta}},\bx_1),\ y({\boldsymbol{\theta}},\bx_2),\dots y({\boldsymbol{\theta}},\bx_n))}^2,
    \label{discrete_loss}
\end{equation}
where, $\{ {\bx_1}\dots {\bx_n}\} $ is the set of collocation points in $\Omega$. Further, the neural network learns the boundary condition \eqref{general_boundary} by considering training data points on $\partial \Omega$.   It may be noted that, if one requires to update the network for a new parameter field $a\in \sA$ or a modified operator $\sG$, it is often unclear on the best layers to re-train \cite{subel2022explaining}. Recent work by Subel et al. \cite{subel2022explaining} shows that common wisdom guided transfer learning in machine learning
literature that involves retraining only the last few layers of the network is often not the best option  and that interpretable hidden layers are necessary to devise meaningful transfer learning strategy. Our approach (PIANN),  attempts to create interpretable hidden layers in a deep network with superior performance compared to traditional PINNs.
\begin{figure}[h!]      
  % \begin{subfigure}[b]{\textwidth}
  \hspace{-0.8 cm}
  \begin{tabular}{c}
      \begin{tabular}{c}
          \centering
          \includegraphics[scale=0.37]{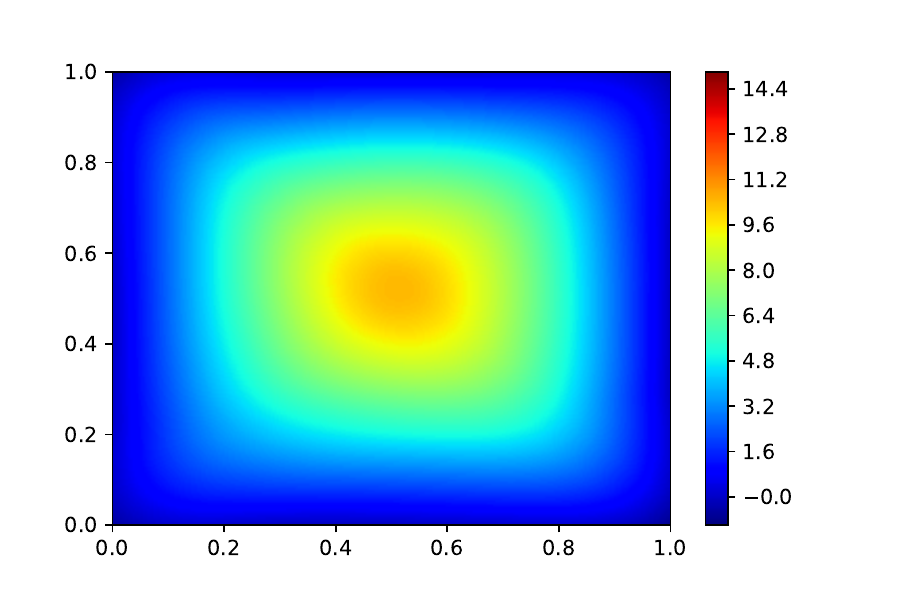}
      \end{tabular}
    \hspace{-1.0 cm}
      \begin{tabular}{c}
          \centering
          \includegraphics[scale=0.37]{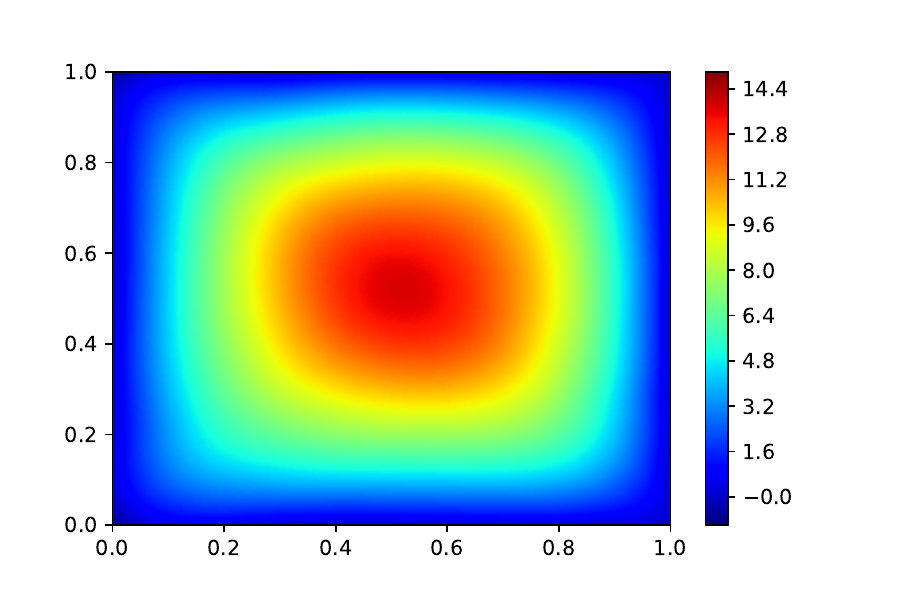}
          % \caption{Effect of noise on DI and Tikhonov solutions}
          % \figlab{some_good_name}
      \end{tabular}
    \hspace{-1.0 cm}
      \begin{tabular}{c}
          \centering
          \includegraphics[scale=0.37]{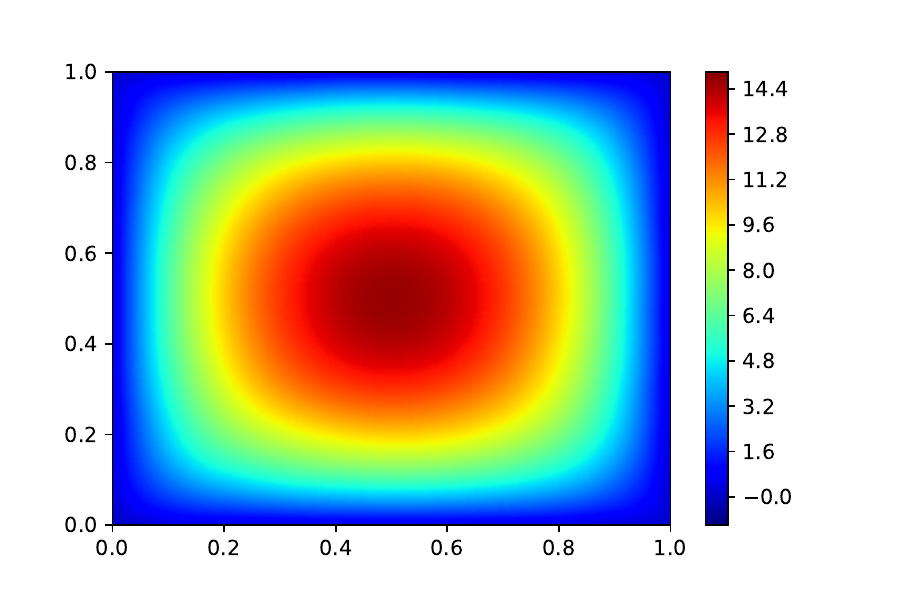}
          % \caption{Effect of noise on DI and Tikhonov solutions}
          % \figlab{some_good_name}
      \end{tabular}
  \end{tabular}
  \caption{Adaptively learning the Poisson's equation (Case a: symmetric boundary condition). Left to right: Solution after training  layer $\sN^{(2)}$;  Solution after training layer $\sN^{(3)}$; Solution after training layer $\sN^{(7)}$.}
  \label{Algo_I_simple_PDE}
\end{figure}

\begin{figure}[h!]      
  % \begin{subfigure}[b]{\textwidth}
  \hspace{-0.8 cm}
  \begin{tabular}{c}

      \begin{tabular}{c}

          \centering
          \includegraphics[scale=0.37]{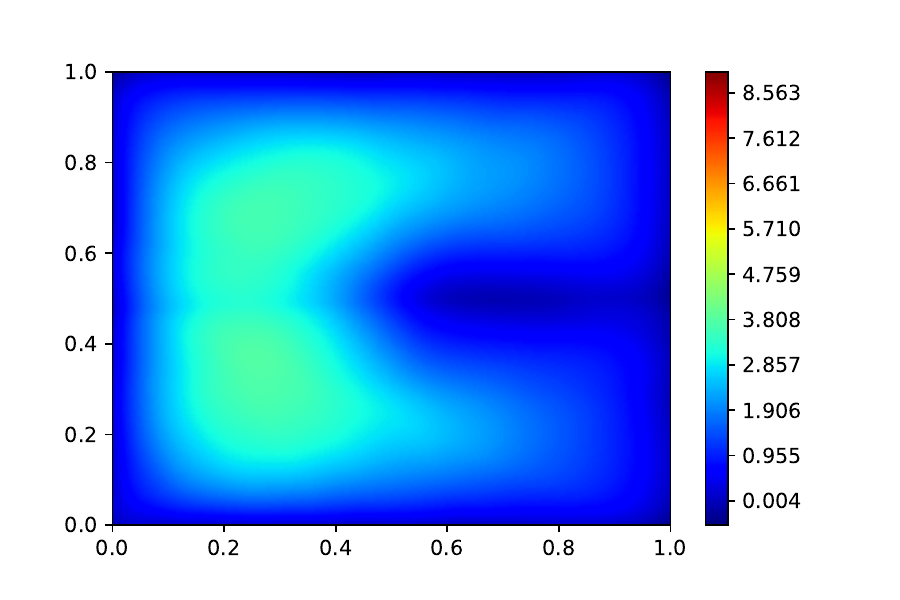}
          % \caption{Effect of noise on DI and Tikhonov solutions}
          % \figlab{some_other_good_name}

      \end{tabular}

    \hspace{-1.0 cm}

      \begin{tabular}{c}

          \centering
          \includegraphics[scale=0.37]{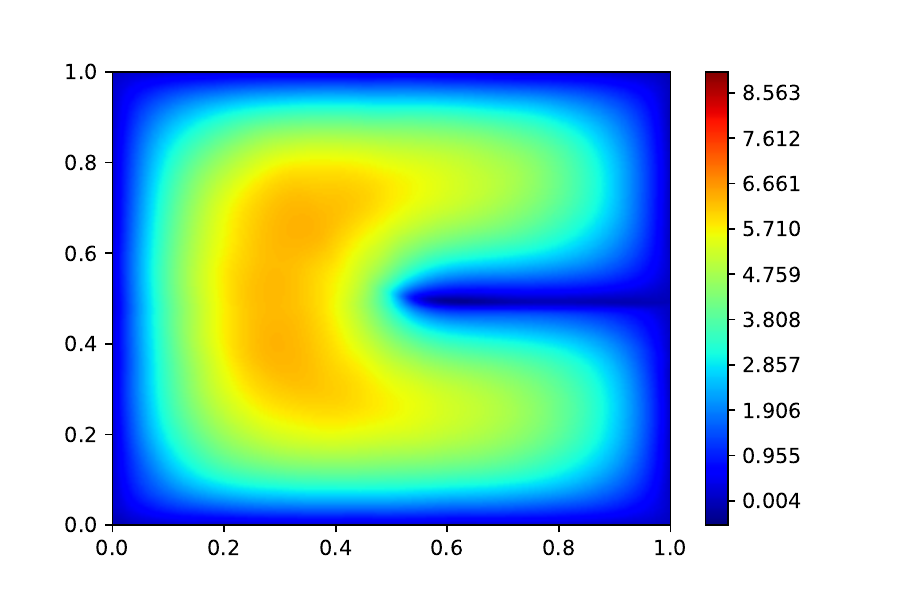}
          % \caption{Effect of noise on DI and Tikhonov solutions}
          % \figlab{some_good_name}

      \end{tabular}

    \hspace{-1.0 cm}

      \begin{tabular}{c}

          \centering
          \includegraphics[scale=0.37]{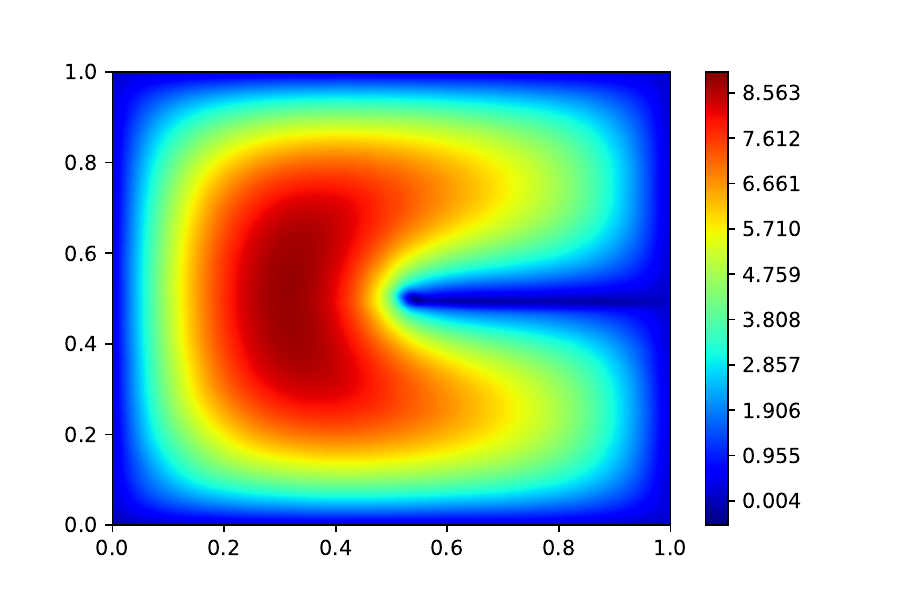}
          % \caption{Effect of noise on DI and Tikhonov solutions}
          % \figlab{some_good_name}

      \end{tabular}

  \end{tabular}
  \caption{Adaptively learning the  Poisson's equation (Case b: Slit in domain). Left to right: solutions after $\sN^{(2)}$ layers, $\sN^{(3)}$ layers, and  $\sN^{(5)}$ layers.}
  \label{Algo_I_complex_PDE}
  % \end{subfigure} \\
\end{figure}     

For the present approach, $\gamma=0$ (manifold regularization) since collocation points are uniformly distributed over the domain and the input data clusters does not exist.  The training data set contains $4000$ data points on the boundary. Further, $n=961$.

Inspired from the continuation method in  optimization \cite{shameli2018continuation}, we propose the layerwise continuation approach where each layer is tasked with learning a simple function that can be minimized efficiently, and gradually transform it to  more complicated cost function upon adding new layers. For achieving this, we choose $\tau^{(i+1)}>\tau^{(i)}$ ($\tau$ is the weight on the physics loss), and thereby also ensuring that a newly added layer is trainable  (Proposition \ref{TPOP}). Also, since $|\tau^{(i+1)}-\tau^{(i)}|\neq 0$ for any $i$, training saturation problem does not exist in this case (\cref{TPOP_cor}) and  \cref{Sequential} is not required. For demonstration of our approach, we consider prototype problems considered by Weinan et al. \cite{yu2018deep}.

\subsubsection{Learning the Poisson equation}

\label{simple_domain}

For demonstration, we consider solving the Poisson equation where the operator $\sG(\bx, \by, a)$ in  \eqref{general} is given as:  
\begin{align}
    \sG(\bx, \by, a)&=-\nabla . (a(x_1,\ x_2)\nabla y)-f(x_1,x_2)  \ \  \text{in} \ \ \Omega \subset\real^2,\\
    %\label{poisson_eq}
    y&=0  \ \ \text{in} \ \partial \Omega,
    \label{darc}
\end{align}
 with $a(x_1,\ x_2)=1$ and $f(x_1,\ x_2)=200$. Further, we experiments are conducted for two different domains:

\begin{enumerate}
    \item Case (a):  $\Omega=(0,1)\times(0,1)$;
    \item Case (b):  $\Omega=(0,1)\times(0,1)-(0.5,1)\times \{0.5\}$ \cite{yu2017deep}
\end{enumerate}

\begin{figure}[h!t!b!]      
  % \begin{subfigure}[b]{\textwidth}
  \hspace{-0.9 cm}
  \begin{tabular}{c}

      \begin{tabular}{c}

          \centering
          \includegraphics[scale=0.36]{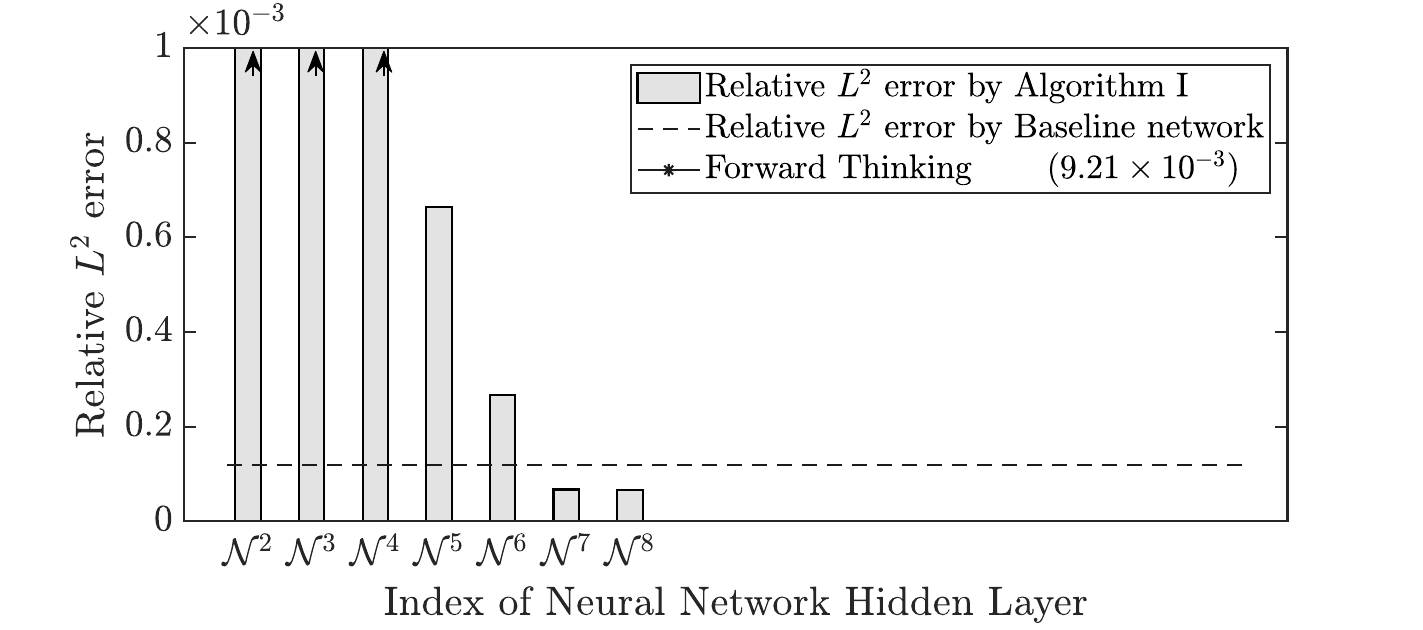}
          % \caption{Effect of noise on DI and Tikhonov solutions}
          % \figlab{some_other_good_name}

      \end{tabular}

    \hspace{-1.4 cm}

      \begin{tabular}{c}

          \centering
          \includegraphics[scale=0.36]{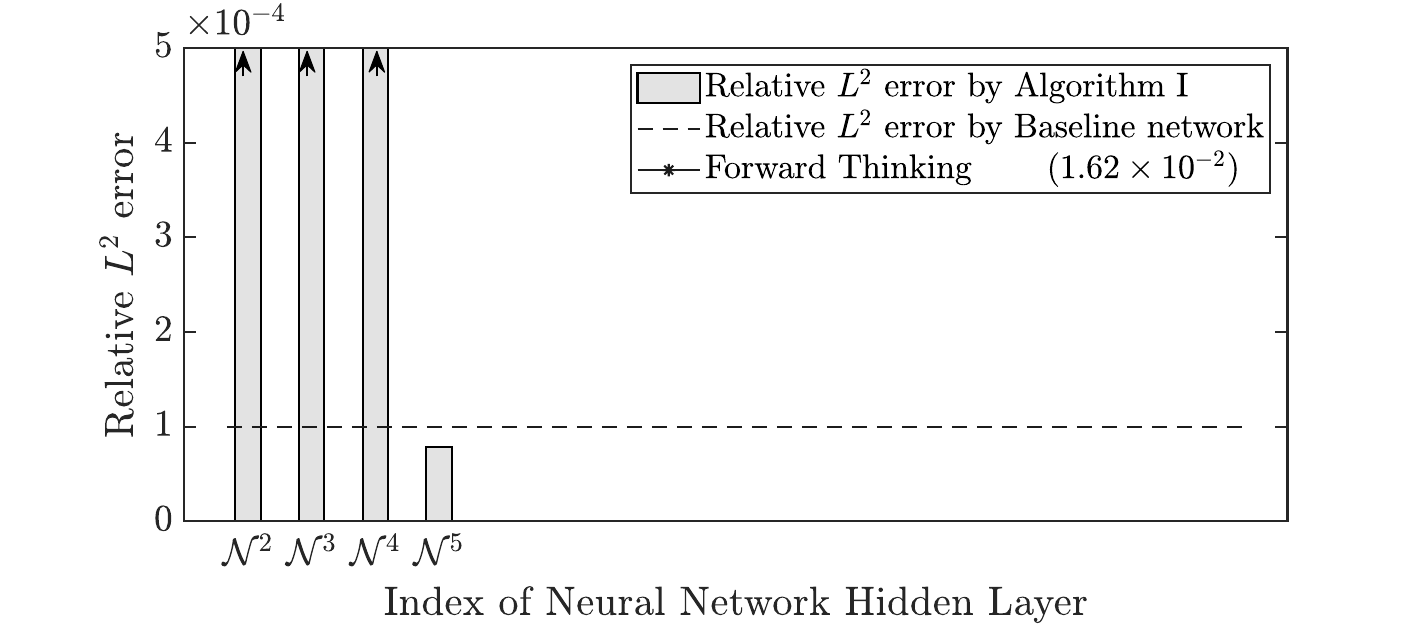}
          % \caption{Effect of noise on DI and Tikhonov solutions}
          % \figlab{some_good_name}

      \end{tabular}

  \end{tabular}
  \caption{%Summary of architecture adaptation results for 
  PIANN problem. Comparison of the proposed approach and others. Case a: Symmetric boundary condition (left figure), Case b:  Slit in domain (right figure). }

  \label{summary_PINNs_a}
  % \end{subfigure} \\
\end{figure} 
Additional details on training process are given in \cref{PIANN_a}. The evolution of PDE solution $y(\bx)$ as  more layers are added is shown in Figure \ref{Algo_I_simple_PDE} and \ref{Algo_I_complex_PDE}: with $5$ layers the solution looks identical to the finite element solution.  Note that learning the solution progressively by our proposed approach (trained for a total of $8000$ epochs) outperformed a baseline (traditional PINNs trained for $15000$ epochs) trained under similar settings as seen from Figure \ref{summary_PINNs_a}.   It can be seen that layerwise training procedure by Hettinger et al. \cite{hettinger2017forward} %(not shown in Figure \ref{summary_PINNs_a})
performed poorly in this case since by this method the network output is not preserved after adding a new layer. The summary of results and parameter efficiency is provided in \cref{PIANN_a}. Further, creating interpretable hidden layers in a deep network helps one in devising efficient transfer learning strategies and make informed decisions on which layers to re-train. An example is provided in \cref{transfer_learning} where we demonstrate the superiority of our approach over traditional transfer learning strategies.
%\krish{This Appendix is what I recently added to show how the method could be helpful in transfer learning.}

%(Appendix \ref{transfer_learning}). 

\subsection{Adaptive learning of  inverse maps from sparse data}
\label{Inverse_problem}

Inverse problems are usually ill-posed and  involves learning a map from low-dimensional space (observation space) to a high dimensional space (parameter space). This pose a challenge for learning the inverse map especially in the low data regime. Further, it is desirable for the inverse map to be well-posed where the solution's behaviour should change continuously with respect to input.  Therefore, it is clear that the notion of $\delta-$stability defined in  \cref{delta-def} is closely related to the concept of well-posedness. Therefore, the developed procedure serves as a natural candidate for this problem and we show how our approach can be applied to this regression task characterized by low availability of data.
In this section, we demonstrate the approach for conductivity coefficient field inversion in a 2D heat equation written as: 
\begin{equation}
\begin{aligned}
     -\nabla \cdot \LRp{e^u \nabla y} & = 20  \quad \text{in } \Omega = \LRs{0,1}^2,\\
    y & = 0 \quad \text{ on } \Gamma^{\text{ext}}, \\
    \textbf{n} \cdot \LRp{e^u \nabla y} & = 0 \quad \text{ on } \Gamma^{\text{root}},
\end{aligned}
\label{heat_equation_o}
\end{equation}
where $\textbf{n}$ is normal vector on the equivalent boundary surfaces. The parameter of interest is denoted as $u$ and the observables are 10 pointwise values of the heat state (measurement locations are shown in Figure \ref{Prob_inverse}), $\by$ at arbitrary location.  The training data is generated by drawing parameter samples as:
\begin{equation}
   \bu(\mb{x}) = \sum_{i =1 }^n \sqrt{\lambda_i}\  \mb{\phi}_i\  x_i  
\end{equation}
where $\lambda_i, \ \mb{\phi}_i$ is the eigen-pair of an exponential two point correlation function and $\textbf{x} = \LRp{x_1,\hdots, x_n}$ is drawn from standard Gaussian distribution \cite{constantine2016accelerating}. Note that $\textbf{x}$ represents the output of the network. For the present study, we choose the output dimension $n = 12$. The observation vector (input of the network) is  generated by solving \eqref{heat_equation_o}. $5 \%$ additive Gaussian noise is added to the observations to represent field condition.

\begin{table}[h!]
    \centering
    \caption{Relative error achieved by baseline network and other  methods}
  \label{tab:adaptation_inversion}
       	\begin{tabular}{|c | c | c | c | c|c|}
        \hline
    Training data  & Equivalent  &   Proposed method & Forward thinking \cite{hettinger2017forward}& Architecture \\ 
 size &  baseline &    &   & search (AS) \cite{li2020random} \\ \hline
    20 & $0.64$    &  0.585 \  & 1.41& 0.625  \\ 
    50   & 0.44\  &   0.42\    & 0.84& 0.427\\\hline
	\end{tabular} 
	\label{summ_inverse}
\end{table}
We consider experiments with two different training data size; $20$ and $50$. The validation data set consists of $20$ data points and the test set contains $500$ data points. Since the number of training data points is too low, forming clusters with the available dataset is impractical. 

\subsubsection{Forming artificial data clusters and training procedure}
\label{artific_clus}
In order to compute  the manifold regularization \eqref{manif_modif}, we generate artificial clusters by first assuming that each input data point (observation) lies on a  different cluster, i.e $\bx_m\in \sM_m\subset \real^{10}$. Each set $\sM_m$ is then populated with $100$ artificial perturbations of the input $\bx_m$ (assuming $1 \%$ additive Gaussian noise) and we further assume that each perturbed data is a valid observation data point.
\begin{figure}[h!]      
  % \begin{subfigure}[b]{\textwidth}
  
  \begin{tabular}{c}
\hspace{-0.5 cm}
      \begin{tabular}{c}

          \centering
          \includegraphics[scale=0.38]{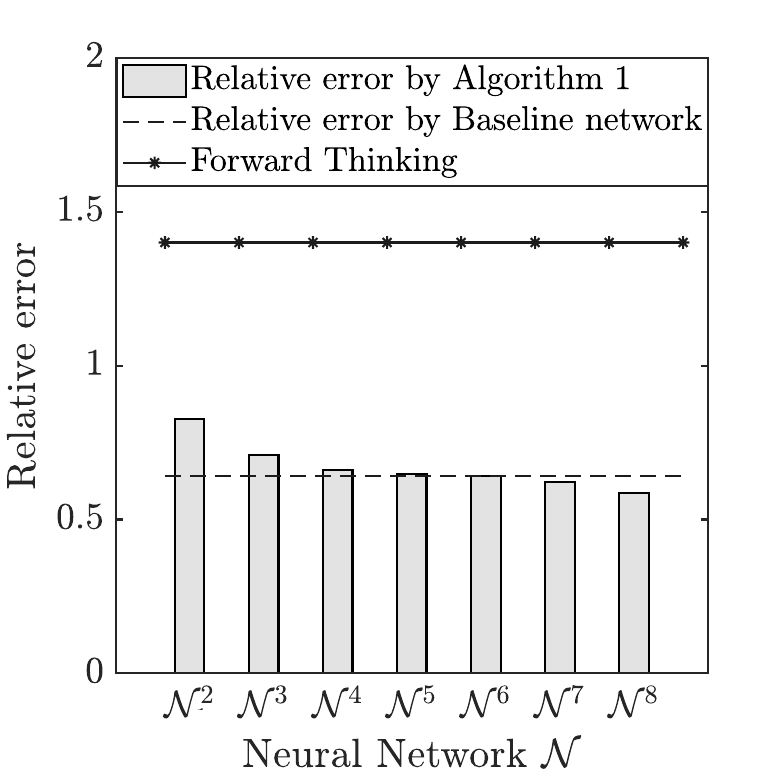}
          % \caption{Effect of noise on DI and Tikhonov solutions}
          % \figlab{some_other_good_name}

      \end{tabular}

     \hspace{-0.4 cm}

      \begin{tabular}{c}

          \centering
          \includegraphics[scale=0.35]{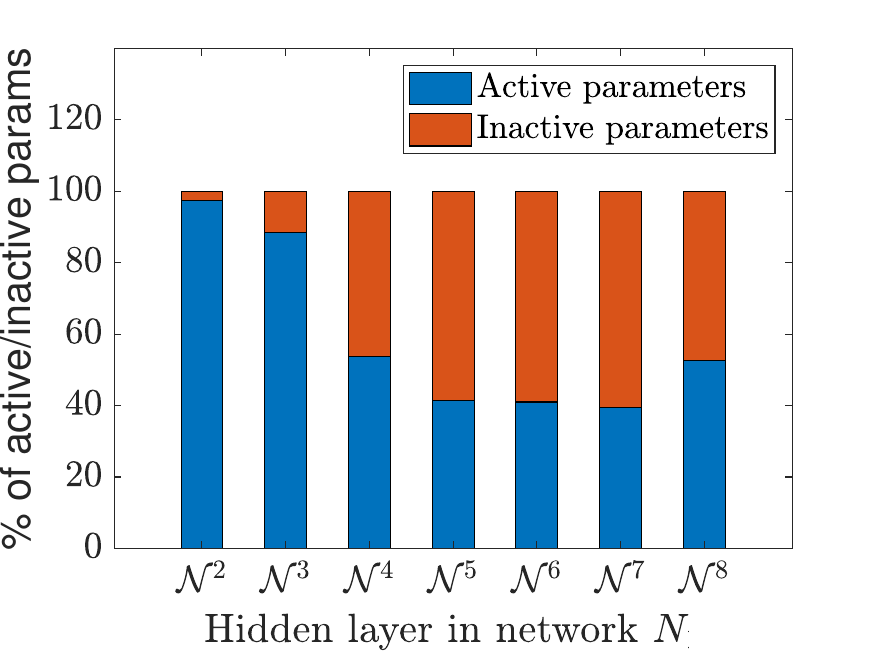}
          % \caption{Effect of noise on DI and Tikhonov solutions}
          % \figlab{some_good_name}

      \end{tabular}
  
      \hspace{-0.4 cm}
      
 \begin{tabular}{c}

          \centering
          \includegraphics[scale=0.35]{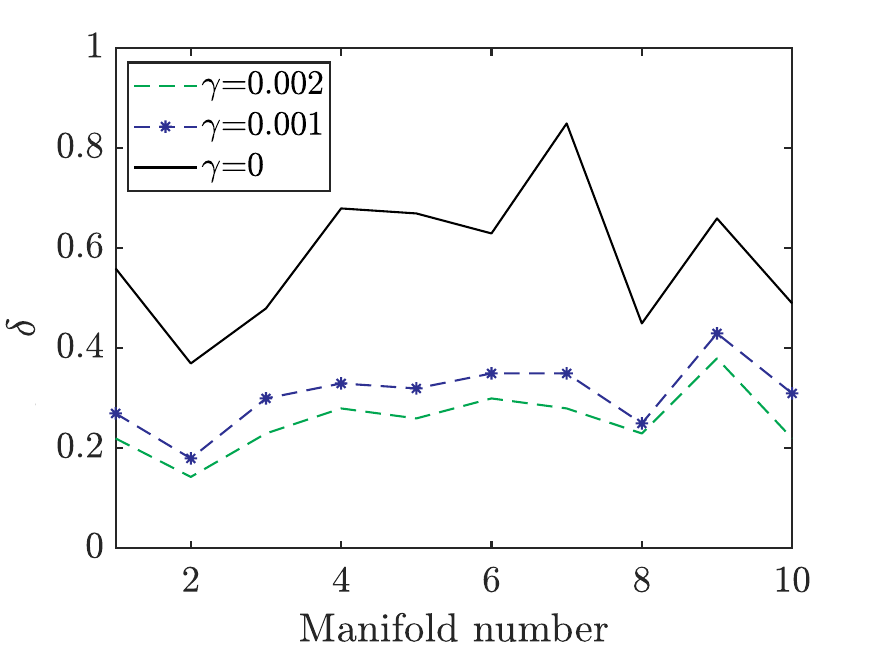}
          % \caption{Effect of noise on DI and Tikhonov solutions}
          % \figlab{some_good_name}

      \end{tabular}
  \end{tabular}
 \caption{Summary of adaptive inversion (trained with 20 data set).\\
 Left to right: Summary of   \cref{AlgoGreedyLayerwiseResNet}; Active and inactive parameters in each hidden layer;   Promoting  $\delta-$stability through manifold regularization. }
  \label{inverse_training}
  % \end{subfigure} \\
\end{figure} 

\begin{figure}[h!]      
  % \begin{subfigure}[b]{\textwidth}

  \begin{tabular}{c}

      \begin{tabular}{c}

          \centering
          \includegraphics[scale=0.37]{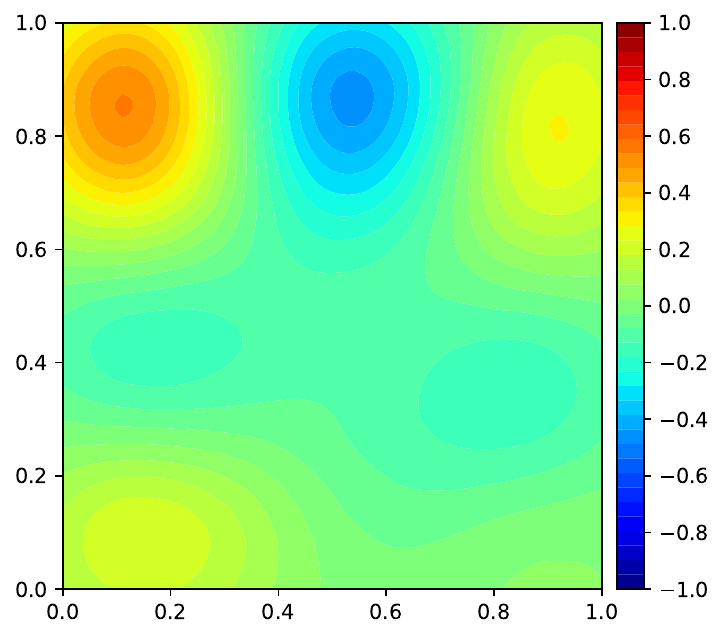}
          % \caption{Effect of noise on DI and Tikhonov solutions}
          % \figlab{some_other_good_name}

      \end{tabular}

    \hspace{0.2 cm}

      \begin{tabular}{c}

          \centering
          \includegraphics[scale=0.392]{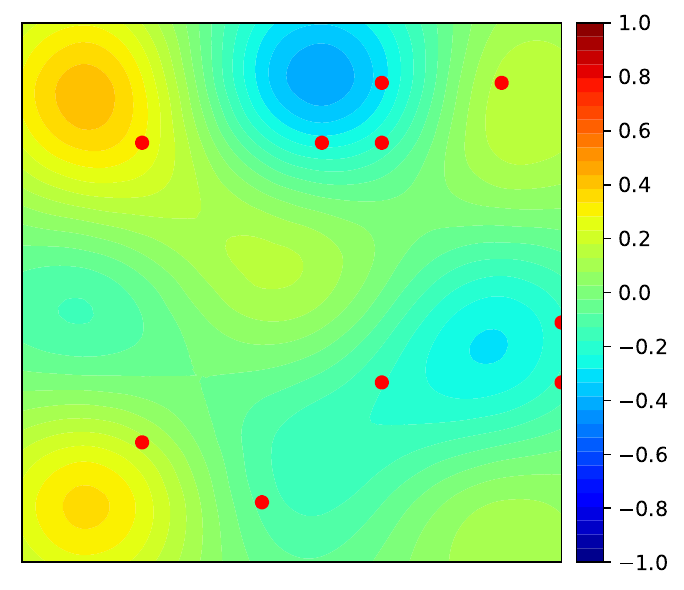}
          % \caption{Effect of noise on DI and Tikhonov solutions}
          % \figlab{some_good_name}

      \end{tabular}

  \hspace{0.2 cm}

      \begin{tabular}{c}

          \centering
          \includegraphics[scale=0.37]{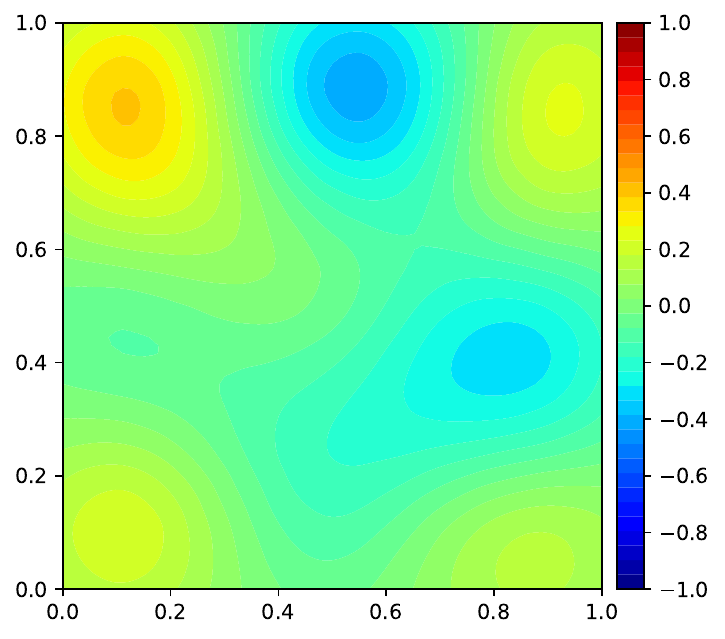}
          % \caption{Effect of noise on DI and Tikhonov solutions}
          % \figlab{some_good_name}

      \end{tabular}

  \end{tabular}
  % \end{subfigure} \\
  \caption{Predicted parameter field for a particular test observation sample using different methods (trained with 50 data sets).   First row left to right: Solution obtained by baseline network; Exact solution (red dot shows the location of sensors measuring the heat state);  Solution obtained by  \cref{AlgoGreedyLayerwiseResNet}.}
  \label{Prob_inverse}
\end{figure} 

In addition,  we consider increasing the manifold regularization weight, $\gamma^{(i+1)}=2\times \gamma^{(i)}$  while adding new layers  where stability is more strongly enforced in the later layers. This strategy is adopted  due to the small data-set size  of the problem where initial layers are weakly regularized to allow for maximum information transfer through the initial layers (note that we freeze the parameters of these layers later).  Further, for this problem setup  we have numerically observed that the data loss $\Phi_d$ in (\ref{loss_total}) only increases marginally  while optimizing the later layers (where $\gamma^{(i)}$ is too large) making this strategy feasible.  Most importantly, note that by Proposition \ref{TPOP}, training does not saturate  since $\gamma^{(i+1)}=2\times \gamma^{(i)}$ (satisfying the necessary condition for trainability) and hence one does not require \cref{Sequential}.

The relative error achieved by different methods is shown in Table \ref{summ_inverse} and Figure \ref{inverse_training} 
 which clearly shows that   \cref{AlgoGreedyLayerwiseResNet} outperforms traditional methods.  Further, Figure \ref{Prob_inverse} also shows that the parameter field predicted by the proposed method is better than the one produced by baseline network.

\begin{figure}[h!]      

  \begin{tabular}{c}

      \begin{tabular}{c}

          \centering
          \includegraphics[scale=0.35]{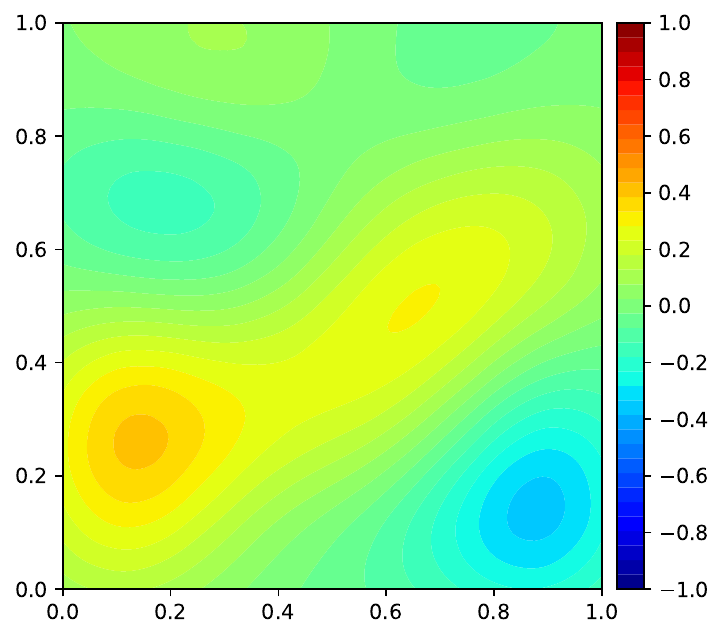}
          % \caption{Effect of noise on DI and Tikhonov solutions}
          % \figlab{some_other_good_name}

      \end{tabular}

    \hspace{0.2 cm}

      \begin{tabular}{c}

          \centering
          \includegraphics[scale=0.35]{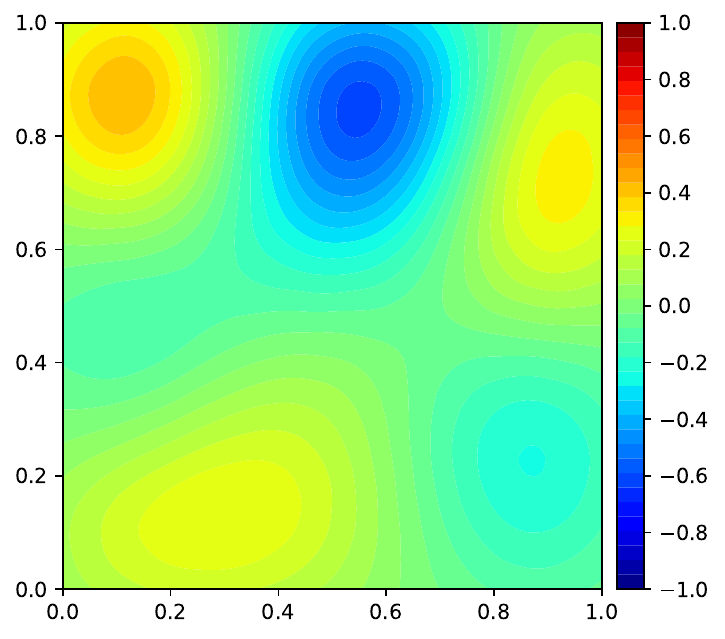}
          % \caption{Effect of noise on DI and Tikhonov solutions}
          % \figlab{some_good_name}

      \end{tabular}

    \hspace{0.2 cm}

      \begin{tabular}{c}

          \centering
          \includegraphics[scale=0.35]{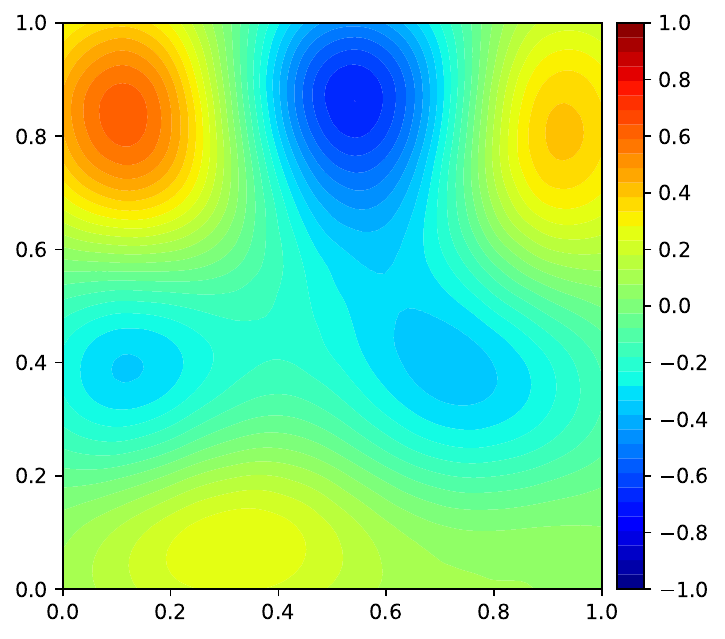}

      \end{tabular}

  \end{tabular}
\\   
  % \begin{subfigure}[b]{\textwidth}

  \begin{tabular}{c}

      \begin{tabular}{c}

          \centering
          \includegraphics[scale=0.35]{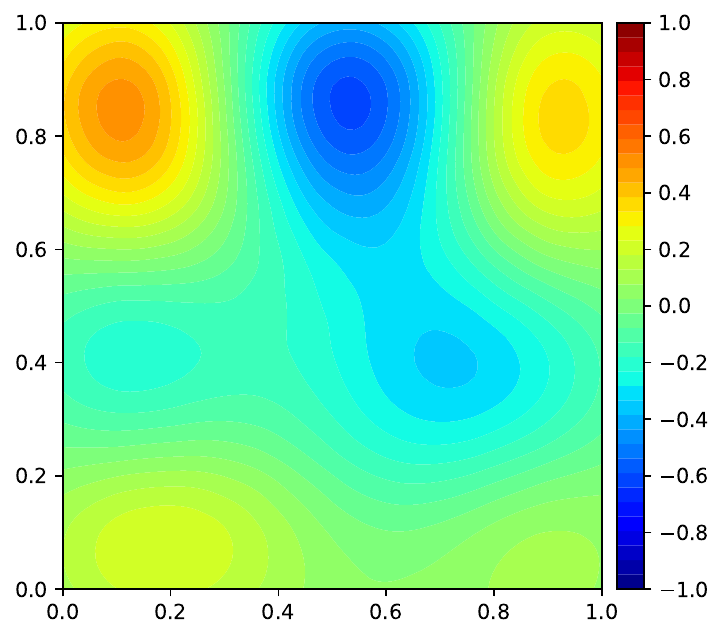}
          % \caption{Effect of noise on DI and Tikhonov solutions}
          % \figlab{some_other_good_name}

      \end{tabular}

    \hspace{0.2 cm}

      \begin{tabular}{c}

          \centering
          \includegraphics[scale=0.35]{Figures_modified/Inverse_problem/trained_parameter.pdf}
          % \caption{Effect of noise on DI and Tikhonov solutions}
          % \figlab{some_good_name}

      \end{tabular}

    \hspace{0.2 cm}

      \begin{tabular}{c}

          \centering
          \includegraphics[scale=0.35]{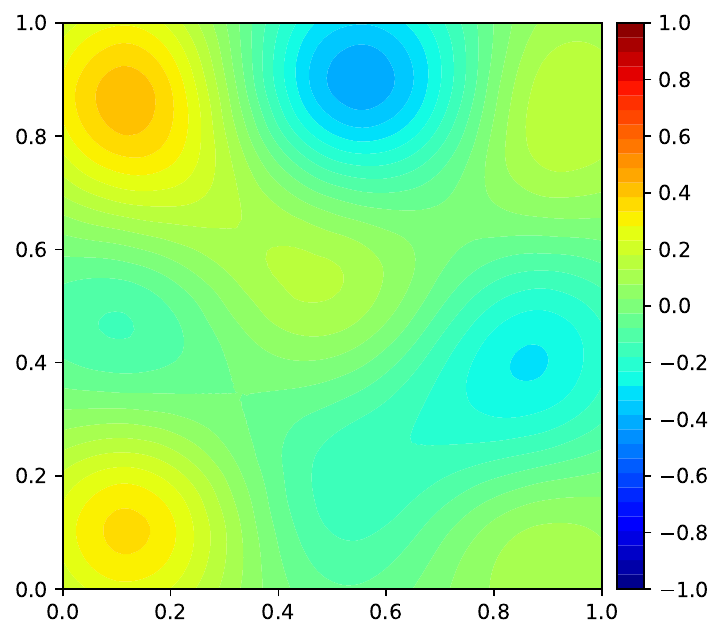}
          % \caption{Effect of noise on DI and Tikhonov solutions}
          % \figlab{some_good_name}

      \end{tabular}

  \end{tabular}
  \caption{Evolution of parameter field across the hidden layers for a particular test observation sample.  \\  First row left to right: Solution after training layer $\sN^{(3)}$;  Solution after training layer $\sN^{(5)}$; Solution after training layer $\sN^{(7)}$. Second row left to right: Solution after training layer $\sN^{(9)}$;  Solution after training layer $\sN^{(12)}$;  Exact solution.}
  \label{stability_pro}
\end{figure}

Since for this problem, data can be artificially generated from each $\sM_m$, one could investigate the effect of manifold regularization numerically through on a sampling based approach.  We study the nature of the stability function $\delta_j^{\sN}(\gamma)$ in (\ref{total_stab})   by training  three networks exactly in the same way but with different $\gamma^{(2)}$. Once the network is trained, $5000$ different points $\{\bx_1,\dots \bx_{5000} \}\in  \sM_m$ is artificially generated.
Then, based on  (\ref{total_stab}),  an upper bound corresponding to the set $\sM_m$ is approximately computed as:
\begin{equation}
 \delta^{\sN}_m\LRp{\gamma^{(2)}}=\max_{i,j} \Big ( \norm{\sN\LRp{\bx_j}-\sN\LRp{\bx_i}}_2\Big ),\quad \bx_i,\ \bx_j \in \sM_m
\end{equation}
 Figure \ref{inverse_training} shows that as $\gamma$ is increased, the numerically computed $ \delta^{\sN}_m\LRp{\gamma^{(2)}}$ decreases almost surely $\forall m$ and provides a feasible way for controlling stability via manifold regularization. In addition, the evolution of solution across the hidden layers for a particular test observation sample is shown in Figure \ref{stability_pro}. It is clear from  Figure \ref{stability_pro} that injecting stability in later layers allows the network to recover fine details in the parameter field when the baseline network fails to do so.

%Another advantage of this approach is 
%the interpretability of different hidden layers associated with the layerwise training procedure.  Note that for this problem, the initial layers tries to overfit on the training data whereas the later layers focus on injecting stability to the learnt inverse map. Therefore,  when new data is available (especially when unlabelled observation data is available)
%one may us the data to retrain the final few layers so as to further improve the stability.
%When labelled data is available,
%the two parts of the network may be trained independently (which is an efficient) while still maintaining the interpretability of hidden layers. 

\subsection{Image classification problem: MNIST data set}
\label{sub_image}
\begin{figure}[h!t!b!]      
  % \begin{subfigure}[b]{\textwidth}
  \hspace{-1  cm}
  \begin{tabular}{c}

      \begin{tabular}{c}

          \centering
          \includegraphics[scale=0.36]{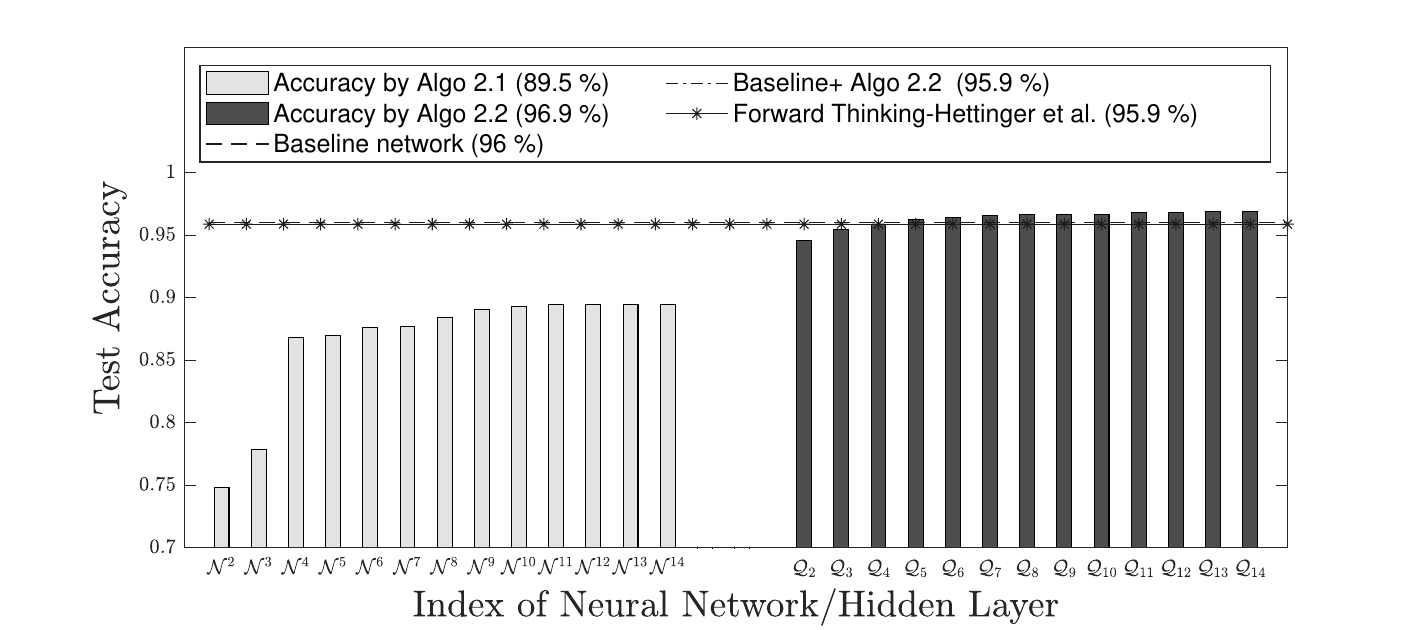}
          % \caption{Effect of noise on DI and Tikhonov solutions}
          % \figlab{some_other_good_name}

      \end{tabular}

    \hspace{-1 cm}

      \begin{tabular}{c}

          \centering
          \includegraphics[scale=0.36]{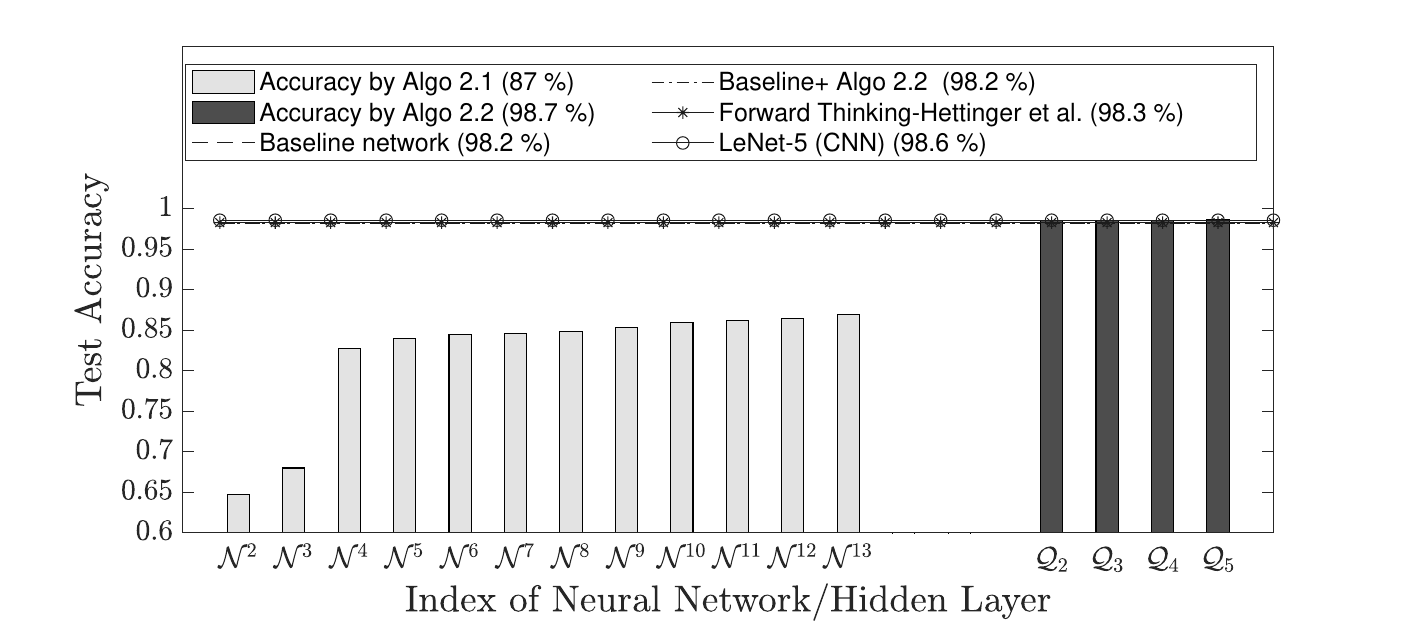}
          % \caption{Effect of noise on DI and Tikhonov solutions}
          % \figlab{some_good_name}

      \end{tabular}

  \end{tabular}
  \caption{%Summary of architecture adaptation results for 
  MNIST classification problem.
 Left figure is with 20 neurons in each hidden layer: our approach provides the best results with 96.9\% testing accuracy.  Right figure is with 500 neurons in each hidden layer: our approach provides the best results with 98.7\% testing accuracy.}
  \label{summary_PINNs_a_b}
  % \end{subfigure} \\
\end{figure}

\begin{figure}[h!]

  \begin{tabular}{c}
\hspace{-1.3 cm}
      \begin{tabular}{c}

          \centering
          \includegraphics[scale=0.055]{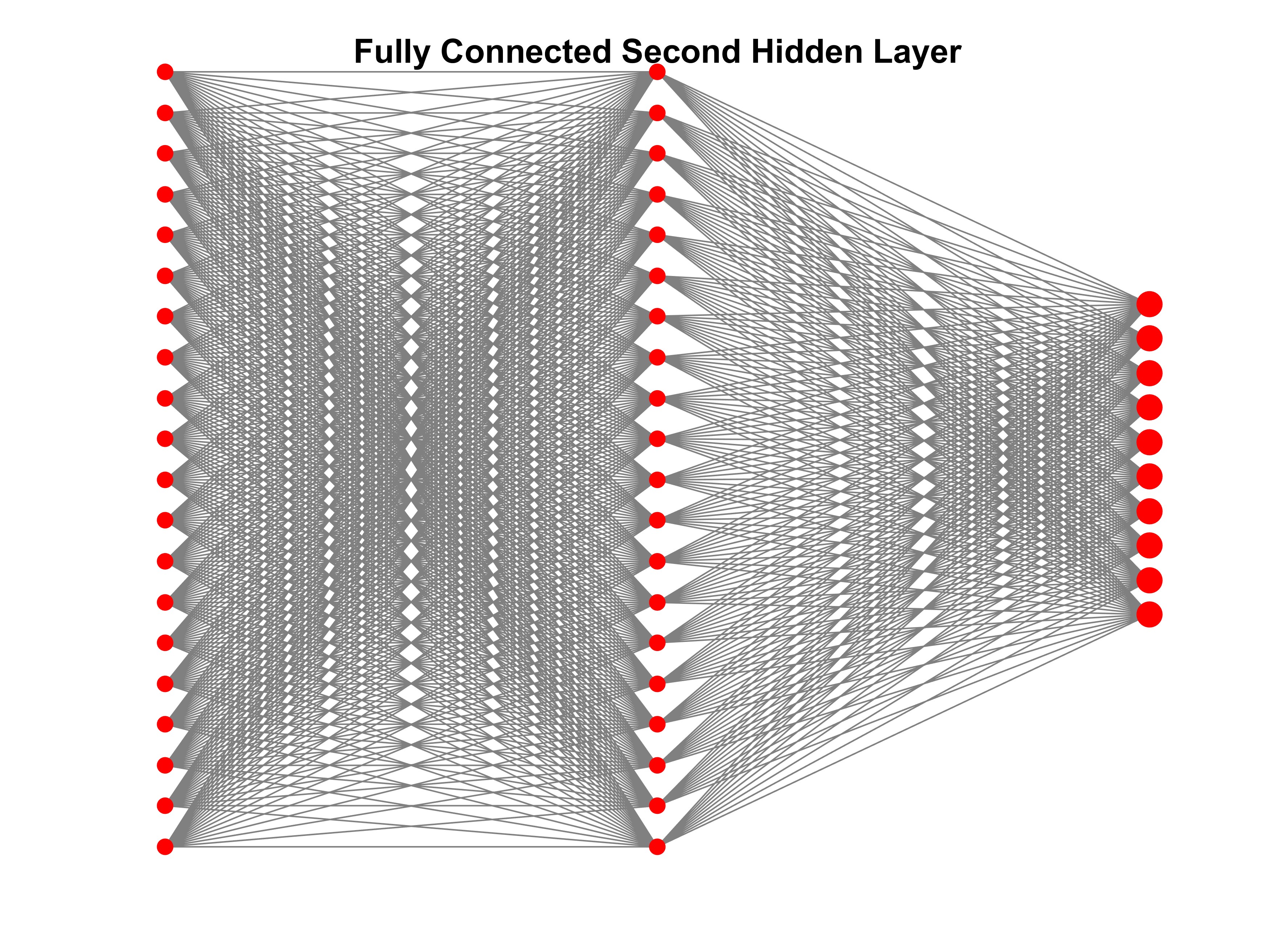}

      \end{tabular}
$\xRightarrow[\text{inactive connections}]{\text{Removing}}$
\hspace{-0.5 cm}
      \begin{tabular}{c}

          \centering
          \includegraphics[scale=0.055]{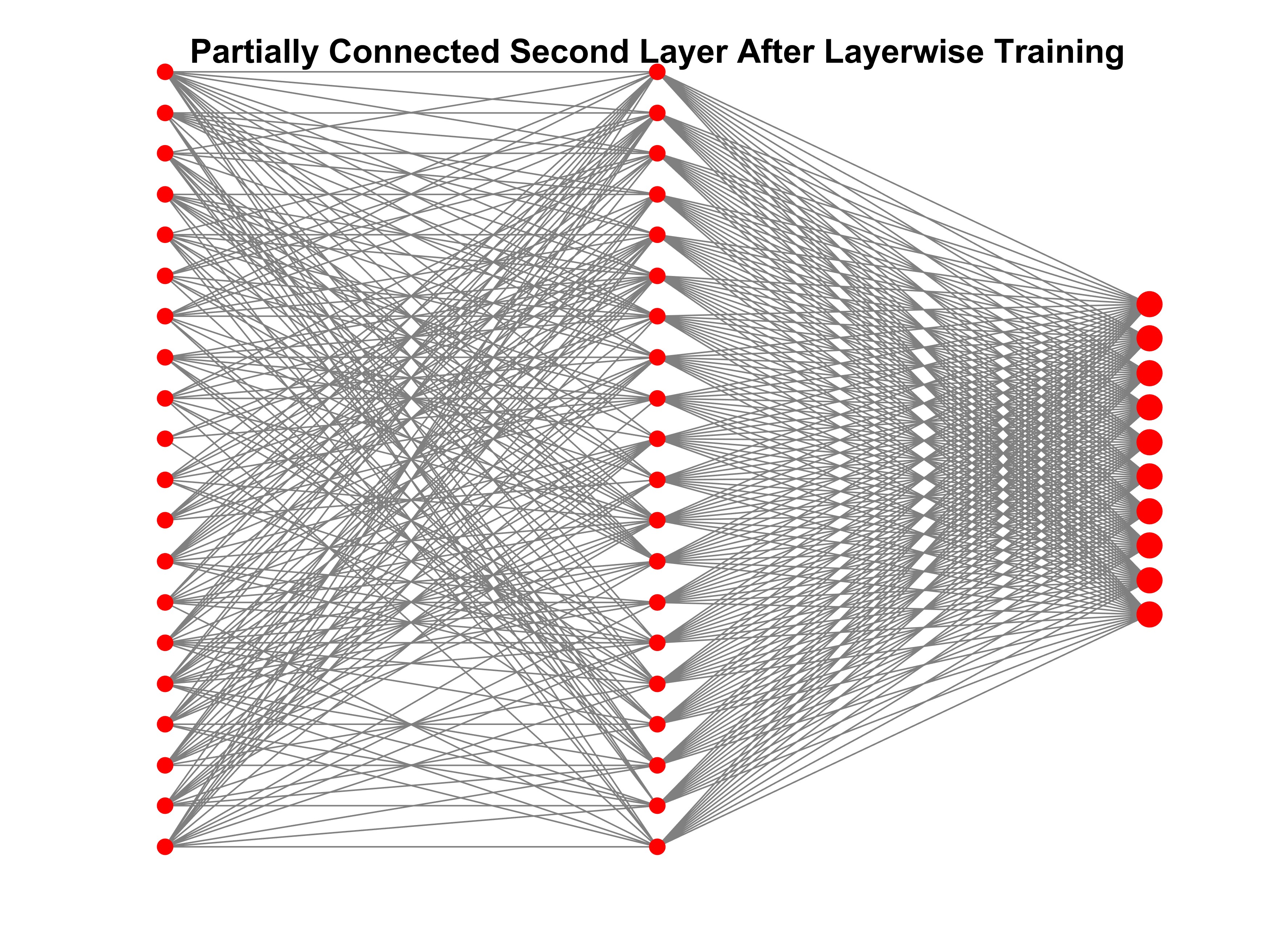}

      \end{tabular}\\ 

  \begin{tabular}{c}
\hspace{-1.3 cm}
          \centering
          \includegraphics[scale=0.055]{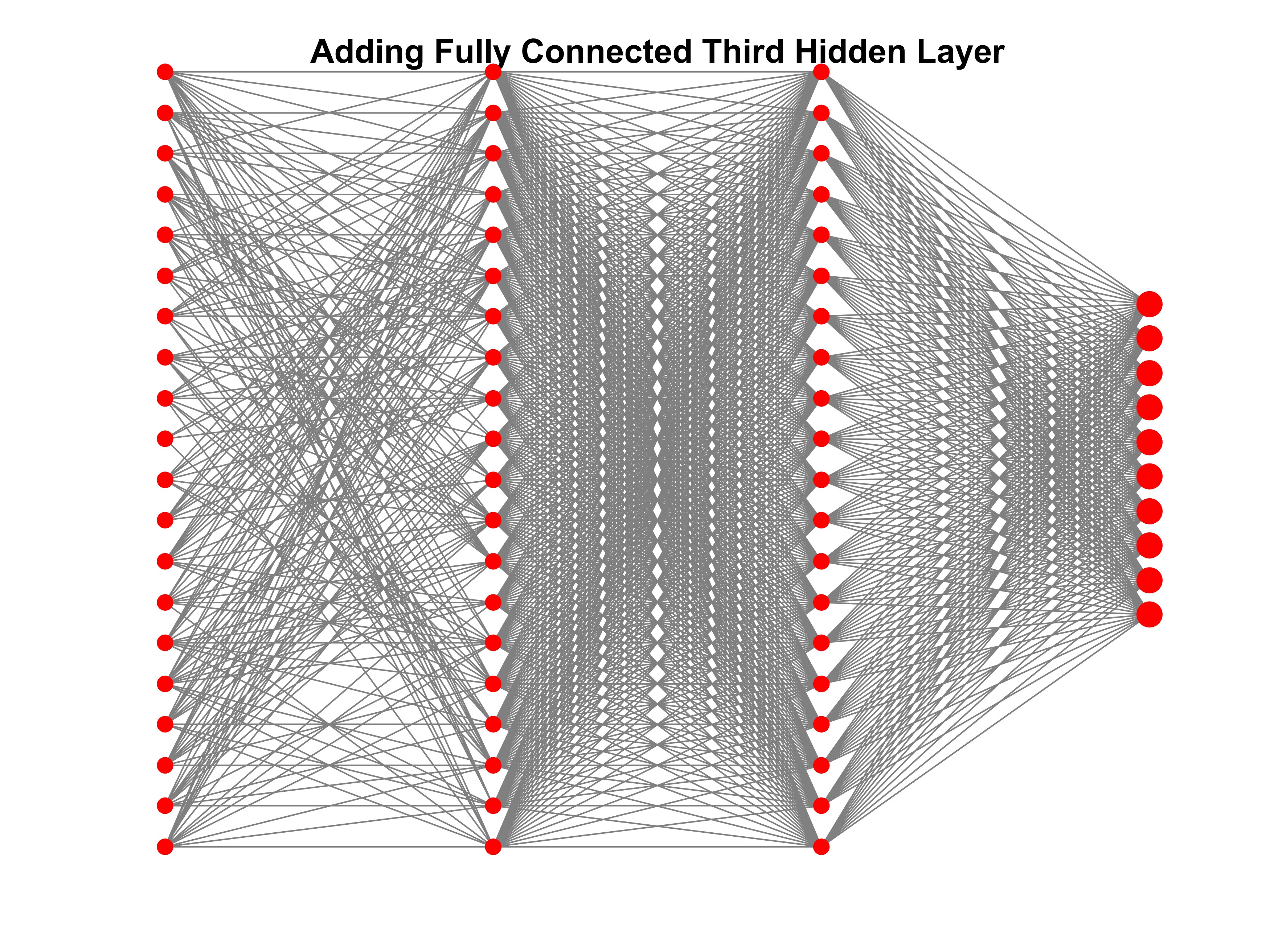}
          % \caption{Effect of noise on DI and Tikhonov solutions}
          % \figlab{some_other_good_name}

      \end{tabular}
$\xRightarrow[\text{inactive connections}]{\text{Removing}}$
\hspace{-0.5 cm}
      \begin{tabular}{c}

          \centering
          \includegraphics[scale=0.055]{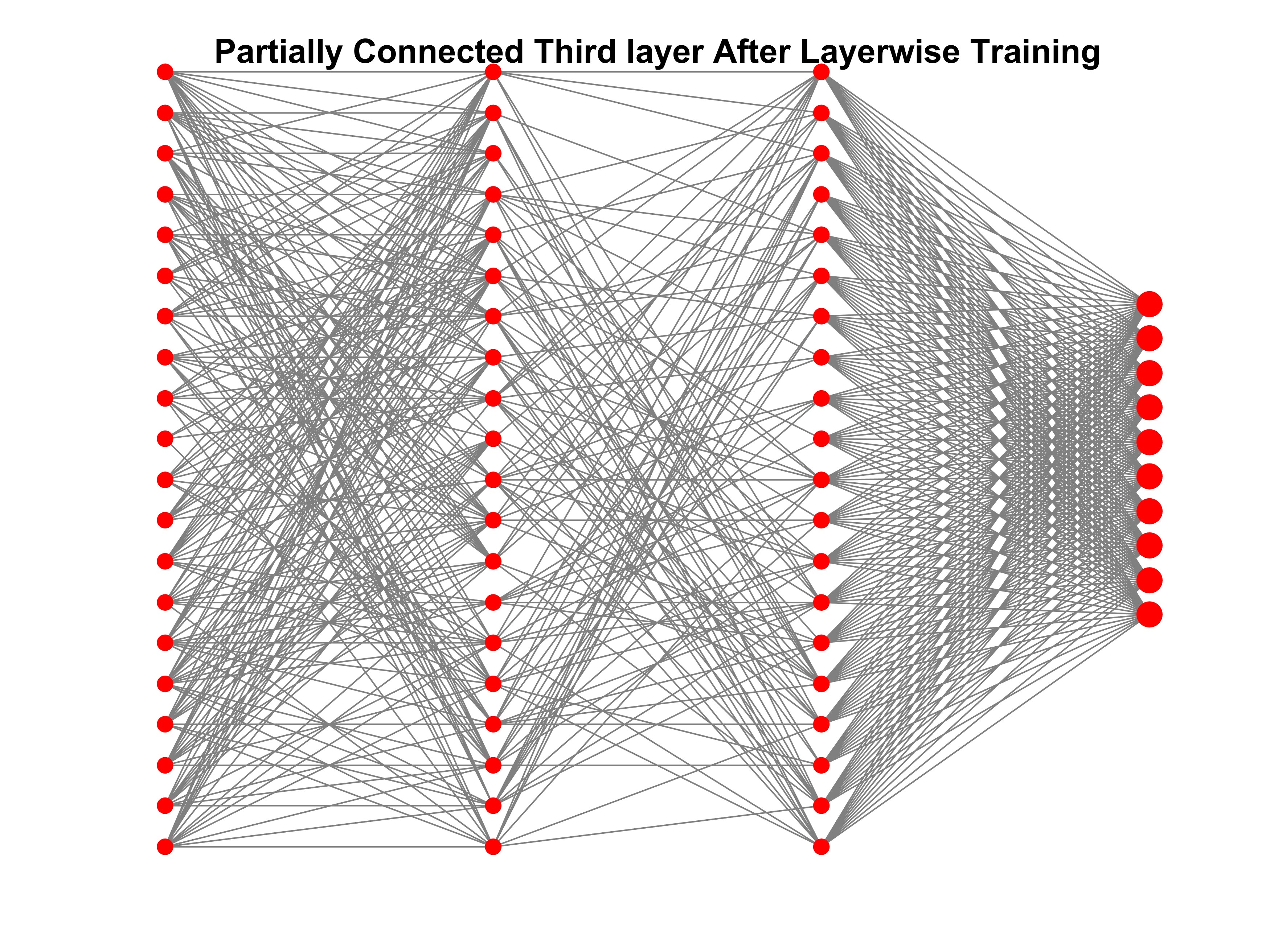}

      \end{tabular}\\

\hspace{0.0 cm}

      \begin{tabular}{c}
\hspace{-1.3 cm}
          \centering
          \includegraphics[scale=0.11]{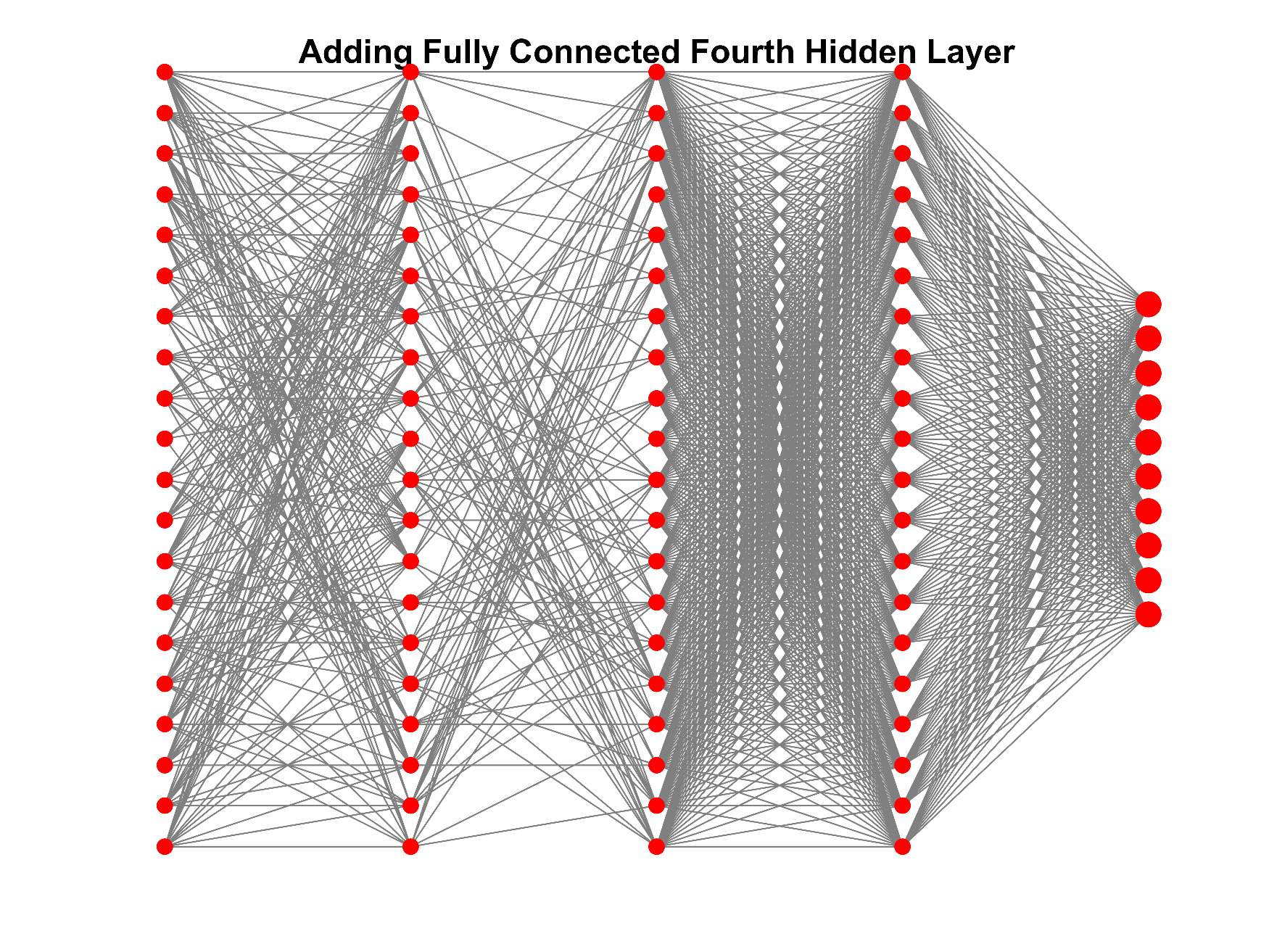}

      \end{tabular}
$\xRightarrow[\text{inactive connections}]{\text{Removing}}$
\hspace{-0.55 cm}
      \begin{tabular}{c}

          \centering
          \includegraphics[scale=0.055]{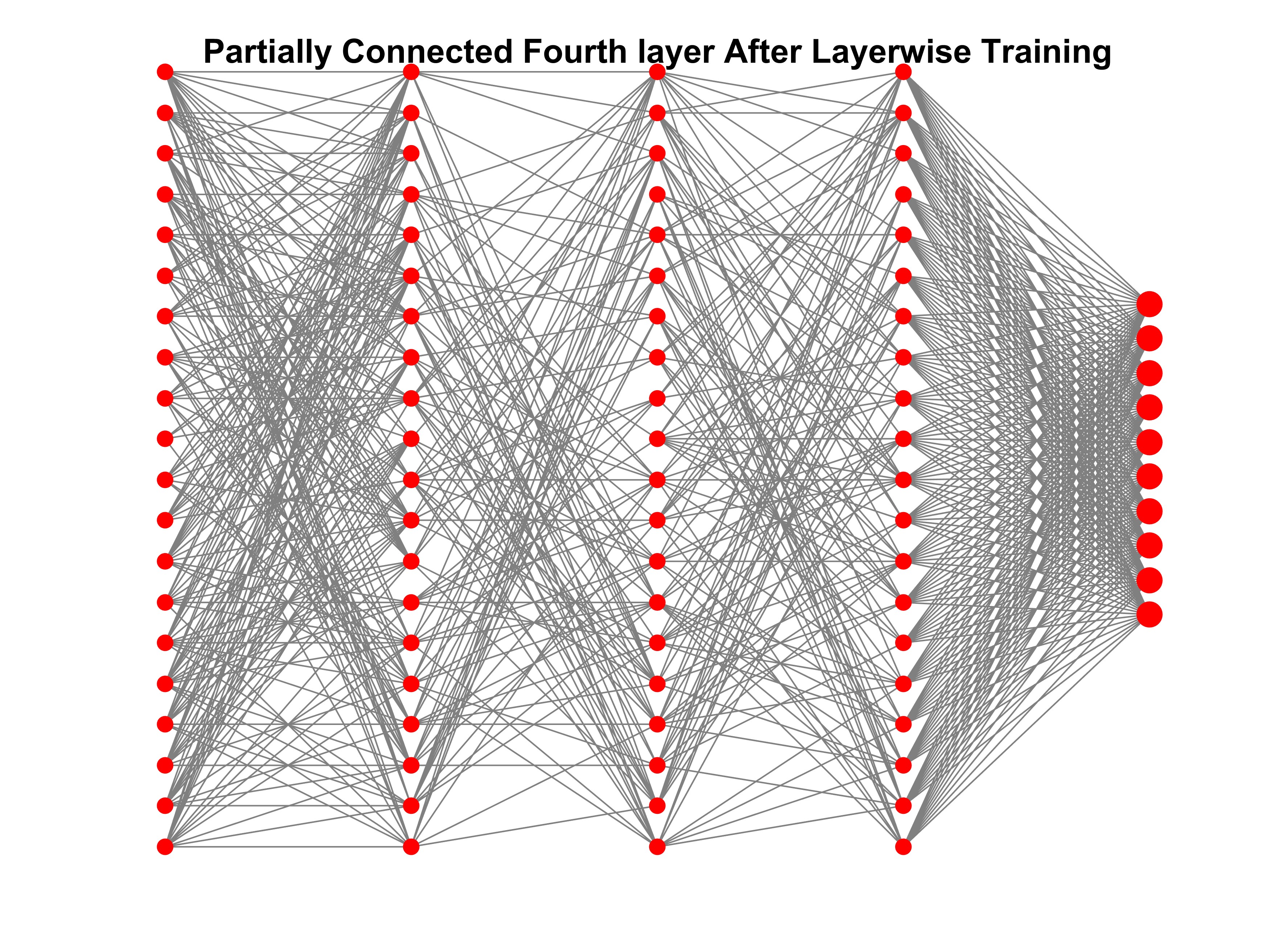}

      \end{tabular}
      
  \end{tabular}
  \caption{Schematic of  \cref{AlgoGreedyLayerwiseResNet}  demonstrated on MNIST dataset (20 neurons in each hidden layer) for the first few layers: Note that the input layer and ResNet connections are not shown here. }
  \label{MNIST_20_partially_connected}
  % \end{subfigure} \\
\end{figure}

Finally, we consider the MNIST handwritten digit classification problem using a fully connected residual neural network. It is noteworthy that, even though  MNIST problem is easier to solve with convolutional architectures, achieving high classification accuracy remains challenging with fully connected networks. In this section we examine how the design of $\epsilon-\delta$ stability promoting algorithm allows us to achieve a high accuracy using fully-connected networks.   We demonstrate the adaptation procedure for two different scenarios: a) Case with 20 neurons in each hidden layer; b) Case with 500 neurons in each hidden layer. Figure \ref{summary_PINNs_a_b} shows that the proposed methodology outperforms the baseline network and other methods by noticeable margins and even exhibiting performance on par to that of a LeNet-5 architecture (a basic CNN architecture). Further, Figure \ref{MNIST_20_partially_connected} shows the schematic of  \cref{AlgoGreedyLayerwiseResNet} applied to the MNIST dataset. Note that the inactive parameters can be removed after training each hidden layer as demonstrated in Figure \ref{MNIST_20_partially_connected} thereby leading to a partially connected  network. Additional details on experiments carried out is provided in \cref{MNIST_appendix}.

\section{Concluding remarks}
\label{conclusion}

In this paper, we presented a two-stage procedure for adaptive learning that promotes {\it{``robustness"}} and thereby generalizing well for a given data-set. The first stage (\cref{AlgoGreedyLayerwiseResNet}) is designed to  grow a neural network along the depth while promoting $\delta-$stability for each hidden layer through the use of manifold regularization.  In addition, non-important parameters are removed using a sparsity promoting regularization. We have proved that for certain hyperparameter settings of \cref{AlgoGreedyLayerwiseResNet}, one faces with the training saturation problem where the newly added layer does not learn ( \cref{TPOP_cor}). In order to further improve the prediction accuracy, we designed  \cref{Sequential} as a post-processing stage  where the main goal is to  promote robustness (\cref{robust_def}). Besides several theoretical results, numerical results on prototype regression and classification tasks, including solving forward and inverse problems governed by elliptic PDEs, suggest that the proposed approach can outperform an ad-hoc baseline and other adaptation strategies in terms of generalization error. The proposed PIANN is perhaps the first PINN approaches that is capable of adapting the network architecture with theoretical results on stabilities.

%One bottleneck with the approach is the 
%computational complexity of  $\mathcal{O}(n^2)$ associated with computing the manifold regularization term, where $n$ is the number of training data samples.

%However, this can be addressed using sub-sampling approaches \cite{li2019approximate}, thereby  substantially improving computational efficiency. 

\section*{Acknowledgement}

This research is partially funded by the National Science Foundation awards NSF-OAC-2212442, NSF-2108320, and NSF-CAREER-1845799; and by the Department of Energy award DE-SC0018147 and DE-SC0022211.
The authors would also like to thank Dr. Hwan Goh,  Dr. Jonathan Wittmer, Hai Van Nguyen, and Jau-Uei Chen for fruitful
discussions. The authors also acknowledge the Texas Advanced Computing Center (TACC)
at The University of Texas at Austin for providing HPC, visualization, database, or grid
resources that have contributed to some of the results reported within this paper. \href{http://www.overleaf.com}{URL:}  \url{http://www.tacc.utexas.edu.}

\bibliographystyle{siamplain}

%\bibliography{references}

\appendix

\begin{section}{Proof of non-asymptotic bound in equation \eqref{sample_influence}}{}
\label{conv_prop}
Let us first define the random variable $X_i$ as follows:
\begin{equation}
    \begin{aligned}
         X_i=& \int_{\sM_p^{(1)}}\norm{ \sN^{(2)}(\bx_i)-\sN^{(2)}(\by)}^2_2\ d\mu_p^{(1)}(\by), \quad \bx_i\sim \mu_p^{(1)}(\bx),\\
         \expect \LRs{X_i}= &\int_{\sM_p^{(1)}}\int_{\sM_p^{(1)}}\norm{ \sN^{(2)}(\bx)-\sN^{(2)}(\by)}^2_2\ d\mu_p^{(1)}(\bx)\ d\mu_p^{(1)}(\by).
    \end{aligned}
\end{equation}
Now using assumption \ref{gl_2} (continuous function $\sN^{(1)}$ maps the compact set $\sM_p^{(0)}$ to $\sM_p^{(1)}$ which is compact) it is easy to see that $X_i$ is a bounded random variable based on the Weierstrass theorem (Theorem 1.18.1 in \cite{oden2017applied}). Therefore, from Hoeffding's inequality \cite{vershynin2018high} we have:
\[ \p \LRp{\snor{\frac{1}{m_p}\sum_{i=1}^{m_p}X_i-\expect \LRs{X_i}}\geq \epsilon}\leq 2 \ \exp \LRp{-2c_1m_p\epsilon^2},
\]
where $c_1$ is a positive constant. Thus,
\begin{equation}
 \p \LRp{\expect \LRs{X_i}\leq \epsilon+\frac{1}{m_p}\sum_{i=1}^{m_p}X_i}\geq \p \LRp{\snor{\frac{1}{m_p}\sum_{i=1}^{m_p}X_i-\expect \LRs{X_i}}< \epsilon} \geq 1-2 \ \exp \LRp{-2c_1m_p\epsilon^2},
    \label{uno_2}
\end{equation}
 Now for any given $\bx_i\in \sM_p^{(1)}$ drawn from $\mu_p^{(1)}(\bx)$, let us define the following random variable:
\begin{equation}
    \begin{aligned}
         Y_j^i=& \norm{ \sN^{(2)}(\bx_i)-\sN^{(2)}(\by_j)}^2_2, \quad \by_j\sim \mu_p^{(1)}(\by),\\
         \expect \LRs{Y_j^i}= &\int_{\sM_p^{(1)}}\norm{ \sN^{(2)}(\bx_i)-\sN^{(2)}(\by)}^2_2\  d\mu_p^{(1)}(\by).
    \end{aligned}
\end{equation}
Again, from Hoeffding's inequality for random variable $Y_j^i$ we have:
\[ \p \LRp{\snor{\frac{1}{m_p}\sum_{j=1}^{m_p}Y_j^i-\expect \LRs{Y_j^i}}\geq \epsilon}\leq 2 \ \exp \LRp{-2c_2m_p\epsilon^2},\]
where $c_2$ is a positive constant. Therefore, we have:
\begin{equation}
   \p \LRp{\expect \LRs{Y_j^i}\leq \epsilon+\frac{1}{m_p}\sum_{j=1}^{m_p}Y_j^i}\geq \p \LRp{\snor{\frac{1}{m_p}\sum_{j=1}^{m_p}Y_j^i-\expect \LRs{Y_j^i}}\leq \epsilon} \geq 1-2 \ \exp \LRp{-2c_2m_p\epsilon^2}.
   \label{abo}
\end{equation}
Now considering the concentration inequality \eqref{abo} for all $i$ (summing up), and using the union bound we have:
\begin{equation}
     \p \LRp{\frac{1}{m_p}\sum_{i=1}^{m_p}\expect \LRs{Y_j^i}\leq \epsilon+\frac{1}{m_p^2}\sum_{i=1}^{m_p}\sum_{j=1}^{m_p}Y_j^i}\geq 1-2m_p \ \exp \LRp{-2c_2m_p\epsilon^2}.
     \label{union_bound_one}
\end{equation}
Now combining the result \eqref{uno_2} and \eqref{union_bound_one} and using the union bound, we have:
\[  \p \LRp{\expect \LRs{X_i}\leq \epsilon+\frac{1}{m_p^2}\sum_{i=1}^{m_p}\sum_{j=1}^{m_p}Y_j^i}\geq 1-2m_p \ \exp \LRp{-c_3m_p\epsilon^2}-2 \ \exp \LRp{-c_4m_p\epsilon^2}=1-\sE_p^{(2)}\LRp{m_p,\epsilon}\]
where, the constants $c_3,\ c_4$ depends on the cluster $C_p$ and the hidden layer. Therefore, we have the final required result:
\begin{align*}
   & \p\LRs{\expect_{\mu_p^{(1)}}\LRs{\norm{ \sN^{(2)}(\bx)-\sN^{(2)}(\by)}^2_2}\leq \epsilon+\frac{1}{m_p^2}\sum_{\bx_i,\bx_j\in C_p}\norm{\sN^{(2)}(\bx_i)-\sN^{(2)}(\by_j)}^2_2}\geq 1-\sE_p^{(2)}\LRp{m_p,\epsilon}.
    % &\expect_{\mu_p^{(1)}}\LRs{\norm{ \sN^{(2)}(\bx)-\sN^{(2)}(\by)}^2_2}= \int_{\sM_p^{(1)}}\int_{\sM_p^{(1)}}\norm{ \sN^{(2)}(\bx)-\sN^{(2)}(\by)}^2_2\ d\mu_p^{(1)}(\bx)\ d\mu_p^{(1)}(\by).
\end{align*}

\end{section}

\begin{section}{Proof of \cref{stabili_cor}}{}
\label{corol_algo}
A sketch of the proof is provided here. 
Since  $\sN$ is a composition of hidden layers (note that $(L+1)^{th}$ layer is the output layer), we have:
 \[ \sN\LRp{\bx}=\sN^{(L+1)}\circ \sR^{(L)} \circ \sN^{(L)}\circ \dots \circ \sR^{(1)} \circ \sN^{(1)}\LRp{\bx}.\]
Since the activation function employed is Lipschitz continuous, we have $\forall \bx,\ \bx' \in \sM_p^{(L-1)}$:
    \begin{align}
     \label{second_nm}
      &  \norm{\sN^{(L)}\LRp{\bx}-\sN^{(L)}\LRp{\bx'}}_2^2 \leq c^{(L)}\times \norm{\bx-\bx'}_2^2,\\
      \label{second_mn}
      &  \norm{\sN^{(L)}\LRp{\bx}-\sN^{(L)}\LRp{\bx'}}_2^2 \leq \tilde{\delta}_{p}\LRp{\kappa^{L-2}\times \gamma,\ \xi^{(L)}},
    \end{align}
where $c^{(L)}\geq 0$, and  (\ref{second_mn}) holds with probability at-least $\LRp{1-\xi^{(L)}-\sE_p^{(L)}\LRp{m_p,\ \xi^{(L)}\epsilon}}$ (from (\ref{stability_transfer})). Now multiplying \eqref{second_nm} and \eqref{second_mn}, we have:
%$\LRp{1-\xi^{(L)}-\sE_p^{(L)}\LRp{M,\ \xi^{(L)}\epsilon}}$:  \krish{align the equation}
\begin{equation}
    \begin{aligned}
&\norm{\sN^{(L)}\LRp{\bx}-\sN^{(L)}\LRp{\bx'}}_2^2\leq \\    
        & \sqrt{ c^{(L)}} \times \sqrt{\tilde{\delta}_{p}\LRp{\kappa^{L-2}\times \gamma,\ \xi^{(L)}}}\times \norm{\sR^{(L-1)}\LRp{\sN^{(L-1)}\LRp{\by}}-\sR^{(L-1)}\LRp{\sN^{(L-1)}\LRp{\by'}}}_2\\
        &\leq c^{(L)*} \times \sqrt{\tilde{\delta}_{p}\LRp{\kappa^{L-2}\times \gamma,\ \xi^{(L)}}}\times \norm{\sN^{(L-1)}\LRp{\by}-\sN^{(L-1)}\LRp{\by'}}_2,
    \end{aligned}
    \label{use}
\end{equation}
where $\by,\ \by'\in \sM_p^{(L-2)}$ and we have used the fact that $\sR^{(L-1)}$ is Lipschitz continuous. Similarly, applying (\ref{second_nm}) and (\ref{second_mn})  recursively  to each $\sN^{(L-1)},\dots \sN^{(2)}$ and using the union bound we have  $\forall \bx,\ \bx' \in \sM_p^{(0)} \subset \sM^{(0)}$:
\begin{equation}
    \begin{aligned}
        \norm{\sN\LRp{\bx}-\sN\LRp{\bx'}}_2^2 &\leq c \prod_{i=2}^L \LRp{\tilde{\delta}_{p}\LRp{\kappa^{i-2}\times \gamma,\ \xi^{(i)}}}^{\frac{1}{2^{(L-i+1)}}} \norm{\bx-\bx'}_2^2\\
        & \leq c\epsilon_p \prod_{i=2}^L \LRp{\tilde{\delta}_{p}\LRp{\kappa^{i-2}\times \gamma,\ \xi^{(i)}}}^{\frac{1}{2^{(L-i+1)}}}=\delta^{\sN}_p\LRp{\gamma,\ \xi^{(2)},\dots \xi^{(L)}},
    \end{aligned}
    \label{aympr}
\end{equation}
holds  with probability at-least $\LRp{1-\sum_{i=2}^L\xi^{(i)}-\sum_{i=2}^L\sE_p^{(i)}\LRp{m_p,\ \xi^{(i)}\epsilon}}$, and $\epsilon_p$ is defined as $\epsilon_p=\underset{\bx,\ \bx'\in \sM_p^{(0)}}{\sup} \norm{\bx-\bx'}_2$. Now in the limit $m_p\rightarrow  \infty$, \eqref{aympr} holds with probability at-least $\LRp{1-\sum_{i=2}^L\xi^{(i)}}$, where the stability function $\tilde{\delta}_{p}\LRp{\gamma,\ \xi^{(2)}}$ (for $i=2$) from \eqref{hoeffn} simplifies as:
\[ \tilde{\delta}_{p}\LRp{ \gamma,\ \xi^{(2)}}=\frac{1}{\xi^{(2)} \times m^2}\sum_{\bx_i,\bx_j\in C_p}\norm{\sN^{{(2)}}_{\boldsymbol{\theta}^*( \gamma)}(\bx_i)-\sN^{(2)}_{\boldsymbol{\theta}^*(\gamma)}(\bx_j)}^2_2,\]
and the subsequent functions $\tilde{\delta}_{p}\LRp{\kappa^{i-2}\times \gamma,\ \xi^{(i)}}$ for $i>2$ follows a similar pattern. Further note that from \eqref{prop_stab}  we also have $\tilde{\delta}_p\LRp{\gamma^u,\ \xi^{(2)}}=0$ and $\tilde{\delta}_p\LRp{0,\ \xi^{(2)}}>0,\ \forall p$.
Therefore, we have $\delta^{\sN}_p\LRp{\gamma^u,\ \xi^{(2)},\dots \xi^{(L)}}=0$ and $\delta^{\sN}_p\LRp{0,\ \xi^{(2)},\dots \xi^{(L)}}>0,\ \forall p$, thereby satisfying condition \ref{qw_1} (with $\epsilon=0$)  for any fixed $\xi^{(2)},\dots \xi^{(L)}$. In addition, since each $\tilde{\delta}_{p}\LRp{\kappa^{i-2}\times \gamma,\ \xi^{(i)}}$ is continuous with respect to $\gamma$ as proved in \cref{stability_prop},  $\delta^{\sN}_p\LRp{\gamma,\ \xi^{(2)},\dots \xi^{(L)}}$ is also continuous with respect to $\gamma$ thereby satisfying condition \ref{qw} of \cref{delta-def}.
\end{section}

\begin{section}{Gradients for layerwise training}{}
\label{back_propagation}
Consider training  the  $(i+1)^{th}$ layer.  Let us denote the different loss components for a particular $j^{th}$ training sample belonging to cluster $C_p$  as follows: 
\begin{equation}
\begin{aligned}
  % &\Phi_d\LRp{\bY^{(o)}_j}= \frac{1}{2}\norm{\bC_j-\bY^{(o)}_j}^2,\\
  &\Phi_d\LRp{\bY^{(o)}_j},\quad \mathrm{(data\ loss)}\\
   & \Phi_m^{(i+1)}\LRp{ \bY^{(i+1)}_j}= \frac{\gamma^{(i+1)}}{m_p^2}\times \sum_{m}\beta_p\norm{\bY^{(i+1)}_j-\bY^{(i+1)}_m}^2=\gamma^{(i+1)}\times \Phi_m\LRp{\bY^{(i+1)}_j},\\
   & \Phi_p^{(i+1)}\LRp{\bY^{(o)}_j}=  \tau^{(i+1)} \times \Phi_p\LRp{\bY^{(o)}_j},\quad \mathrm{(physics\ loss)}\\
 %&  \Phi_s^{(i+1)}(\bW^{(i+1)}, \ {\bb}^{(i+1)}) =\alpha^{(i+1)} \norm{\boldsymbol{\theta}}_1, \ \ \ \ \boldsymbol{\theta}=\{ \bW_{11},\  \bW_{12}, \dots  \ \bb_1,\ \bb_2, \dots\}.
\end{aligned}
\label{two-cost_a}
\end{equation}
where, $\bY^{(i+1)}_j$ represents the output of the $(i+1)^{th}$ layer for the $j^{th}$ training sample and $\bY^{(o)}_j$ represents the output of the network. Therefore,  training problem for the new $(i+1)^{th}$ layer can be formulated as:
\begin{equation}
\begin{aligned}
       \min_{{\bb}^{(i+1)}, \bW^{(i+1)}, \bW_{\mathrm{pred}}, {\bf b}_{\mathrm{pred}}} &\Phi_d\LRp{\bY^{(o)}_j}+\Phi_m^{(i+1)}\LRp{ \bY^{(i+1)}_j}+ \Phi_p^{(i+1)}\LRp{\bY^{(o)}_j}\\
     & \text{such that:} \\
       & \bY^{(i+1)}_j = \sR^{(i)}\LRp{\bY^{(i)}_j} + h^{(i+1)}\LRp{\bW^{(i+1)}\sR^{(i)}\LRp{\bY^{(i)}_j} + \bb^{(i+1)}},\\
       &  \bY^{(o)}_j =h_{\mathrm{pred}}\LRp{\bW_{\mathrm{pred}}\sR^{(i+1)}\LRp{\bY^{(i+1)}_j}+ {\bf b}_{\mathrm{pred}}}.
\end{aligned}
\label{two-cost}
\end{equation}
The Lagrangian for the above constrained optimization problem is defined as:
\begin{equation}
\begin{aligned}
    \lag_j=&   \Phi_d\LRp{\bY^{(o)}_j}+\Phi_m^{(i+1)}\LRp{ \bY^{(i+1)}_j}+\Phi_p^{(i+1)}\LRp{\bY^{(o)}_j}\\
    & +{(\Blambda_2)}^T\LRs{\bY^{(o)}_j -h_{\mathrm{pred}}\LRp{\bW_{\mathrm{pred}}\sR^{(i+1)}\LRp{\bY^{(i+1)}_j}+ {\bf b}_{\mathrm{pred}}}}\\
    &+{(\Blambda_1)}^T\LRs{\bY^{(i+1)}_j - \sR^{(i)}\LRp{\bY^{(i)}_j} - h^{(i+1)}\LRp{\bW^{(i+1)}\sR^{(i)}\LRp{\bY^{(i)}_j} + \bb^{(i+1)}}}.
\end{aligned}
\label{lagrangian}
\end{equation}
The first order optimality conditions are as follows:
\begin{equation}
   \pp{\lag_j}{\bY^{(i+1)}_j}=0, \hspace{0.15 cm}\pp{\lag_j}{\bY^{(o)}_j}=0,\hspace{0.15 cm} \pp{\lag_j}{\bW^{(i+1)}}=0,\hspace{0.15 cm}\pp{\lag_j}{\bb^{(i+1)}}=0, \hspace{0.15 cm} \pp{\lag_j}{\bW_{\mathrm{pred}}}=0,\hspace{0.15 cm}\pp{\lag_j}{ {\bf b}_{\mathrm{pred}}}=0.
\end{equation}
\begin{equation*}
 \pp{\lag_j}{\bY^{(i+1)}_j}=0 \implies {\Blambda_1}=\bR^{(i+1)}\bW_{\mathrm{pred}}^T\LRs{ h_{\mathrm{pred}}'\LRp{\bW_{\mathrm{pred}}\sR^{(i+1)}\LRp{\bY^{(i+1)}_j}+ {\bf b}_{\mathrm{pred}}} \circ{\Blambda_2}}-\pp{ \Phi_m^{(i+1)}}{\bY^{(i+1)}_j},
\end{equation*}
where  $\sR^{(i+1)}\LRp{\bY^{(i+1)}_j}=\bR^{(i+1)}\ \bY^{(i+1)}_j+\bb_r$ such that $\bR^{(i+1)}$ is a diagonal matrix and $\bb_r$ is a constant vector. Similarly, other optimality conditions can be derived as follows:
\begin{equation*}
% \pp{\lag_j}{\bY^{(o)}_j}=0 \implies {\Blambda_2}=\LRp{\bC_j-\bY^{(o)}_j}-\pp{\Phi_p^{(i+1)}}{\bY^{(o)}_j},
\pp{\lag_j}{\bY^{(o)}_j}=0 \implies {\Blambda_2}=-\LRp{\pp{\Phi_d}{\bY^{(o)}_j}+\pp{\Phi_p^{(i+1)}}{\bY^{(o)}_j}},
\end{equation*}
\begin{equation}
\begin{aligned}
\pp{\lag_j}{\bW^{(i+1)}} =-\LRs{{\Blambda_1}\circ {\LRp{h^{(i+1)}}'\LRp{\bW^{(i+1)}\sR^{(i)}\LRp{\bY^{(i)}_j} + \bb^{(i+1)}}}}{\LRp{\sR^{(i)}\LRp{\bY^{(i)}_j}}}^T,
\end{aligned}
\end{equation}
\begin{equation}
\begin{aligned}
&\pp{\lag_j}{\bb^{(i+1)}}=-{{\Blambda_1}\circ {\LRp{h^{(i+1)}}'\LRp{\bW^{(i+1)}\sR^{(i)}\LRp{\bY^{(i)}_j} + \bb^{(i+1)}}}},
\end{aligned}
\label{bias_gradient}
\end{equation}

%\begin{equation}
%\begin{aligned}
%&\pp{\lag}{\bb^{(i+1)}}=-\LRs{{\Blambda_1}\circ %\LRp{{h^{(i+1)}}'(\bY^{(i)}_j\bW^{(i+1)} + %\bb^{(i+1)})}}+\pp{\Phi_s^{(i+1)}}{\bb^{(i+1)}}\\
%&=-\LRs{\LRp{\bW_{\mathrm{pred}}^T\LRs{ %h_{\mathrm{pred}}'(\bY^{(i+1)}_j\bW_{\mathrm{pred}}+ {\bf %b}_{\mathrm{pred}}) \circ{\Blambda_2}}-\pp{ \Phi_m^{(i+1)}}{\bY^{(i+1)}_j}}\circ \LRp{{h^{(i+1)}}'(\bY^{(i)}_j\bW^{(i+1)} + \bb^{(i+1)})}}\\
%&\hspace{10 cm} +\pp{\Phi_s^{(i+1)}}{\bb^{(i+1)}}
%\end{aligned}
%\end{equation}

\begin{equation}
\begin{aligned}
 \pp{\lag_j}{\bW_{\mathrm{pred}}}=-\LRs{{\Blambda_2}\circ {h_{\mathrm{pred}}'\LRp{\bW_{\mathrm{pred}}\sR^{(i+1)}\LRp{\bY^{(i+1)}_j}+ {\bf b}_{\mathrm{pred}}}}}{\LRp{\sR^{(i+1)}\LRp{\bY^{(i+1)}_j}}}^T,\\
%&=-\LRs{\LRp{\LRp{\bC_j-\bY^{(o)}_j}-\pp{\Phi_p^{(i+1)}}{\bY^{(o)}_j}}\circ {h_{\mathrm{pred}}'\LRp{\bY^{(i+1)}_j\bW_{\mathrm{pred}}+ {\bf b}_{\mathrm{pred}}}}}{\bY^{(i+1)}_j}^T,
\end{aligned}
\end{equation}
\begin{equation}
\begin{aligned}
\hspace{0.5 cm}\pp{\lag_j}{ {\bf b}_{\mathrm{pred}}}=-\LRs{{\Blambda_2}\circ \LRp{h_{\mathrm{pred}}'\LRp{\bW_{\mathrm{pred}}\sR^{(i+1)}\LRp{\bY^{(i+1)}_j}+ {\bf b}_{\mathrm{pred}}}}}.\\
%&=-\LRs{\LRp{\LRp{\bC_j-\bY^{(o)}_j}-\pp{\Phi_p^{(i+1)}}{\bY^{(o)}_j}}\circ {h_{\mathrm{pred}}'\LRp{\bY^{(i+1)}_j\bW_{\mathrm{pred}}+ {\bf b}_{\mathrm{pred}}}}}.
\end{aligned}
\end{equation}

\end{section}

\begin{section}{Proof of Proposition \ref{Lipschitz} }
    \label{proof_lips}
    \begin{proof}
First  note that $\sQ_i$ is Lipschitz continuous  when employing Lipschitz continuous activation functions. Therefore, we have $\forall \bx,\ \bx'\in \sM_p^{(0)}$:
\begin{equation}
    \begin{aligned}
        \norm{\sQ\LRp{\bx}-\sQ\LRp{\bx'}}_2&=\norm{\sQ_2(\bx)+\dots \sQ_{r-1}(\bx)-\LRp{\sQ_2(\bx')+\dots \sQ_{r-1}(\bx')}}_2\\
       & \leq \norm{\sQ_2\LRp{\bx}-\sQ_2\LRp{\bx'}}_2+\dots \norm{\sQ_{r-1}\LRp{\bx}-\sQ_{r-1}\LRp{\bx'}}_{2}\\
       & \leq \sL^{(2)}\LRp{\sM_p^{(0)}} \times \epsilon_p+\dots \sL^{(r-1)}\LRp{\sM_p^{(0)}} \times \epsilon_p, 
    \end{aligned}
    \label{comb_1}
\end{equation}
where $\sL^{(i)}(\sM_p^{(0)})$ is the local Lipschitz constant of $\sQ_i$ with respect to $\sM_p^{(0)}$, and $\epsilon_p=\underset{\bx,\ \bx'\in \sM_p^{(0)}}{\sup} \norm{\bx-\bx'}_2$.
Now note  that $\sL^{(i)}(\sM_p^{(0)})\leq \sL^{(i)}(\real^S)$ since $\sM_p^{(0)}\subset \sM^{(0)} \subset \real^S$. Further, noting the fact that the network $\sQ_i$ is a composition of Lipschitz functions, we have for each $i$:
%the following upper bound for $\sL^{(i)}$:
%\[ \sL^{(i)}(\real^S) \leq \LRp{\prod_{k=1}^3 \norm{\bW^{(k)}_{{i}}}_2}.\]
\begin{equation}
    \begin{aligned}
       &    \sL^{(i)}(\sM_p^{(0)}) \times \epsilon_p \leq \sL^{(i)}(\real^S)\times \epsilon_p \leq \LRp{\prod_{k=1}^3 \norm{\bW^{(k)}_{{i}}}_2}\times \epsilon_p\\
       &  \leq \ell^3\times \LRs{\prod_{k=1}^3 \LRp{\sum_{l=0}^m\norm{\LRp{\bg^{(k)}_l}_i}_2}}\times \epsilon_p
    \end{aligned}
     \label{gradient_local}
\end{equation}
where the last line follows from the standard gradient descent update formulae and applying a triangle inequality, $\LRp{\bg^{(k)}_l}_i$ represents the gradients for the $k^{th}$ layer weight matrix at the $l^{th}$ iteration of gradient descent algorithm for network $\sQ_i$. Applying the transformation $\zeta=\frac{1}{m+1}$, and combining \eqref{comb_1} and \eqref{gradient_local} we can recover the stability function that satisfies condition \ref{qw_22}
as follows:
\[ \norm{\sQ\LRp{\bx}-\sQ\LRp{\bx'}}_2 \leq  \sum_{i=2}^{r-1} \ell^3\times \LRs{\prod_{k=1}^3 \LRp{\sum_{l=0}^{(1/\zeta)-1}\norm{\LRp{\bg^{(k)}_l}_i}_2}}\times \epsilon_p=\sum_{i=2}^{r-1}\delta_p^{(i)}(\zeta)=\delta_p^{\sQ}(\zeta).  \]
Now in order to ensure that $\delta_j^{\sQ}(\zeta)$ is a valid stability function as per \cref{discrete_ep_del}, it is enough  to check if each  $\delta_p^{(i)}(\zeta)$ satisfies condition \ref{qw_1} in \cref{delta-def}, and condition \ref{qw_discrete} in \cref{discrete_ep_del}.

Note that  $\delta_j^{(i)}(\zeta)$ is now a discrete function with domain $\Big \{\frac{1}{m^u+1},\ \frac{1}{m^u},\dots 1\Big \}$, where  $m^u \in \mathbb{Z}^+$ is some chosen upper bound for the training iterations. Note that $\delta_p^{(i)}(1)=0$ since $\LRp{\bg^{(3)}_0}_i={\bf{0}}$ due to the output layer being initialized with zeros in \cref{Sequential}. Further, $\delta_p^{(i)}\LRp{\frac{1}{m^u+1}}>0$ since the gradients in the subsequent iterations must be non-zero for decreasing the loss. Therefore, we have $\delta_p^{(i)}(1)<\delta_p^{(i)}\LRp{\frac{1}{m^u+1}}$ satisfying condition \ref{qw_1} in \cref{delta-def}.
In addition, it is clear that $\delta_p^{(i)}(\zeta)$ is monotonically decreasing with respect to $\zeta$ thereby satisfying condition \ref{qw_discrete} in \cref{discrete_ep_del}.
%with respect to $\zeta$ when the domain $\Big \{\frac{1}{m^u+1},\ \frac{1}{m^u},\dots 1\Big \}$ is equipped with  the discrete topology thereby satisfying condition \ref{qw}.

%https://math.stackexchange.com/questions/1895906/proof-for-continuous-function-on-a-discrete-topology#:~:text=If%20X%20is%20equipped%20with,subset%20of%20X%20is%20open.

%such that the decrease in stability between two adjacent points can precisely be quantified as:

%\[\delta_j^{(i)}\LRp{\frac{1}{m+1}}-\delta_j^{(i)}\LRp{\frac{1}{m}}=\]

%Further, $\delta_j^{(i)}(\zeta)$ is well-defined $\forall \zeta$. 
\end{proof}

\end{section}

\begin{section}{General setting for numerical experiments}{}
\label{general_setting}
All codes were written in Python using TensorFlow. Throughout the study, we have employed the Adam optimizer \cite{kingma2014adam} for minimizing the cost function.  The manifold regularization term is computed over mini-batches.  We consider a decay of manifold regularization weight $\gamma^{(i+1)}=0.5\times \gamma^{(i)}$ on adding a new layer unless otherwise stated. We choose $\beta_p=1,\ \forall p$ in \eqref{manifold_strength}. Our proposed approach is compared with a number of different approaches.
Description of different algorithms adopted for comparison with the proposed approach
\begin{equation*}
 \begin{aligned}
\text{Algo. \ref{AlgoGreedyLayerwiseResNet}}&: \text{layerwise training algorithm described in  \cref{reguls}}.\\
\text{ Algo. \ref{Sequential} }&: \text{Sequential residual learning  after Algo. \ref{AlgoGreedyLayerwiseResNet} (\cref{seq_des})}.\\
\text{Baseline network}&: \text{A fully-connected  deep neural network with the same depth and}\\
& \text{\ \ \ width as  that produced by \cref{AlgoGreedyLayerwiseResNet}} .\\
\text{ Algo. \ref{Sequential}  (1 network)}&: \text{Sequential residual learning after Algo. \ref{AlgoGreedyLayerwiseResNet} using  1 network }\\
& \text{\ \ \ in  Algo. \ref{Sequential} trained for a large number of epochs.} \\
%\text{Regularized baseline}&: \text{Baseline network regularized with sparsity and manifold regularization}\\
%& \text{\ \ \ on the output layer.}\\
\text{Baseline + Algo. \ref{Sequential}}&: \text{\cref{Sequential} applied on top of the residuals produced by}\\
& \text{\ \ \ the baseline network}.\\
\text{Forward Thinking}&: \text{Algorithm for layerwise adaptation by  Hettinger et al. \cite{hettinger2017forward}. This}\\
& \text{\ \ \ strategy is also similar to the algorithm by Belilovsky et al. \cite{belilovsky2019greedy}  }\\
& \text{\ \ \ and Trinh et al. \cite{trinh2019greedy}.} \\
\text{Approach  \cite{hettinger2017forward}+ Algo. \ref{Sequential}}&: \text{\cref{Sequential} applied on top of the residuals produced by}\\
& \text{\ \ \ Forward thinking \cite{hettinger2017forward}}.\\
\text{Architecture Search (AS)}&: \text{Random search with early stopping proposed by  Liam Li  }\\
& \text{\ \ \ and Ameet Talwalkar \cite{li2020random}.}\\
\end{aligned}  
\end{equation*}
When training the baseline,   $100$ different random initializations are considered and the best result is chosen for comparison.  For architecture search (AS), the search space is defined by setting the maximum width (for each hidden layer) and maximum depth the same as 
 that produced by  \cref{AlgoGreedyLayerwiseResNet}.

\begin{subsection}{Choice of $\sR^{(i)}$}{}
\label{choice}
For the results provided in this work, we choose  $\sR^{(i)}$ in \eqref{Res_two} as the identity map for computational efficiency. This avoids recomputing the map $\sR^{(i)}$ in each training iteration. We observed numerically that the feature scaling  introduced in (\ref{gl_3}) from a pure theoretical purpose did not provide improvement in results.
\end{subsection}

\begin{subsection}{Tuning parameters $\gamma$ and $\zeta$ for $\delta-$stability}{}
\label{tuning}
Note that the  hyperparameters $\gamma$ (in \cref{AlgoGreedyLayerwiseResNet}) and $E_e$ (in \cref{Sequential}) are tuned based on the validation dataset to achieve the best results. The best parameters are the ones that minimizes the validation loss such as depicted in Figure \ref{Boston_training}. Note that tuning parameter $E_e$ in  \cref{Sequential} corresponds to tuning the parameter $\zeta$ in proposition \ref{Lipschitz}. We consider a strategy of training the first two networks in \cref{Sequential} for $E_e^1$ number of epochs and training the remaining networks  for $E_e^2$ number of epochs. This leads to tuning $\{E_e^1,\ E_e^2\}$ in \cref{Sequential}.  That is, \cref{Sequential} is run for different choices of $\{E_e^1,\ E_e^2\}$ and the best choice is the one that provides the lowest validation loss. This strategy is adopted since the residuals becomes increasingly nonlinear and subsequent networks requires more number of epochs to promote training (i.e decrease the loss).
\end{subsection}

\end{section}

\begin{section}{Boston house price prediction problem}{}
\label{boston_setting}
For this problem, the manifold regularization is computed by employing the K-means clustering algorithm on the input features \cite{macqueen1967some} and the number of clusters $K$ is chosen as $5$.  
\end{section}

\begin{section}{Physics informed adaptive neural network (PIANN)}
\label{PIANN_a}
This section contains additional details on the PIANN problem discussed in \cref{PIANN}.  The relative $L^2$ error achieved after adding each layer is shown in Table \ref{PDE_extended_accuracy} (case a) and  Table \ref{PDE_algo_complex} (case b). The active parameters in each hidden layer  is shown in  Table \ref{zeros_PDE} (case a) and Table \ref{zeros_PDE_complex} (case b).
The  results achieved by the baseline network as well as other layerwise training method \cite{hettinger2017forward} is also provided in Table \ref{tab:adaptation_base_PINNs_b} (case a) and Table \ref{tab:adaptation_base_PINNs} (case b).

\begin{table}[h!]
\hspace{0.5 cm}
%   \resizebox{10cm}{!}{
    \begin{minipage}{.5\textwidth}
    \centering
        \caption{Relative $L^2$ error on layer addition (Case a)}
       	\label{PDE_extended_accuracy}
           	\begin{tabular}{|c | c | c | }
           	\hline
		 Layer  & $\tau^{(i)}$   &  Relative \\
No.   &   &      $L^2$ error \\
		\hline
		2 & 10 &$1.723\times 10^{-1}$\\
		3 &  15&  $2.254 \times 10^{-2}$\\
		4 &  20&$4.974 \times 10^{-3}$\\
		5 &  25& $6.649 \times 10^{-4}$ \\
		6 &  30& $2.660 \times 10^{-4}$\\
		7 &  35& $6.720 \times 10^{-5}$ \\
  \hline
	\end{tabular} 
    \end{minipage}%
    \hspace{-0.7 cm}
    \begin{minipage}{.5\textwidth}
    \centering    
        \caption{$\%$ of non-zero parameters in each layer (Case a)}
         \label{zeros_PDE}
       	\begin{tabular}{|c | c |}
       	\hline
		 Layer  & $\%$ of non-zero  \\
    No. & parameters \\
		\hline
			2 &   99.5 $\%$\\
		3 &  97.7 $\%$\\
		4 & 98.9 $\%$\\
		5 &  95.4 $\%$\\
		6 &  92.5 $\%$\\
		7 &  96.7 $\%$ \\
  \hline
	\end{tabular} 
    \end{minipage}
%}  
\end{table}

\begin{table}[h!]
    \centering
             \caption{Performance of Baseline Network and other layerwise training methods (Case a)}
  \label{tab:adaptation_base_PINNs_b}
       	\begin{tabular}{|c | c | c |}
        \hline
    Method &  Relative $L^2$ error  &  Parameters  trained  \\ 
        &    &  simultaneously  \\ \hline
            Proposed method & $6.7\times 10^{-5}$  & $10,501$\\
    Baseline  network & $1.0\times 10^{-4}$& $61, 001$ \\
Forward Thinking \cite{hettinger2017forward} & $9.2\times 10^{-2}$&  $10,501$ \\ \hline
	\end{tabular} 
\end{table}

\begin{table}[h!]
\hspace{0.5 cm}
%   \resizebox{10cm}{!}{
    \begin{minipage}{.5\textwidth}
    \centering
        \caption{Relative $L^2$ error on layer addition (Case b)}
       	\label{PDE_algo_complex}
           	\begin{tabular}{|c | c | c | }
           	\hline
		 Layer  & $\tau^{(i)}$  &  Relative  \\
    No. &  &     $L^2$ error \\
		\hline
		2 & 5 &  $3.59\times 10^{-1}$\\
		3 &  10&  $4.75 \times 10^{-2}$\\
		4 &  15&$1.13 \times 10^{-3}$\\
		5 &  20& $7.86 \times 10^{-5}$ \\
  \hline
	\end{tabular} 
    \end{minipage}%
    \hspace{-0.7 cm}
    \begin{minipage}{.5\textwidth}
    \centering    
        \caption{$\%$ of non-zero parameters in each layer (Case b)}
         \label{zeros_PDE_complex}
       	\begin{tabular}{|c | c |}
       	\hline
		 Layer  &  $\%$ of non-zero  \\
   		 No. &   parameters \\
		\hline
		2 &   97.52 $\%$\\
		3 &  91.29 $\%$\\
		4 & 97.35 $\%$\\
		5 &  90.00$\%$\\
  \hline
	\end{tabular} 
    \end{minipage}
%}  
\end{table}

\begin{table}[h!]
    \centering
             \caption{Performance of Baseline Network and other layerwise training methods (Case b)}
  \label{tab:adaptation_base_PINNs}
       	\begin{tabular}{|c | c | c |}
        \hline
    Method &  Relative $L^2$ error  &  Parameters  trained  \\ 
        &    &  simultaneously  \\ \hline
            Proposed method & $7.8\times 10^{-5}$  & $10,501$\\
    Baseline  network & $1.1\times 10^{-4}$& $40, 801$ \\
Forward Thinking \cite{hettinger2017forward} & $1.6\times 10^{-2}$&  $10,501$ \\ \hline
	\end{tabular} 
\end{table}

\end{section}

\begin{section}{Interpretability of hidden layers and  transfer learning strategy}
\label{transfer_learning}
In this section, we demonstrate how the interpretability of different layers in PIANN helps one in devising efficient transfer learning strategies and make informed decisions on which layers to re-train.   For demonstration, consider the operator in (\ref{darc}) with $a(x_1,\ x_2)=e^{x_1+x_2}$ and $f=200$. We reuse the pre-trained model that learnt the solution in Figure \ref{Algo_I_complex_PDE}. The objective is to correct this pre-trained model for the  new coefficient $a(x_1,\ x_2)=e^{x_1+x_2}$. 

\begin{figure}[h!]      
  % \begin{subfigure}[b]{\textwidth}
  \hspace{-0.5 cm}
  \begin{tabular}{c}

      \begin{tabular}{c}

          \centering
          \includegraphics[scale=0.35]{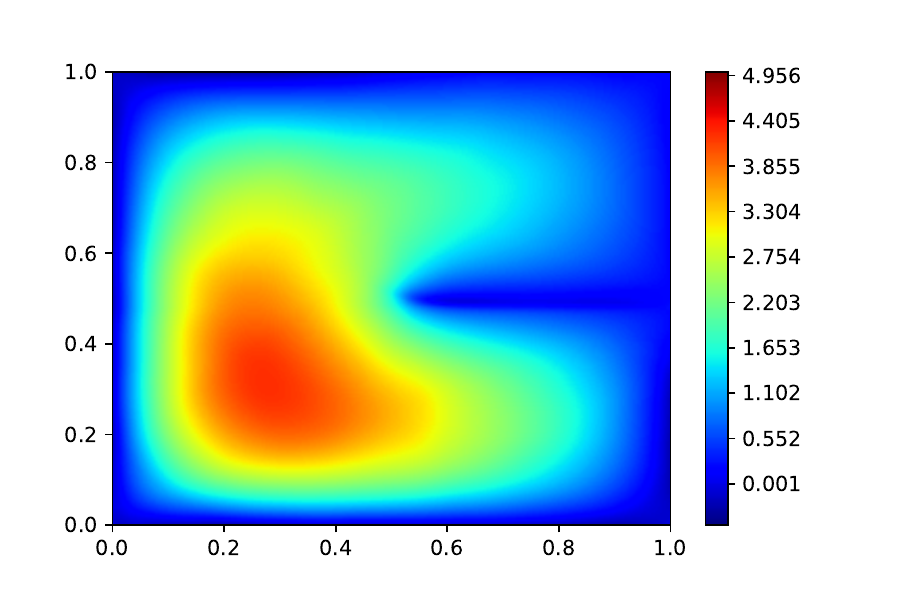}
          % \caption{Effect of noise on DI and Tikhonov solutions}
          % \figlab{some_other_good_name}

      \end{tabular}

    \hspace{-0.5 cm}

      \begin{tabular}{c}

          \centering
          \includegraphics[scale=0.35]{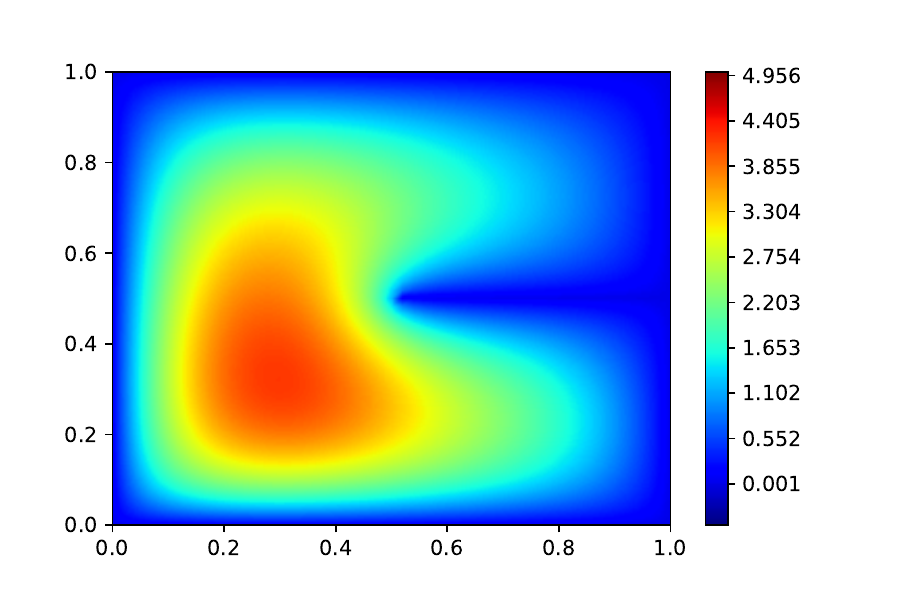}
          % \caption{Effect of noise on DI and Tikhonov solutions}
          % \figlab{some_good_name}

      \end{tabular}

    \hspace{-0.5 cm}

      \begin{tabular}{c}

          \centering
          \includegraphics[scale=0.35]{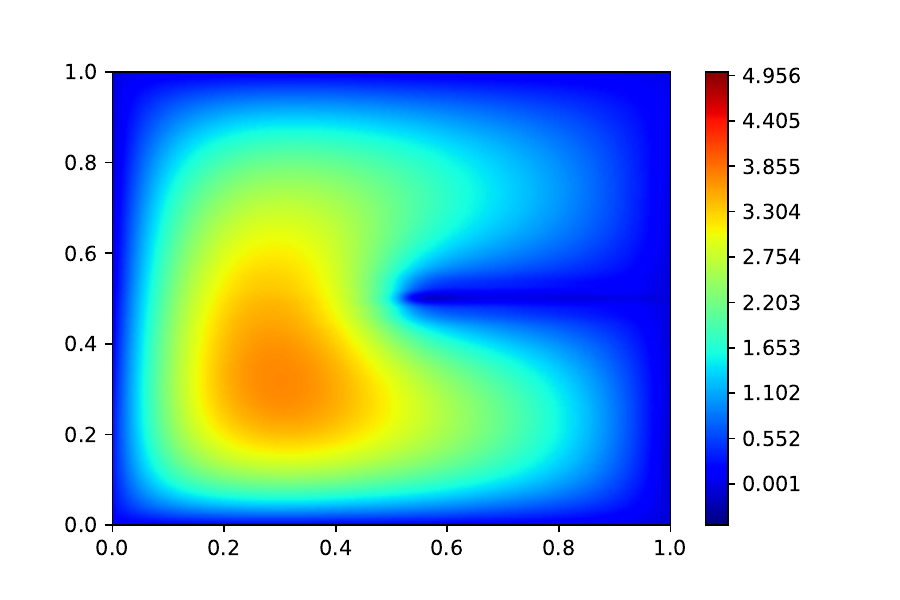}
          % \caption{Effect of noise on DI and Tikhonov solutions}
          % \figlab{some_good_name}

      \end{tabular}

  \end{tabular}
  \caption{Transfer learning strategy. Left to right:  Transfer learning on interpretable network achieved by our approach  (relative $L^2$ error= $4.432\times 10^{-3}$  at the end of $100$ epochs); True solution; Traditional transfer learning strategy on a baseline (relative $L^2$ error= $8.72\times 10^{-3}$  at the end of $100$ epochs). }
  \label{Transfer_Learning}
  % \end{subfigure} \\
\end{figure} 
Since in our approach (PIANN), the first few layers focus mainly on enforcing the boundary conditions (Figure \ref{Algo_I_complex_PDE}) it is anticipated that only the last layers needs retraining for model correction. In our approach for transfer learning, we consider retraining the last two layers reinitialized with zero weights and biases. The relative $L^2$ error achieved and the learnt  solution after  retraining the last two layers for $100$ epochs is shown in  Figure \ref{Transfer_Learning}. In addition, we also consider the traditional transfer learning strategy of retraining the last two layers of a pre-trained baseline network that lacks the interpretability of its layers \cite{subel2022explaining}. Note that in this case it is unclear on the best layers to re-train and a discussion on this aspect is provided in \cite{subel2022explaining}.  Comparison of efficiency between the two approaches is shown in Figure \ref{Transfer_Learning}. It is clear from Figure \ref{Transfer_Learning} that our approach learns the new solution faster in comparison to the traditional transfer learning approach.

\end{section}

\begin{section}{Performance on MNIST dataset}
\label{MNIST_appendix}

\subsection{Training with 20 neurons in each hidden layer}

 The inputs for this experiment is given in \cref{inputs}.   The  parameter efficiency achieved for the case of using 20 neurons in the hidden layers is provided in 
Figure \ref{MNIST_training}.
  Table \ref{tab:adaptation_MNIST} shows the comparison of results with baseline and other layerwise training method.

\begin{figure}[h!]      
  % \begin{subfigure}[b]{\textwidth}
\hspace{-0.5 cm}
  \begin{tabular}{c}

      \begin{tabular}{c}

          \centering
          \includegraphics[scale=0.5]{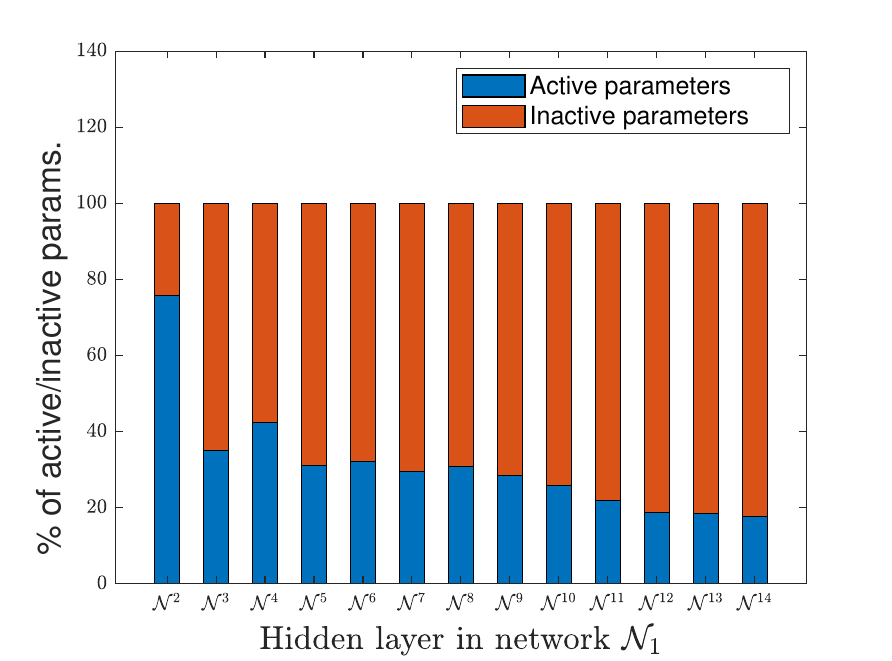}
          % \caption{Effect of noise on DI and Tikhonov solutions}
          % \figlab{some_other_good_name}

      \end{tabular}

\hspace{0.0 cm}
      \begin{tabular}{c}
          \centering
            \includegraphics[scale=0.5]{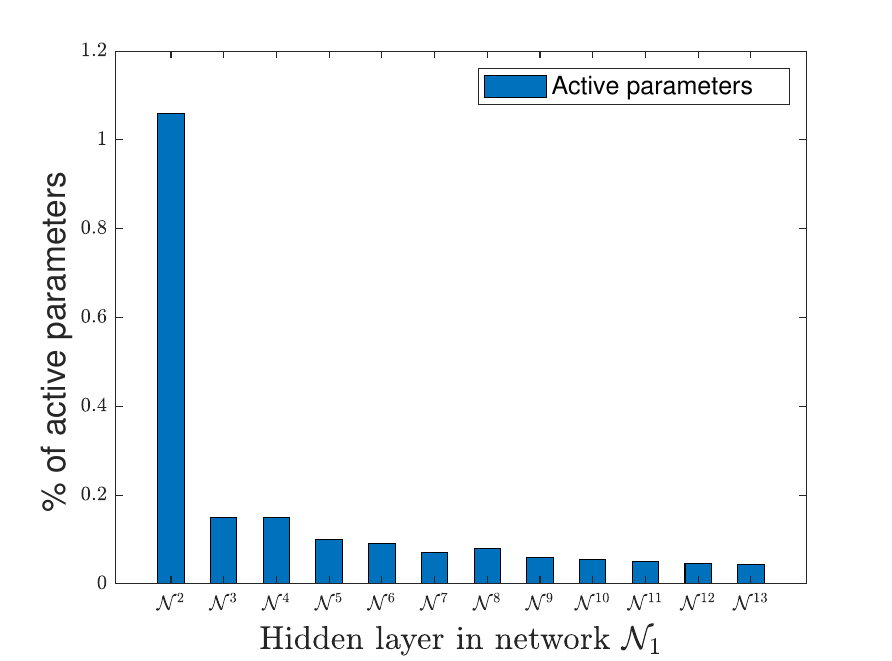}
          % \caption{Effect of noise on DI and Tikhonov solutions}
          % \figlab{some_good_name}

      \end{tabular}

  \end{tabular}
  \caption{ Left to right: Active and inactive parameters in each hidden layer (20 neurons in each hidden layer); Active and inactive parameters in each hidden layer (500 neurons in each hidden layer) }
  \label{MNIST_training}
  % \end{subfigure} \\
\end{figure}

\begin{table}[h!]
    \centering
    \caption{MNIST: Performance of baseline network and other layerwise training methods (20 neuron case)}
  \label{tab:adaptation_MNIST}
       	\begin{tabular}{|c | c | c |}
        \hline
    Method &  Test Accuracy &  Parameters trained\\ 
    &$\%$  &  simultaneously \\ \hline
    Proposed method &  96.90  & 16,330  \\ \hline
    Baseline network & 96.00 & 21,370 \\ \hline
     Baseline + Algo. \ref{Sequential}  & 95.90 &  21,370 \\ \hline
    Forward Thinking \cite{hettinger2017forward}  & 95.90 & 16,330 \\\hline
    Approach \cite{hettinger2017forward} + Algo. \ref{Sequential}  & 96.00 & 16,330 \\\hline
    Architecture search (AS) & 96.47 & - \\\hline
	\end{tabular} 
\end{table}
Note that our experiments showed that,  forward thinking model \cite{hettinger2017forward}  quickly over-fits on the training data while adding the initial few layers. Further, forward thinking model is not $\delta-$ stable which is a necessary condition for  \cref{Sequential} to achieve robustness (Remark \ref{stability_2}). Numerically, it is observed that the residual defined in line 4 of  \cref{Sequential} is very low in magnitude, and it is hardly possible to extract information from the residual and improve the results. Hence, the performance of forward thinking model +  \cref{Sequential} has been not been included in the comparison results shown in Figure \ref{summary_PINNs_a_b} since the behaviour is similar to baseline+\cref{Sequential}. 

% Algo. \ref{Sequential}  (1 network)   & 96.30 & 16, 330 \\ \hline

\subsection{Training with 500 neurons in each hidden layer}

The parameter efficiency and the comparison of results with baseline is shown in Figure \ref{MNIST_training} and Table \ref{tab:adaptation_MNIST_new}.

\begin{table}[h!t!b!]      
\centering
\caption{MNIST: Performance of baseline network and other layerwise training methods (500 neuron case).}
  \label{tab:adaptation_MNIST_new}
           	\begin{tabular}{|c | c | c |}
        \hline
    Method &  Test    &  Params.  trained  \\ 
        &  accuracy $\%$  &  simultaneously  \\ \hline
            Proposed method  & 98.7 & $6,48010$\\ \hline
    Baseline  network & 98.2 & $34, 03510$ \\ \hline
     Baseline + Algo. \ref{Sequential}  & 98.2 & $34, 03510$ \\ \hline
Forward  Thinking \cite{hettinger2017forward}  & 98.3 &  $6,48010$\\ \hline
Approach \cite{hettinger2017forward} + Algo. \ref{Sequential}  & 98.4 &  $6,48010$\\ \hline
Architecture search (AS) & 98.3 & - \\\hline
	\end{tabular} 
  % \end{subfigure} \\
\end{table}

%  Algo. \ref{Sequential}  (1 network)   & 98.3 & $6,48010$ \\ \hline

\end{section}

\begin{section}{Details of input parameter values for different problems}{}
\label{inputs}
The description of each problem is given below:

\begin{equation*}
 \begin{aligned}
\text{I}&: \text{Boston house price prediction problem}\\
\text{II (a)}&: \text{Physics informed adaptive neural network (PIANN), Symmetric boundary}.\\
\text{II (b)}&: \text{Physics informed adaptive neural network (PIANN), Slit in the Domain}.\\
%\text{III}&: \text{Physics reinforced adaptive neural network (PRANN).}\\
\text{III (a)}&: \text{Adaptive learning for inverse problems (20 data-set problem).}\\
\text{III (b)}&: \text{Adaptive learning for inverse problems (50 data-set problem).}\\
\text{IV (a)}&: \text{MNIST classification (20 neuron in hidden layers).}\\
\text{IV (b)}&: \text{MNIST classification (500 neuron in hidden layers).}\\
\end{aligned}  
\end{equation*}
  
\begin{table}[h!]
	\caption{Details of input parameters in  \cref{AlgoGreedyLayerwiseResNet} for different problems} \label{Input_values_different_problems}
	\renewcommand{\arraystretch}{1.3}
	\centering	
	\begin{tabular}{|c | c | c | c | c | c | c | c | c |}
 \hline
		\centering
		 Problem &$\epsilon_\eta$ &$\rho$ &$(\alpha^{(2)}, \tau^{(2)}, \gamma^{(2)})$ & $\ell$ & $d_l$ & $E_e$ & $N_o$ & Batch size\\
		\hline
		I\phantom{I(a)}& $0.035$ &$10^{-6}$ & $(0.05,\ 0,\ 0.06)$& $0.001$ & $0.9$ &$50$& $100$ & $70$ \\
		II (a) &  $0.8$&$10^{-6}$ & $(0.001,\ 10,  \ 0)$& $ 0.0005$ & $1 $ & $ 1000 $& $100$ & $500$\\
		II (b)& $0.5$& $10^{-6}$ & $(0.001,\ 5,  \ 0)$& $0.0005$ & $1 $ & $ 2000 $&$100$ & $500$ \\
		III (a)& 0.035& $10^{-6}$  & $(0.0001,\ 0,  \ 0.001)$& $0.001$ & $1 $ & $400$& $100$ & $20$ \\
		III (b)& 0.0015& $10^{-6}$  & $(0.0001,\ 0,  \ 0.001)$& $0.001$ & $1 $ & $400$& $100$ & $50$ \\
		IV (a)& $0.005$&  $10^{-6}$ &$(0.001,\ 0,\ 0.0035)$& $0.001$ & $0.9$ & $100$& $20$& $900$ \\
		IV (b)& $0.005$&  $10^{-6}$ &$(0.001,\ 0,\ 0.005)$& $0.001$ & $0.9$ & $100$& $500$& $900$ \\
  \hline
	\end{tabular} 
\end{table}
Details of input parameters used in  \cref{AlgoGreedyLayerwiseResNet} is provided in Table \ref{Input_values_different_problems}. $\ell$ represents the learning rate, $d_l$ represent the decay of learning rate, and $E_e$ represents the number of epochs for which each layer is trained. Details of input parameters used in \cref{Sequential} is provided in Table \ref{Input_values_algo_II}. In Table \ref{Input_values_algo_II}, Epoch $\{E_e^1,\ E_e^2\}$ means that the first two networks in \cref{Sequential} is trained for $E_e^1$ number of epochs and the remaining networks are trained for $E_e^2$ number of epochs.  
%\begin{table}[h!]
%	\caption{Details of input parameters in \cref{Sequential} for different problems} \label{Input_values_algo_II}
%	\renewcommand{\arraystretch}{1.3}
%	\centering	
%	\begin{tabular}{|c | c | c | c | c | c | c | c | c |}
% \hline
%		\centering
%		 Problem &$\epsilon_e$ &$N_n$ &Network \tablefootnote{FC represents fully connected network.} & Activation & No. of   & No. of & Epoch & Batch\\
 %  & & & & &  layers &  neurons &$\{ \text{lower},\ \text{higher}\}$ & size \\
%		\hline
%		I\phantom{I(a)}& $0.1$ &$5$ & \textit{FC}& \textit{relu} & $1$ &$10$& $\{70,\ 100\}$&70 \\
%		IV (a)& $0.05$&  $20$ &\textit{FC}&\textit{elu} & $2$ & $20$ &$\{100,\ 200\}$& 900 \\
%		IV (b) & $0.05$&  $4$ &\textit{FC}&\textit{elu} & $2$ & $500$ & $\{200,\ 400\}$& 900\\
%  \hline
%	\end{tabular} 
%\end{table}

\begin{table}[h!]
	\caption{Details of input parameters in \cref{Sequential} for different problems} \label{Input_values_algo_II}
	\renewcommand{\arraystretch}{1.3}
	\centering	
	\begin{tabular}{|c | c | c | c | c | c | c | c | }
 \hline
		\centering
		 Problem &$N_n$ &Network \tablefootnote{FC represents fully connected network.} & Activation & No. of   & No. of & Epoch $E_e$& Batch\\
   & & & &   layers &  neurons &$\{ E_e^1,\ E_e^2\}$ & size \\
		\hline
		I\phantom{I(a)} &$5$ & \textit{FC}& \textit{relu} & $1$ &$10$& $\{70,\ 100\}$&70 \\
		IV (a)&  $20$ &\textit{FC}&\textit{elu} & $2$ & $20$ &$\{100,\ 200\}$& 900 \\
		IV (b) & $4$ &\textit{FC}&\textit{elu} & $2$ & $500$ & $\{200,\ 400\}$& 900\\
  \hline
	\end{tabular} 
\end{table}

\end{section}

\end{document}